\pdfoutput=1

\documentclass[11pt]{article}

\usepackage[preprint]{acl}
\usepackage{authblk}

\usepackage{times}
\usepackage{latexsym}
\usepackage{color-edits}
\addauthor{sw}{blue}

\usepackage[T1]{fontenc}

\usepackage[utf8]{inputenc}

\usepackage{microtype}

\usepackage{inconsolata}

\usepackage{graphicx}

\usepackage{tabularx}
\usepackage{booktabs}
\usepackage{subcaption}
\usepackage{placeins}
\usepackage{amsmath}
\usepackage{amsfonts}

\usepackage{float}

\title{
Predicting Language Models’ Success at Zero-Shot Probabilistic Prediction
}

\author{
 \textbf{Kevin Ren \textsuperscript{1*,}$^{\dagger}$},
 \textbf{Santiago Cortes-Gomez\textsuperscript{2}},
 \textbf{Carlos Miguel Patiño\textsuperscript{2}},
 \textbf{Ananya Joshi\textsuperscript{2}},
\\
 \textbf{Ruiqi Lyu\textsuperscript{2}},
 \textbf{Jingjing Tang\textsuperscript{2}},
 \textbf{Alistair Turcan\textsuperscript{2}},
 \textbf{Khurram Yamin\textsuperscript{2}},
\\
 \textbf{Steven Wu\textsuperscript{2}},
 \textbf{Bryan Wilder\textsuperscript{2}}
\\
\\
 \textsuperscript{1}Cornell Tech,
 \textsuperscript{2}Carnegie Mellon University
 \\
 \textsuperscript{$^{\dagger}$}Work done while at Carnegie Mellon University
\\
 \small{
   \textbf{Correspondence:} \href{mailto:kevinren@cs.cornell.edu}{kevinren@cs.cornell.edu}
 }
}

\begin{document}
\maketitle
\begin{abstract}
    Recent work has investigated the capabilities of large language models (LLMs) as zero-shot models for generating individual-level characteristics (e.g., to serve as risk models or augment survey datasets). However, when should a user have confidence that an LLM will provide high-quality predictions for their particular task? To address this question, we conduct a large-scale empirical study of LLMs' zero-shot predictive capabilities across a wide range of tabular prediction tasks. We find that LLMs' performance is highly variable, both on tasks within the same dataset and across different datasets. However, when the LLM performs well on the base prediction task, its predicted probabilities become a stronger signal for individual-level accuracy. Then, we construct metrics to predict LLMs' performance at the \textit{task} level, aiming to distinguish between tasks where LLMs may perform well and where they are likely unsuitable. We find that some of these metrics, each of which are assessed without labeled data, yield strong signals of LLMs' predictive performance on new tasks \footnote{We release our code at \\  \href{https://github.com/kkr36/llm-eval/tree/camera-ready}{\texttt{https://github.com/kkr36/llm-eval/tree/camera-ready}}.}.

\end{abstract}

\section{Introduction}

    There is increasing interest in using large language models (LLMs) as predictive models, leveraging the world knowledge encoded by their pretraining corpora to make zero-shot predictions in domains without any labeled data. While this predictive capability was first investigated for traditional tasks within Natural Language Processing (NLP), such as text classification or question-answering  \cite{wang2023_zeroshot}, recent work has utilized LLMs as predictive models in a broader sense. For instance, LLMs have been used to provide medical risk scores \citep{chung2024large}, predict fraud risk in financial applications \citep{xie2024finben} and impute unsurveyed fields in social science surveys \citep{park2024generative, dominguez2024questioning}. More generally, LLMs can effectively consume text serializations of tabular data; the prevalence of tabular data across many domains likely contributes to this increasing interest across application areas. These applications differ from traditional text-based tasks \citep{cruz2024_riskscores} because the label is not determined fully by the input: people with identical features may have different outcomes. We refer to tasks with this property as \textit{probabilistic prediction}, and the predicted probabilities from the LLM as \textit{risk scores}.
   
   While the zero-shot prediction capabilities of LLMs offer exciting opportunities to scientists and practitioners, it is likely (as we empirically verify) that LLMs' performance varies widely across settings. Then, how can practitioners tell whether an LLM will perform well as a predictive model, prior to observing labeled data? This is a question with no easy answer. The appeal of using a pretrained model in many domains lies in avoiding the cost of collecting labeled data. However, validating conclusions from foundation models without labeled-data confirmation is far from straightforward.

   This challenge is especially pronounced in the fully zero-shot case, where users lack access to ground-truth labels altogether. We distinguish performance at two levels of granularity: at the \textbf{individual} level, referring to which examples an LLM is likely to predict accurately, and at the \textbf{task} level, referring to which overall prediction problems, defined by a dataset and outcome variable, the LLM is likely to perform well on. The ability to quantify uncertainty at both levels allows practitioners to judge which individuals and overall predictive tasks may result in inaccurate predictions. 

    Previous work has primarily studied uncertainty at the individual level, finding mixed results. Abstention methods use measures of individual-level confidence  to flag dubious predictions that should be examined manually by a human expert, or ignored altogether ~\cite{tomani2024uncertainty, feng2024don}. However, both answer-token probabilities and verbalized confidence scores from LLMs have been found to be badly calibrated for probabilistic prediction  \cite{cruz2024_riskscores} and also for a variety of question-answering tasks \cite{xiong2023can}, typically due to overconfidence. Despite this, multiple approaches train a post-processing step to improve calibration using only the outputs or last-layer representations of models \cite{shen2024thermometer,ulmer2024calibrating}. Confidence scores have also been found to be useful in conformal prediction frameworks \cite{kumar2023conformal,mohri2024language}, suggesting that they can be post-processed to yield informative decisions about when to provide specific information. 

    Analogously, practitioners may wish to know whether a task is likely suitable for an LLM before using its outputs, via some metric of uncertainty at the task level. Yet, to our knowledge, no previous work considers uncertainty quantification at the task level, at least in the context of probabilistic prediction. This presents a significant challenge, as in many real-world scenarios, practitioners would benefit from heuristics to assess whether LLMs will perform well \textit{a priori}. However, doing so typically requires labeled data—a costly resource that pretrained models are meant to help avoid.

    In this work, we conduct a large-scale empirical study on the performance of LLMs for probabilistic prediction on 316 tasks across 31 tabular datasets. The primary question we ask is: \textbf{given only unlabeled data, is it possible to anticipate how well the model will perform on a zero-shot prediction task?} We provide the first empirical evidence using task-level strategies to assess signals of LLM performance across prediction tasks. Additionally, we provide more nuanced results about individual-level uncertainty quantification; previous results on LLM calibration for probabilistic prediction \cite{cruz2024_riskscores} are restricted to data from the US Census while we employ a much larger number of tabular datasets across many subject areas. Our empirical study reveals several findings that can inform how LLMs are employed and evaluated in predictive settings:
    \begin{enumerate}
        \item The \textit{distribution} of LLMs'  predictions on unlabeled data encodes substantial information about their suitability for a task. We propose simple heuristics and more elaborate model-based strategies that provide a strong signal of LLMs' predictive performance, using only unlabeled data. While we do not suggest that practitioners forgo labeled-data evaluation in high-stakes settings, our results could be useful to provide an initial assessment of which candidates from a set of prediction tasks are more promising for further development—or to screen out applications that have a lower chance of success.
        \item At the task level, naive ``elicited confidence" strategies (e.g., asking LLMs to rate their skill level given a description of the task) are comparatively unreliable predictors of success. 
        \item Substantial variation in LLMs' performance on different prediction tasks is \textit{not} explained by broader patterns of ``subject matter expertise"; within different tasks defined on the same dataset, predictive performance exhibits very high variance. This implies that attempts to validate LLMs' suitability must be specific to individual predictive tasks, and should not solely utilize information at a dataset or general subject level. For example, validating a social simulator by demonstrating that the LLM predicts observed fields well carries a high degree of risk because success on observed fields often fails to generalize to success on a specific, unobserved field.
        \item At the individual level, LLMs' responses to probabilistic prediction tasks are typically poorly calibrated. Beyond overconfidence as reported in previous work \cite{cruz2024_riskscores}, we find that LLMs' responses in a given domain are often describable as simply being over- or under-predictions, where risk scores are consistently too large or too small.  
        \item Despite a lack of calibration in individual-level predictions, in many tasks, individual-level responses still provide an informative signal for abstention decisions because LLMs are more accurate on examples for which they output more extreme risk scores. This conclusion empirically holds even if the numerical scale of the scores is highly distorted. This echoes our first two findings at the task level: LLMs' responses contain considerable latent information about performance at both levels, but this information often requires postprocessing to elicit meaningful results.
    \end{enumerate}

    Our results provide a pathway towards more rigorous decisions about which tasks and individual instances are appropriate for LLMs.

\section{Related Work}
\paragraph{LLMs for Tabular Data:}
Recent work has shown that LLMs can effectively process tabular data using simple prompting strategies, achieving strong performance \citep{hegselmann2023_tabllm}. Pretrained models like TaBERT \citep{yin2020_tabert}, TAPAS \citep{herzig2020_tapas}, and TURL \citep{deng2021_turl} focus on tabular data for QA tasks, while others leverage chain-of-thought prompting \citep{sui2023_tablemeets, jin2023_tabcot} and fact verification \citep{chen2019_tabfact,eisenschlos2020_intermediate}. Broader generalization strategies include UniPredict \citep{wang2023_unipredict} and instruction tuning \citep{yang2024_predictive}. More recent efforts highlight LLMs’ ability to perform zero-shot tabular predictions \citep{shi2024_surprisingly, wen2023_gtl, gardner2024_tabulal8b}. As opposed to developing methods to optimize LLMs for the purposes of understanding tabular data, our work seeks to empirically distinguish general factors predicting LLMs' success and failure across prediction tasks.

\paragraph{Elicited Confidence Scores From LLMs:} LLM predictions on tabular data can suffer from pretraining-induced biases~\cite{liu-etal-2024-confronting}, and their uncertainty estimates are often poorly calibrated~\cite{cruz2024_riskscores}. Methods like multicalibration and prompt-based scoring~\cite{xiong2023can, detommaso2024multicalibration} aim to improve calibration. In contrast to prior work, we primarily study uncertainty estimation at the task level. En route, we also provide a more nuanced picture of individual-level uncertainty on a wider range of tasks than previous work.

\section{Methods}

We describe our experimental setup, the problem of predicting LLM performance, and the set of proxy methods that we assess for performance prediction.

\subsection{Experimental Setup}

We conduct experiments on 31 tabular datasets spanning domains such as social surveys, finance, medicine, and transportation (see Appendix \ref{appendix:datadesc} for details). Each dataset is associated with a binary classification task. Using the \texttt{folktexts} library \cite{cruz2024_riskscores}, we serialize 1,000 randomly sampled rows per dataset (or the full dataset if smaller) into text prompts, followed by a multiple-choice question requesting the label. Predicted probabilities (risk scores) are derived from the token-level output distribution. We evaluate four models that expose token-probability APIs: GPT-4o-Mini, GPT-4o, Mistral-7B-Instruct-v0.1, and Llama-3.1-8b-Instruct. Each model also generates a verbalized confidence score per row (see Appendix \ref{appendix:prompts} for details). Final evaluations use the ground-truth labels to compute accuracy, AUC, and expected calibration error (ECE).

Beyond the designated “label” column for each dataset, we also treat other features as additional prediction targets, expanding the number of tasks substantially. In each case, one feature is treated as the prediction target while the others serve as inputs. For continuous features, we define binary labels relative to the median to standardize outputs, while for categorical features, we predict whether the value equals the mode. Features with >70\% missing values, or categorical features where >99\% of rows equal the mode or <10\% equal the mode, are excluded.  We sample 10 features per dataset to construct auxiliary prediction tasks. For example, given features A, B, C, and an outcome D, we remove D and create three tasks, (A,B→C), (B,C→A), and (A,C→B). We then compute zero-shot predictions on each task and average the AUCs, yielding 285 additional proxy evaluations.

\subsection{Predicting task-level performance}

We define and empirically evaluate metrics for predicting LLMs' zero-shot performance over domains. Many of these are intuitive extensions of individual-level uncertainty quantification strategies to the task level, and part of our goal is to give practitioners guidance about which extensions perform well empirically and which do not. We group our strategies into several broad categories.

\paragraph{Task-level confidence elicitation:} Perhaps the simplest strategy to predict LLMs' performance at a new task is to ask the LLM itself whether it will perform well, analogous to verbalized confidence strategies at the individual level \cite{tian2023just}. We provide the LLM with a text description of the dataset and its target variable (see Appendix \ref{appendix:prompts} for the exact prompt). We assess several strategies that prompt the LLM to output assessments of its own expected performance, given that LLMs are sensitive to the manner in which information is elicited. \textbf{Direct AUC prediction} asks the LLM to output a prediction of its own AUC at the task. \textbf{Integer scoring} asks the LLM to rate its confidence at the task as a number between 1 (no confidence) and 5 (full confidence). Finally, \textbf{Decimal scoring} asks the LLM for a continuous rating between 0.0 (no confidence) and 1.0 (full confidence). 

\paragraph{Aggregating individual-level confidence: } We utilize LLM outputs for each row of a dataset, given a prediction task, to design proxies for task-level AUC. For each row, we obtain the risk score $\hat{p}_i$ and verbalized confidence score $c_i$. One natural strategy is to aggregate these individual-level measures of uncertainty to the task level, reasoning that LLMs will perform well on tasks where they are confident in many individual examples. We evaluate four metrics as proxies for task-level performance. First, \textbf{average confidence}, defined for task $j$ as $\frac{1}{n_j}\sum_{i = 1}^{n_j} c_i$ (where $n_j$ is the number of samples for task $j$). Second, \textbf{average Maximum Class Probability (MCP)}, defined as $\frac{1}{n_j}\sum_{i = 1}^{n_j} \max\{\hat{p_i}, 1- \hat{p_i}\}$. This measures how close predictions are to 0 or 1, which is a proxy for confidence. Finally, we include two additional metrics, \textbf{standard deviation of confidence} and \textbf{standard deviation of risk scores}, the empirical standard deviations of the sets $\{c_i\}$ and $\{\hat{p}_i\}$, respectively. These are motivated by the anecdotal observation that one common failure mode LLMs encounter is outputting (near) identical responses for every row. One potential proxy to account for this is simply whether the LLM makes a wide range of predictions.

\paragraph{Masking:} Finally, we might think that an LLM will output high-quality predictions of a label $y$ if it performs well at other predictive tasks on the same dataset: predicting each feature $x^i$ from the other features $x^{-i}$. This procedure is motivated by the hypothesis that strong performance on these proxy tasks signals broader task-relevant understanding by the LLM. We collect risk scores from a sample of such masked prediction tasks for each dataset. The \textbf{masking} strategy takes the average of the AUCs in these simulated tasks as a proxy for the AUC from predicting the true label $y$.

\section{Results}

Our analysis is structured as follows. We begin by examining the zero-shot classification performance of the LLMs on our curated datasets, with a focus on the quality of individual-level predictions. We then broaden the scope to analyze the predictability of aggregate-level LLM performance across datasets. Results are shown for a spread of tested LLMs, with the full scope of results added to the Appendix.

\subsection{Overall Trends in Performance.}

\begin{figure}[ht]
\begin{subfigure}{.5\linewidth}
  \centering
  \includegraphics[width=\linewidth]{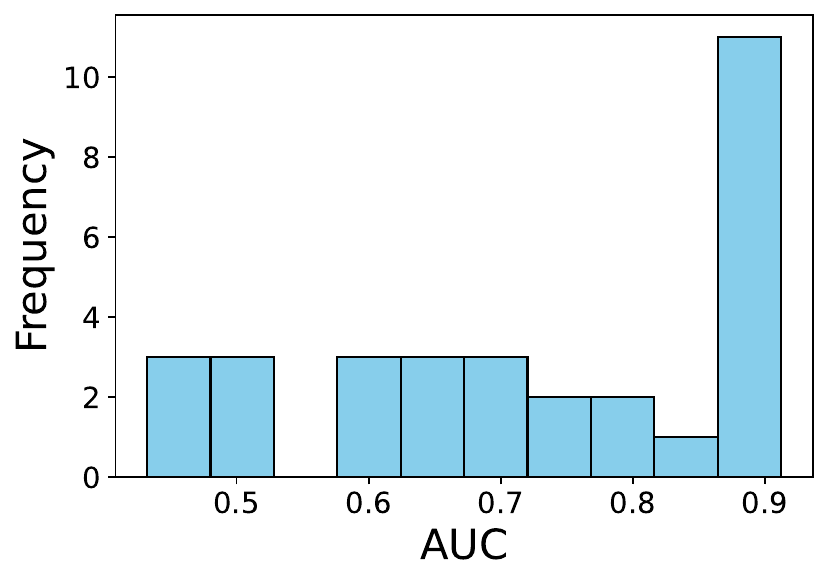}
  \caption{}
  \label{fig:sfig1-1}
\end{subfigure}%
\begin{subfigure}{.5\linewidth}
  \centering
  \includegraphics[width=\linewidth]{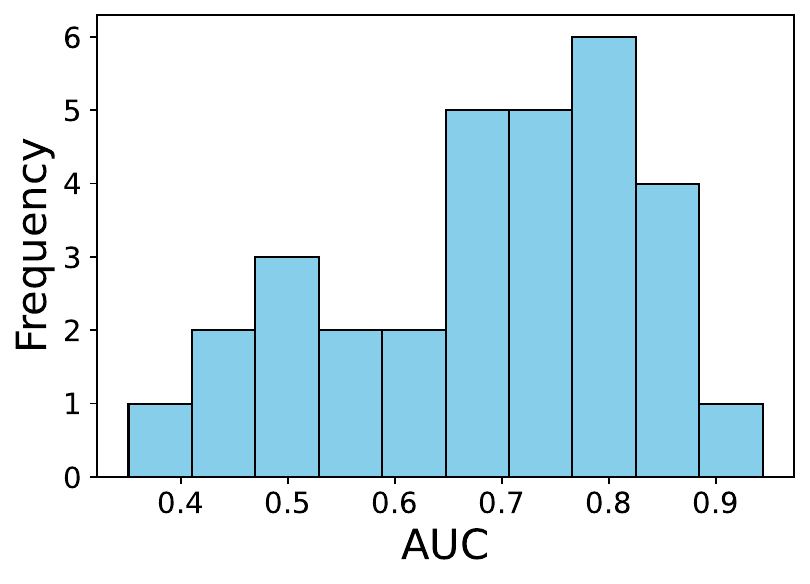}
  \caption{}
  \label{fig:sfig1-2}
\end{subfigure}

\begin{subfigure}{.5\linewidth}
  \centering
  \includegraphics[width=\linewidth]{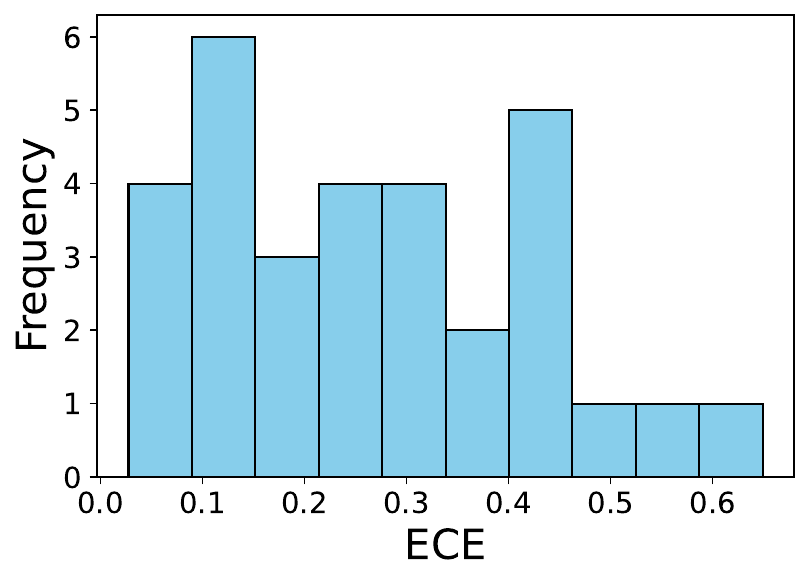}
  \caption{}
  \label{fig:sfig1-3}
\end{subfigure}%
\begin{subfigure}{.5\linewidth}
  \centering
  \includegraphics[width=\linewidth]{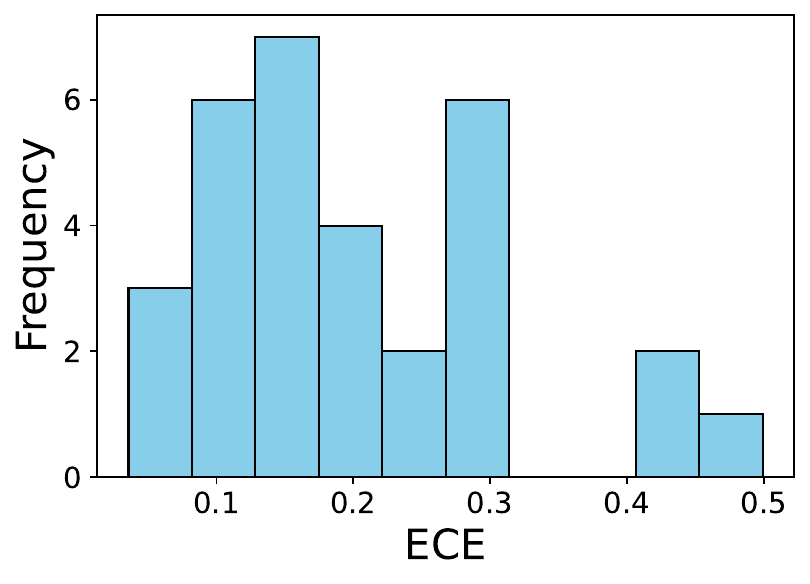}
  \caption{}
  \label{fig:sfig1-4}
\end{subfigure}

\caption{Histograms of AUC and ECE over all datasets, for GPT-4o-mini (a,c) and Llama-3.1-8b-Instruct (b,d).}
\label{fig:fig1}
\end{figure}

\textbf{\textit{LLMs have significant spread in their prediction capabilities, both across datasets and across prediction tasks from the same dataset}}. In Figures \ref{fig:sfig1-1} and \ref{fig:sfig1-2}, we observe that both GPT-4o-mini and Llama-3.1-8b-Instruct have nontrivial zero-shot predictive capabilities, with a median AUC of 0.7232 for the GPT-4o-mini and 0.7080 for the Llama model. The range is wide, with AUCs above 0.9 for some tasks, but at near-random (or worse than random) levels for others. This confirms that practitioners must take steps to assess the appropriateness of LLM zero-shot inference for a given task. See Appendix \ref{appendix:metrics} for a full set of AUC and ECE scores over all datasets and LLMs.

Within individual datasets, the quality of LLM predictions can vary substantially when using different columns as outcome variables (i.e., different prediction tasks). In Figure~\ref{fig:fig9}, we plot the distribution of AUC scores across columns within each dataset for GPT-4o-Mini (see Appendix \ref{appendix:rawauc} for similar plots for the other LLMs.). These results show that intra-dataset variation is often considerable: many datasets contain prediction tasks with AUC scores below 0.5 as well as tasks with scores above 0.9. To quantify this result, we compute an intra-class correlation coefficient, defined as the ratio of the variance in AUC within datasets vs overall, measuring the fraction of variance at the dataset level. We find that only 19\% of the variance is explained by the dataset for GPT-4o-mini (for Llama-3.1-8b-Instruct, 12.80\%), with 81\% persisting within datasets. Perhaps surprisingly, this indicates that checking the performance of an LLM on some tasks in a given domain offers practitioners little confidence that it will perform well in unseen tasks from the same domain. 

For deeper analysis, we also examine LLM performance relative to the \textit{best achievable}, as some within-dataset variation may stem from inherent differences in column difficulty, independent of model skill. To test for dataset-level variation in relative LLM skill, we compute the ratio between the LLM’s AUC and that of an XGBoost model trained on labeled data~\cite{chen2016xgboost}, as a proxy for optimal performance. This normalized metric is more concentrated within datasets than AUC (Appendix \ref{appendix:normauc}), with the intraclass correlation increasing to 53.02\% for GPT-4o-mini (47.68\% for Llama-3.1-8b-Instruct), indicating more variation is explained at the dataset level. From the perspective of scientific understanding of LLMs' capabilities, this suggests there are meaningful differences in skill across domains after accounting for the inherent difficulty of a task (although practitioners may more heavily weigh absolute performance, where our earlier results show high within-dataset variation). Interestingly, GPT-4o-Mini's and Llama-3.1-8b-Instruct's AUC scores correlate strongly across tasks ($R^2 = 0.497$, Figure \ref{fig:fig4}), suggesting certain tasks are more amenable to LLM-based inference than others. This correlation is stronger than either model’s correlation with XGBoost performance (Figure \ref{fig:fig5}), implying that shared LLM performance factors are not reducible to the difficulty of the base  task. Analogous figures for GPT-4o and Mistral-7b-Instruct-v0.1 are shown in the Appendix~\ref{appendix:agreement}.

\begin{figure}[!ht]

    \centering
    \includegraphics[width=\linewidth]{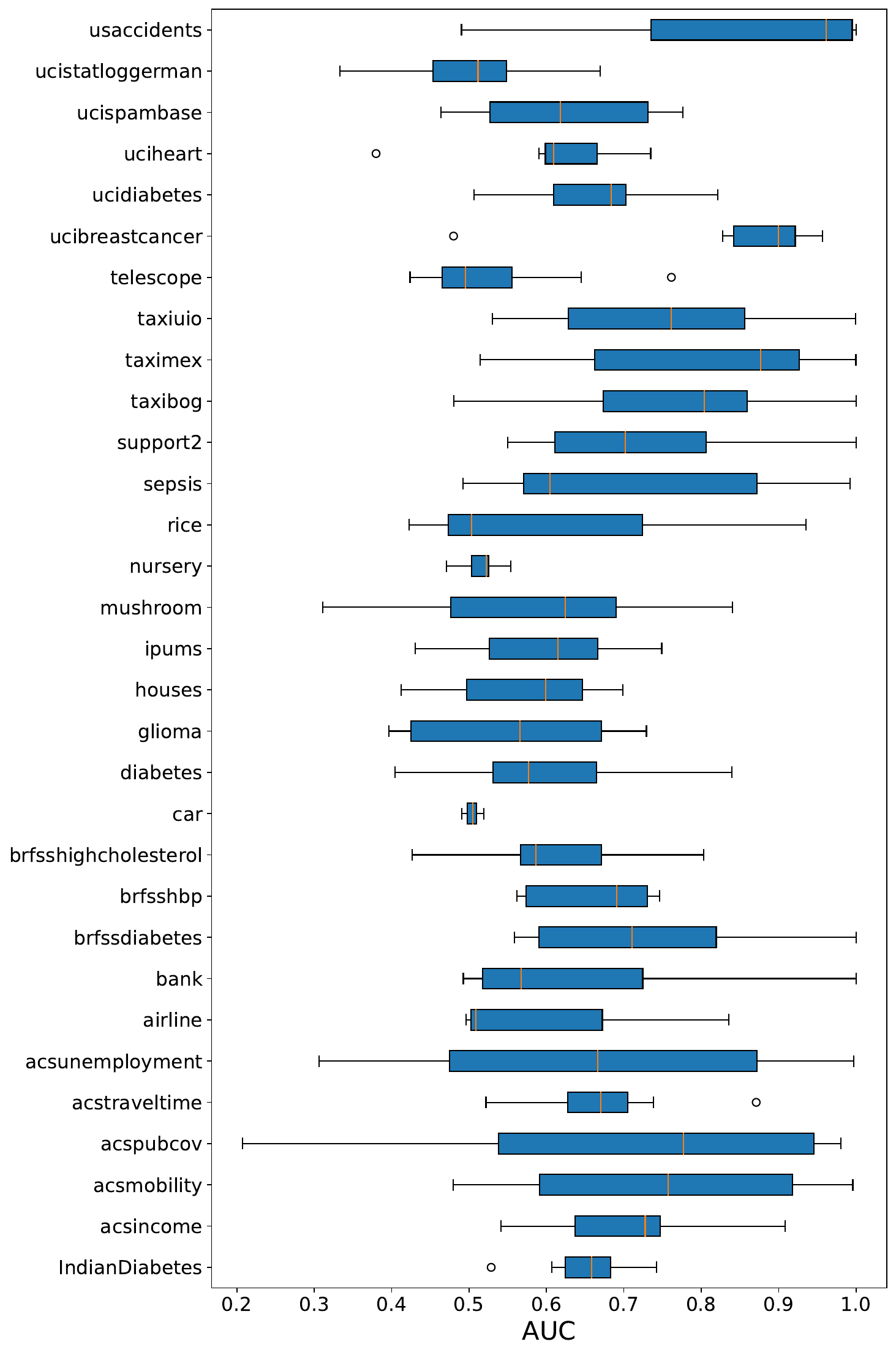}

\caption{Box plots of AUC scores over masked-out columns in the Masking experiment, for all datasets. Results shown for GPT-4o-mini.}

\label{fig:fig9}

\end{figure}

\begin{figure}[!ht]
  \includegraphics[width=0.5\textwidth]{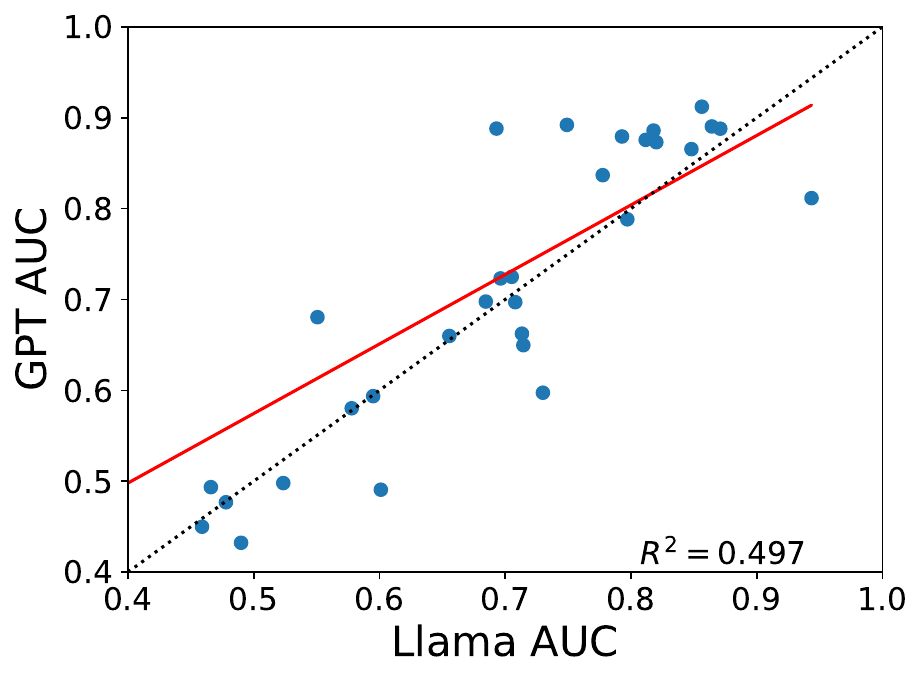}

\caption{Plot of AUC scores for each of the datasets, for both Llama-3.1-8b-Instruct and GPT-4o-mini. Best-fit line with $R^2$ value plotted in red.}

\label{fig:fig4}
\end{figure}

\begin{figure}[!t]

\begin{subfigure}{.5\linewidth}
    \centering
    \includegraphics[width=\linewidth]{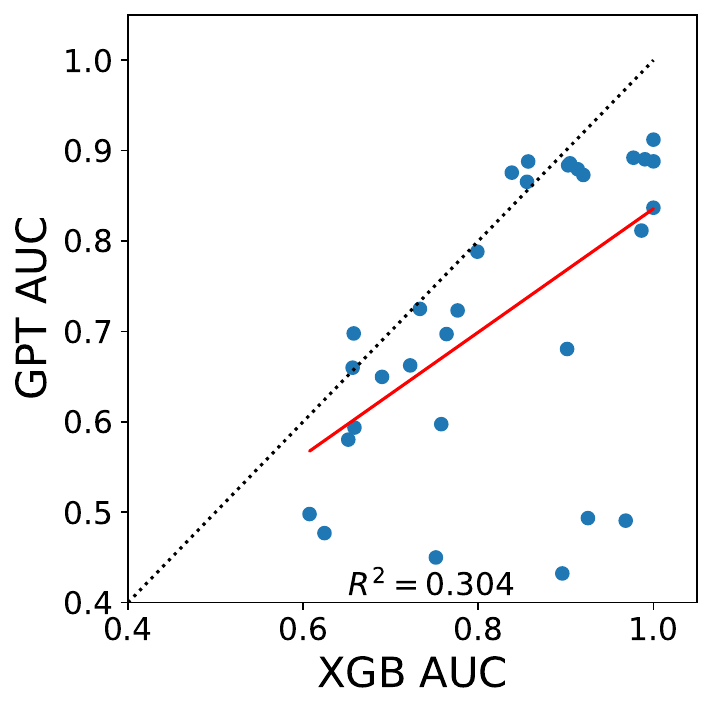}
    \caption{}
    \label{fig:sfig5-1}
\end{subfigure}%
\begin{subfigure}{.5\linewidth}
    \centering
    \includegraphics[width=\linewidth]{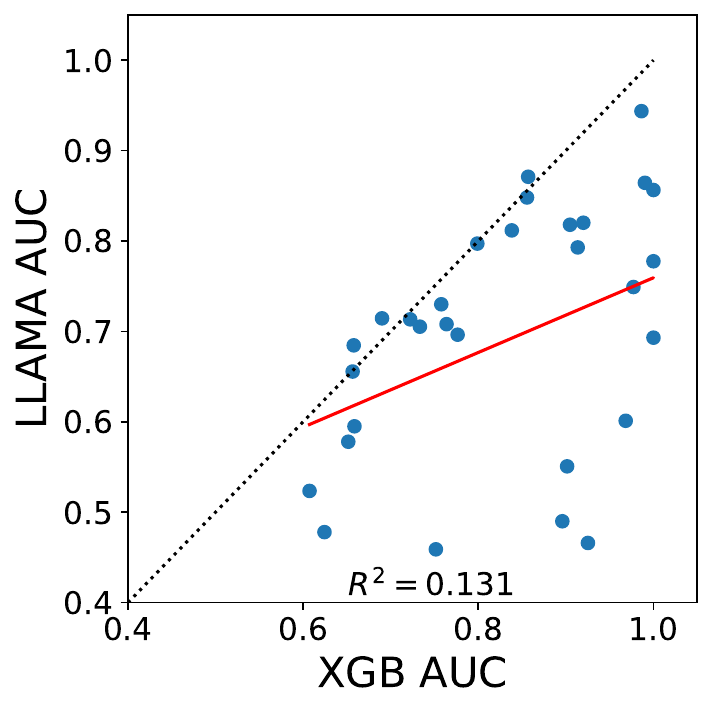}
    \caption{}
    \label{fig:sfig5-2}
\end{subfigure}

\caption{Correlation between AUC scores of GPT-4o-mini (a) and Llama (b) over prediction tasks on each dataset, along with the AUCs achieved by training an XGBoost classification model on a subset of the training set, and evaluating on a disjoint validation set. 
}

\label{fig:fig5}

\end{figure}

\subsection{Individual-Level Results.}

\begin{figure}[!ht]
\begin{subfigure}{.5\linewidth}
    \centering
    \includegraphics[width=\linewidth]{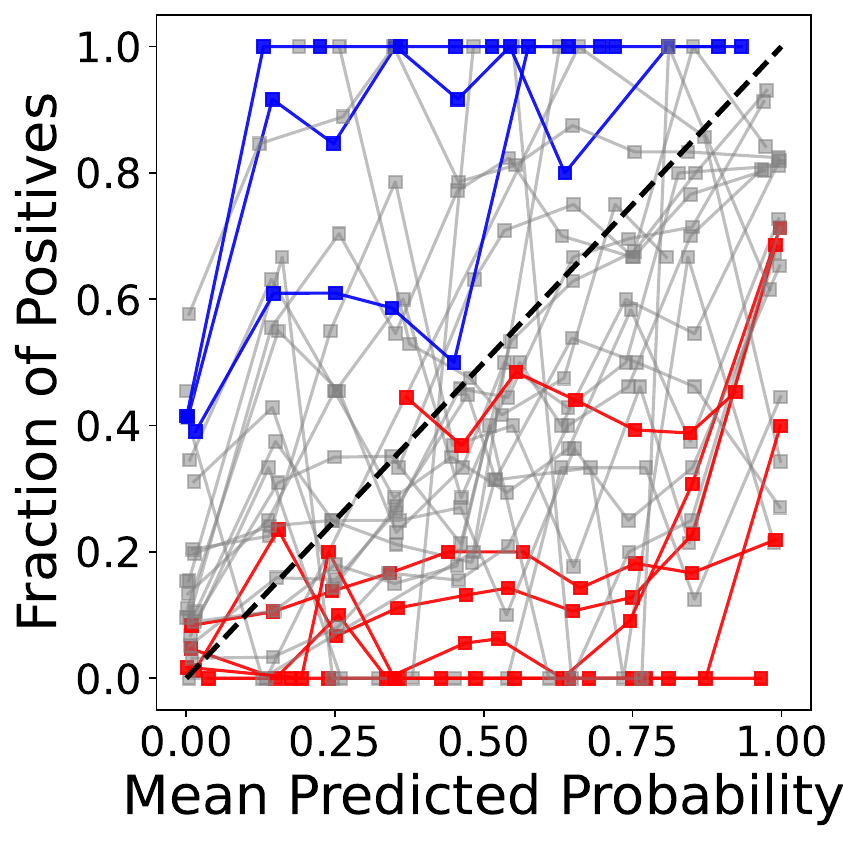}
    \caption{}
    \label{fig:sfig2-1}
\end{subfigure}%
\begin{subfigure}{.5\linewidth}
    \centering
    \includegraphics[width=\linewidth]{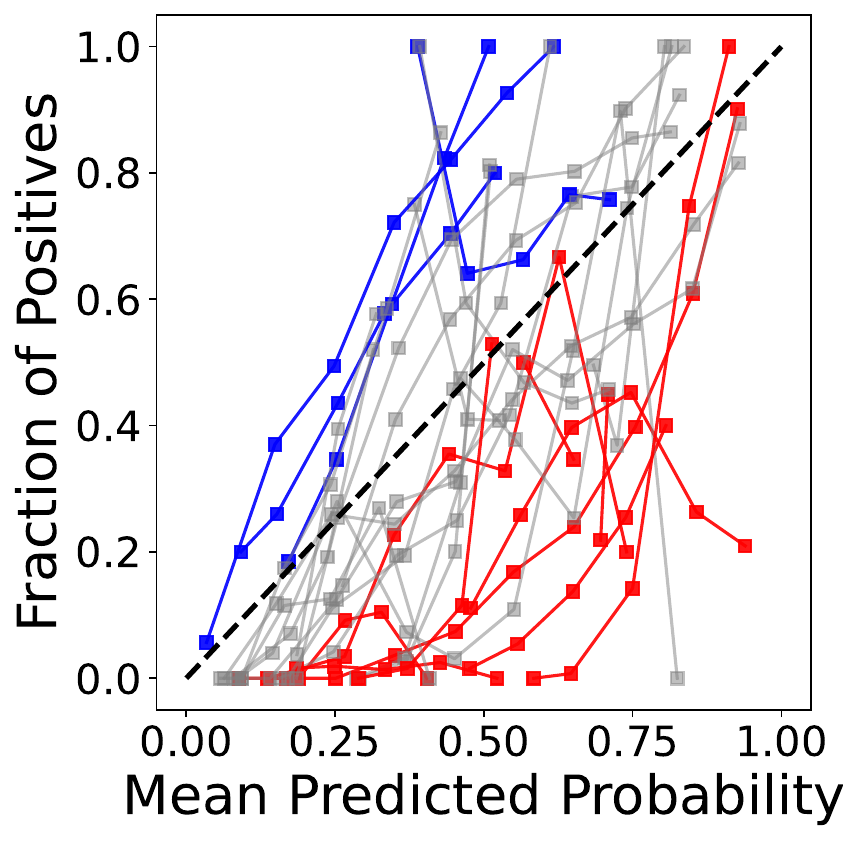}
    \caption{}
    \label{fig:sfig2-2}
\end{subfigure}

\caption{Calibration curves for GPT-4o-mini (a) and Llama-3.1-8b-Instruct (b) across 31 datasets. Each curve corresponds to a prediction task. Curves crossing the identity line are shown in grey; those consistently above or below are blue and red, respectively. Concretely, all curves that a) are on average 0.2 above the identity line and b) have no points more than .1 below the identity line are colored in blue; curves on average .2 below the identity line and with no points more than .1 above are colored in red.}
\label{fig:fig2}
\end{figure}

\textbf{\textit{Elicited risk scores from LLMs are poorly calibrated, but are often useful for abstention tasks.}} Figure \ref{fig:sfig1-3} and \ref{fig:sfig1-4} show median ECEs around 0.2 for GPT-4o-mini and Llama (0.2426 and 0.1722, respectively), with GPT-4o-mini exhibiting greater variability. This corroborates previous findings of poor LLM calibration in US census tasks \cite{cruz2024_riskscores} in a larger set of probabilistic prediction tasks. While prior work reports overconfident, inverted-sigmoid calibration curves from instruction-tuned models, we observe curves (see Figure \ref{fig:fig2}, Appendix~\ref{appendix:fail}) that often remain entirely above or below the identity line, indicating predictions are consistently too high or too low. This suggests that LLMs often misjudge the absolute scale of their risk scores, even when preserving relative ranking accuracy (as reflected by high AUCs). Our findings thus contradict prior notions of overconfidence: instead, LLMs ostensibly have difficulty scaling their predictions to fit the marginal distribution of the label, even while correctly identifying which features correlate well with the label, which was a previously unknown phenomenon.

\begin{figure}[!ht]
\begin{subfigure}{.5\linewidth}
    \centering
    \includegraphics[width=\linewidth]{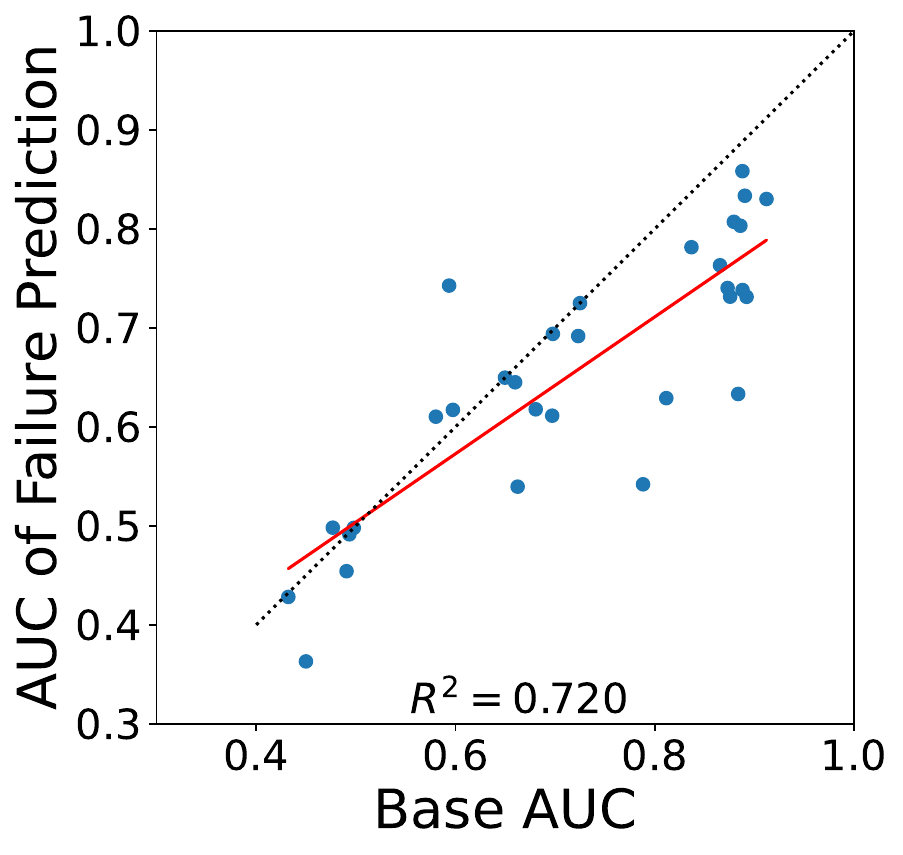}
    \caption{}
    \label{fig:sfig3-1}
\end{subfigure}%
\begin{subfigure}{.5\linewidth}
    \centering
    \includegraphics[width=\linewidth]{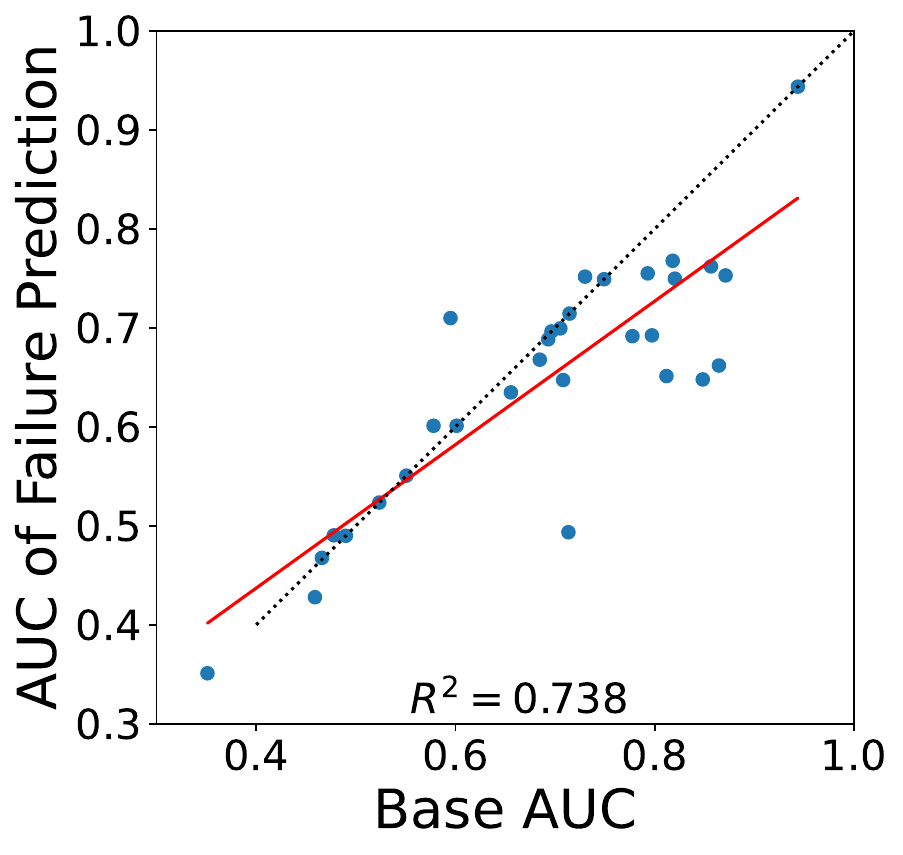}
    \caption{}
    \label{fig:sfig3-2}
\end{subfigure}

\caption{
Correlation between AUC scores of failure prediction and predicting the outcome variable for all datasets, for GPT-4o-mini (a) and Llama (b).
}
\label{fig:fig3}
\end{figure}

Despite poor numerical calibration, predictions closer to 0 or 1 (higher confidence) tend to be more accurate. We simulate abstention systems with LLM outputs by examining the degree to which MCP, a proxy of confidence in the predicted label, predicts individual-level accuracy, a task referred to by ~\cite{xiong2023can} as \textit{failure prediction}. We observe that LLM outputs are nontrivially successful at failure prediction for many tasks (see Figure \ref{fig:fig3}, Appendix~\ref{appendix:failpred}). On the high end, we find AUCs for failure prediction of nearly 0.9, although performance varies across tasks (ranging from around 0.4 to 0.9). Strikingly, this effect is stronger for tasks where the LLM already performs well: the AUC of the original prediction task is highly correlated with AUC of failure prediction, indicating that \textbf{\textit{when a model has a strong baseline ability, its confidence is better aligned with accuracy}}. Thus, risk scores—despite calibration issues—can potentially support abstention strategies on domains where LLM usage is well-motivated to begin with, as LLMs often distinguish effectively between more and less reliable predictions on those tasks.

\subsection{Task-Level Results.}

\begin{figure*}[!ht]
\begin{subfigure}{.25\linewidth}
  \centering
\includegraphics[width=\linewidth]{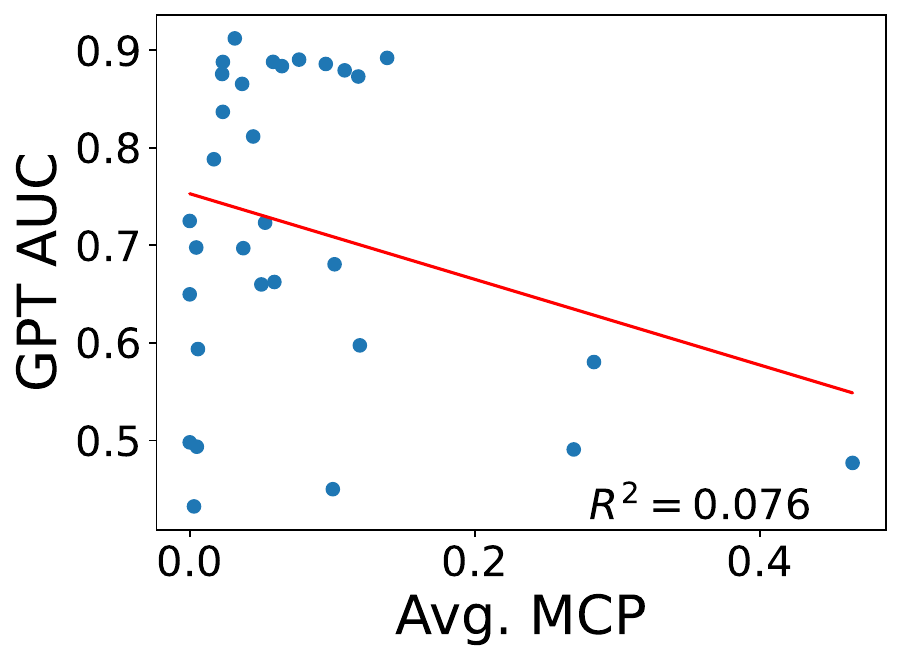}
  \caption{}
  \label{fig:sfig6-1}
\end{subfigure}%
\begin{subfigure}{.25\linewidth}
  \centering
\includegraphics[width=\linewidth]{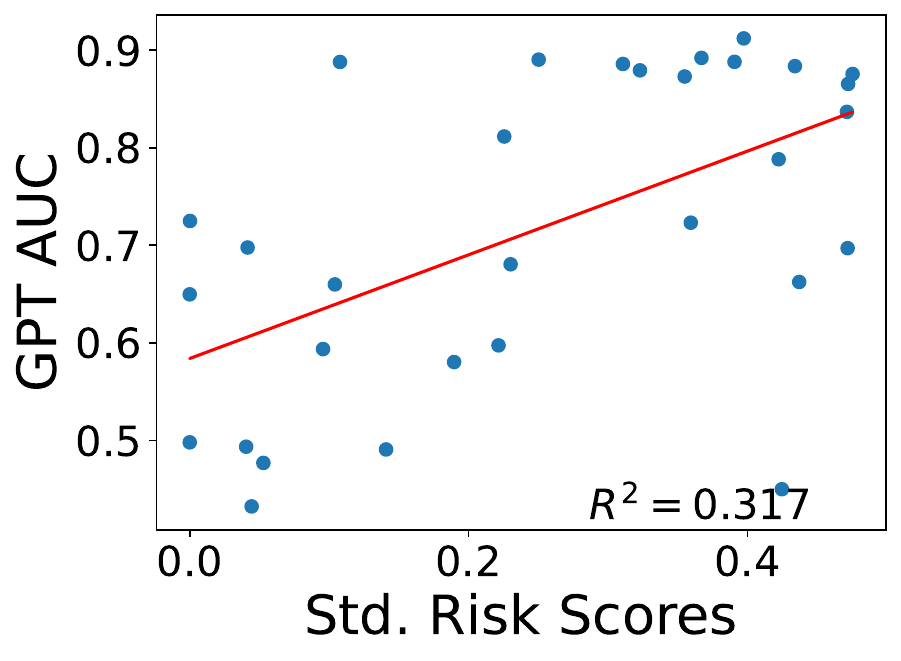}
  \caption{}
  \label{fig:sfig6-2}
\end{subfigure}%
\begin{subfigure}{.25\linewidth}%
  \centering
\includegraphics[width=\linewidth]{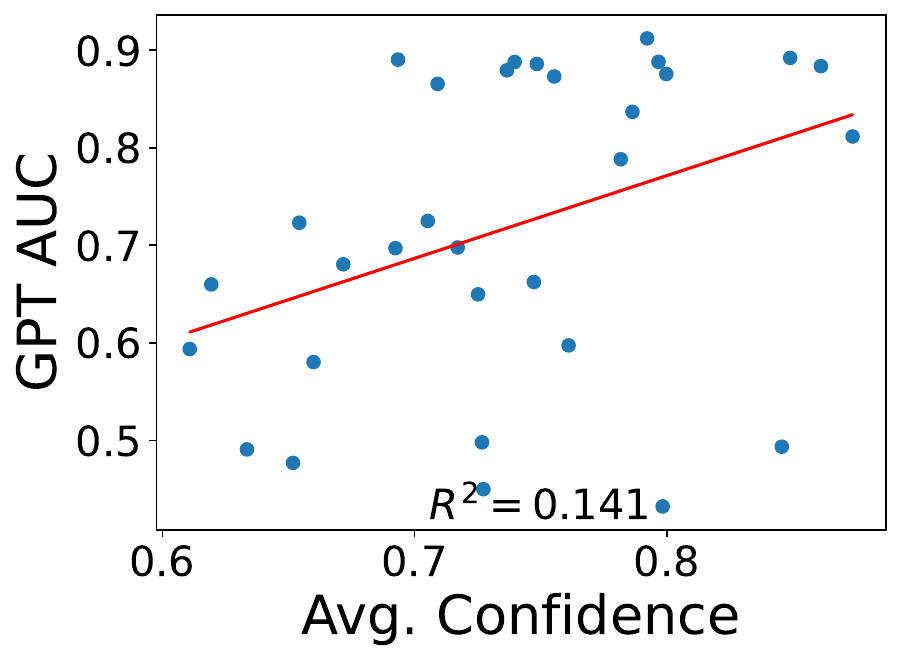}
  \caption{}
  \label{fig:sfig6-3}
\end{subfigure}%
\begin{subfigure}{.25\linewidth}
  \centering
\includegraphics[width=\linewidth]{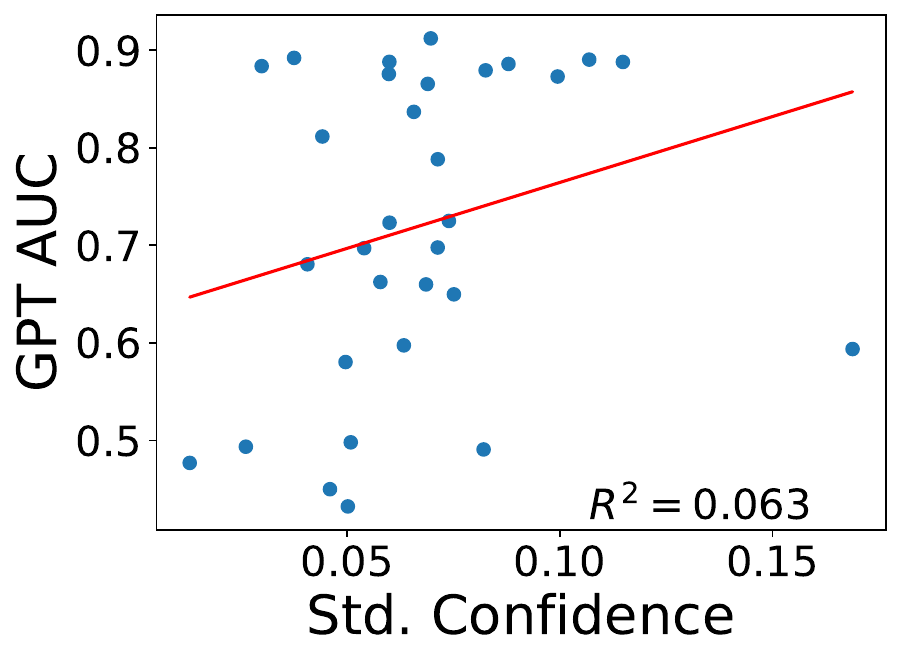}
  \caption{}
  \label{fig:sfig6-4}
\end{subfigure}

\begin{subfigure}{.25\linewidth}
  \centering
  \includegraphics[width=\linewidth]{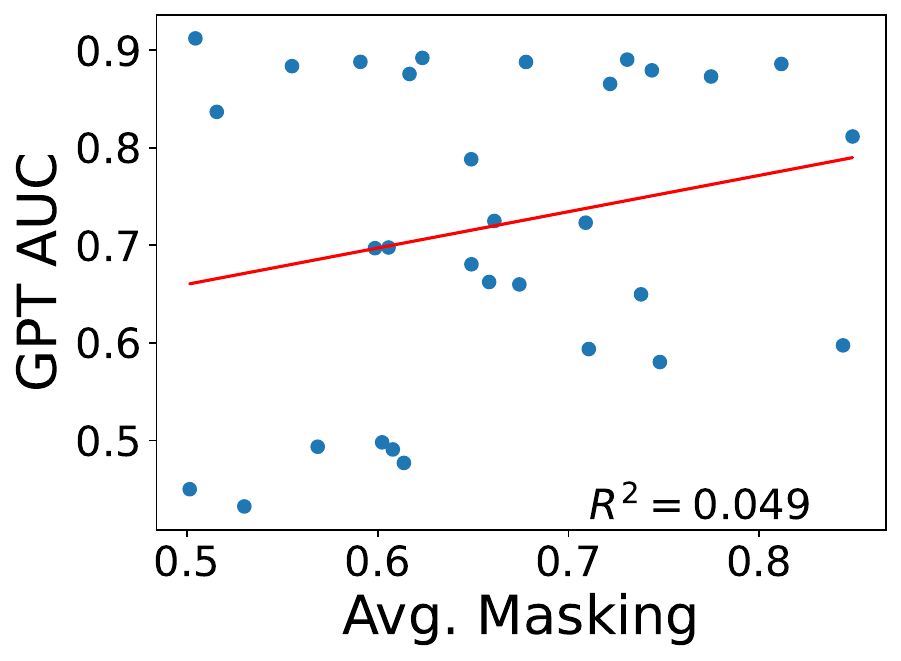}
  \caption{}
  \label{fig:sfig6-5}
\end{subfigure}%
\begin{subfigure}{.25\linewidth}
  \centering
  \includegraphics[width=\linewidth]{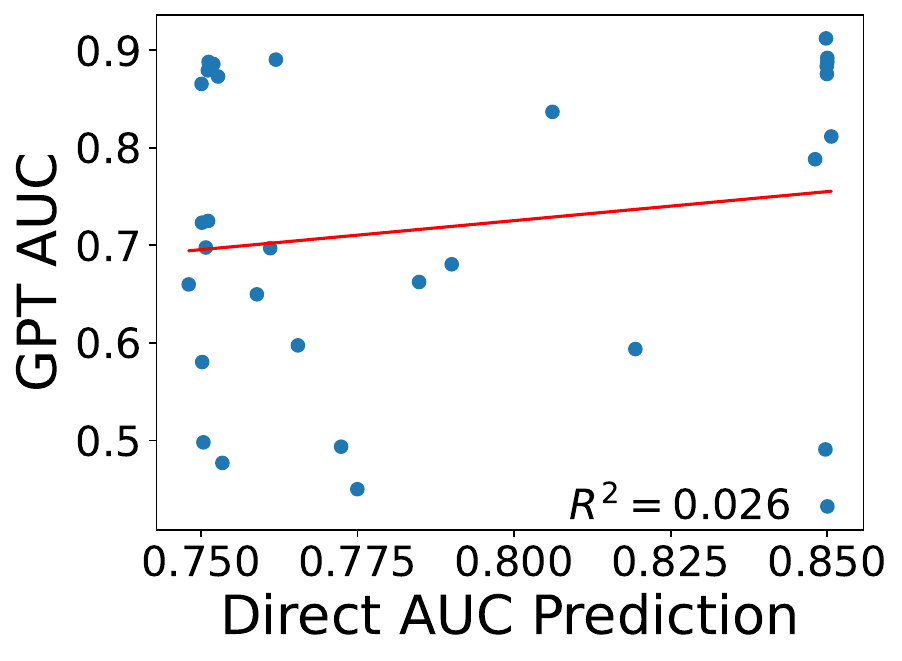}
  \caption{}
  \label{fig:sfig6-6}
\end{subfigure}%
\begin{subfigure}{.25\linewidth}
  \centering
  \includegraphics[width=\linewidth]{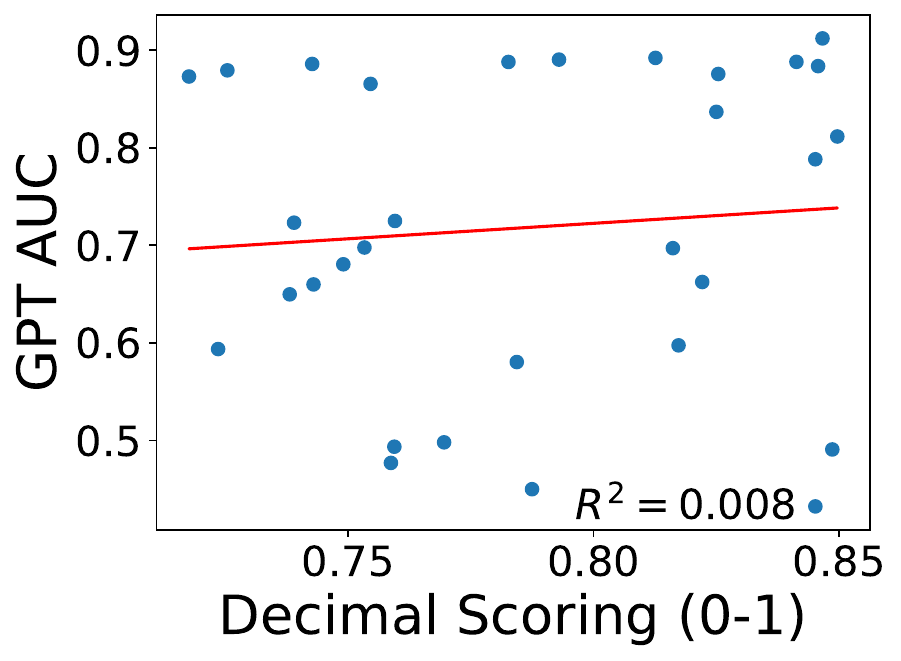}
  \caption{}
  \label{fig:sfig6-7}
\end{subfigure}%
\begin{subfigure}{.25\linewidth}
  \centering
  \includegraphics[width=\linewidth]{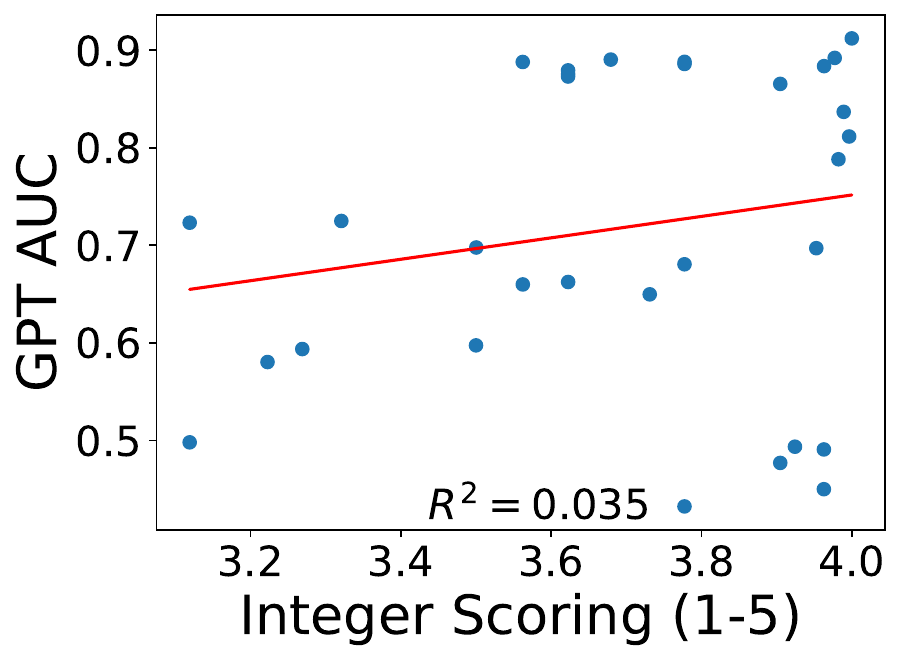}
  \caption{}
  \label{fig:sfig6-9}
\end{subfigure}
\caption{Correlation between aggregate metrics derived from our experiments on the unlabeled datasets and the AUC scores of GPT-4o-mini on each of the datasets, where each point represents one dataset. We plot the best-fit line with its corresponding $R^2$ value for each metric. See Appendix \ref{appendix:llamacor} for the same set of plots made for Llama.}
\label{fig:fig6}

\end{figure*}

\begin{figure*}[!ht]
\begin{subfigure}{.25\linewidth}
  \centering
\includegraphics[width=\linewidth]{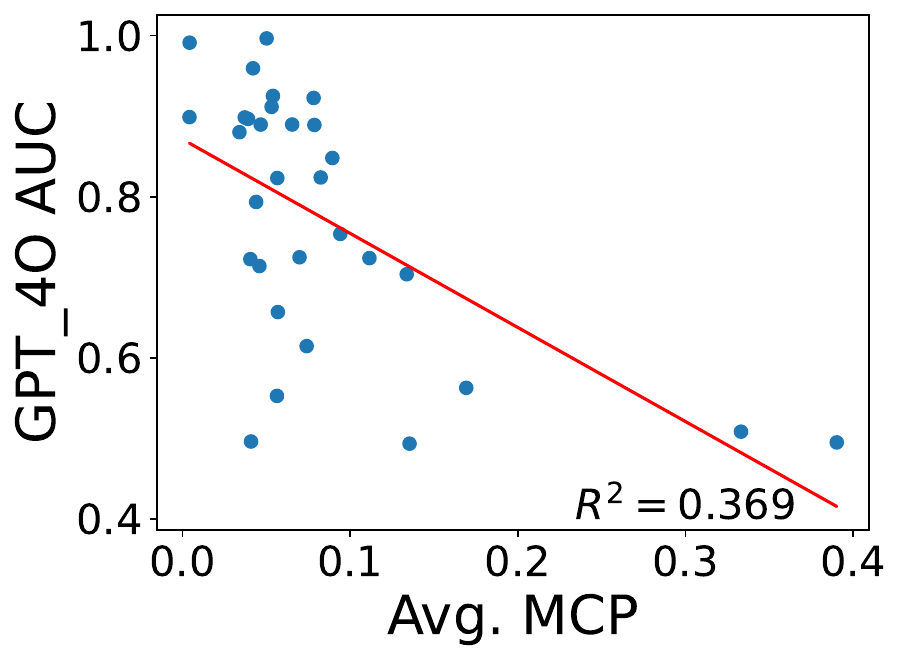}
  \caption{}
  \label{fig:sfig4o1-1}
\end{subfigure}%
\begin{subfigure}{.25\linewidth}
  \centering
\includegraphics[width=\linewidth]{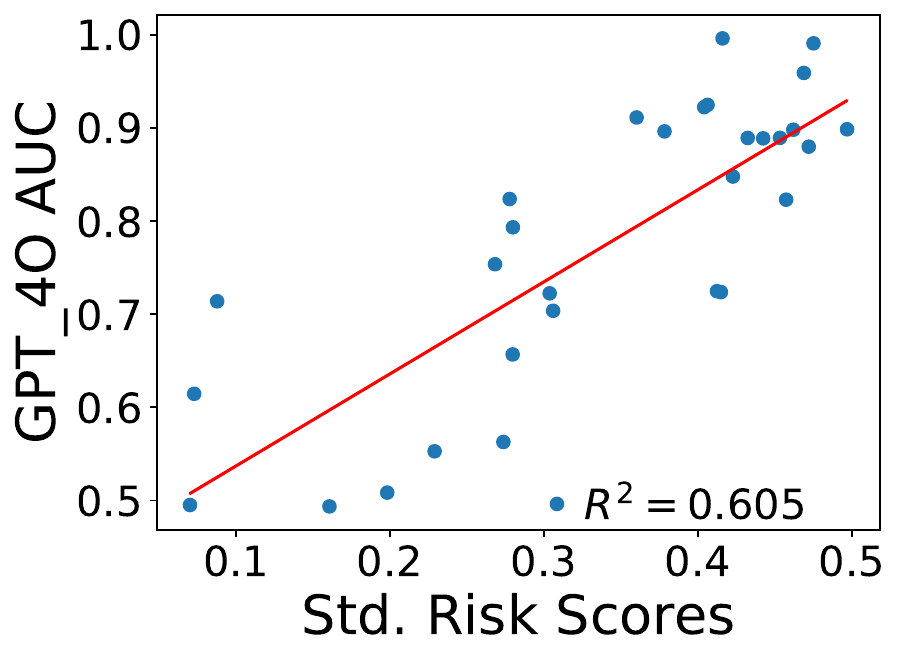}
  \caption{}
  \label{fig:sfig4o1-2}
\end{subfigure}%
\begin{subfigure}{.25\linewidth}%
  \centering
\includegraphics[width=\linewidth]{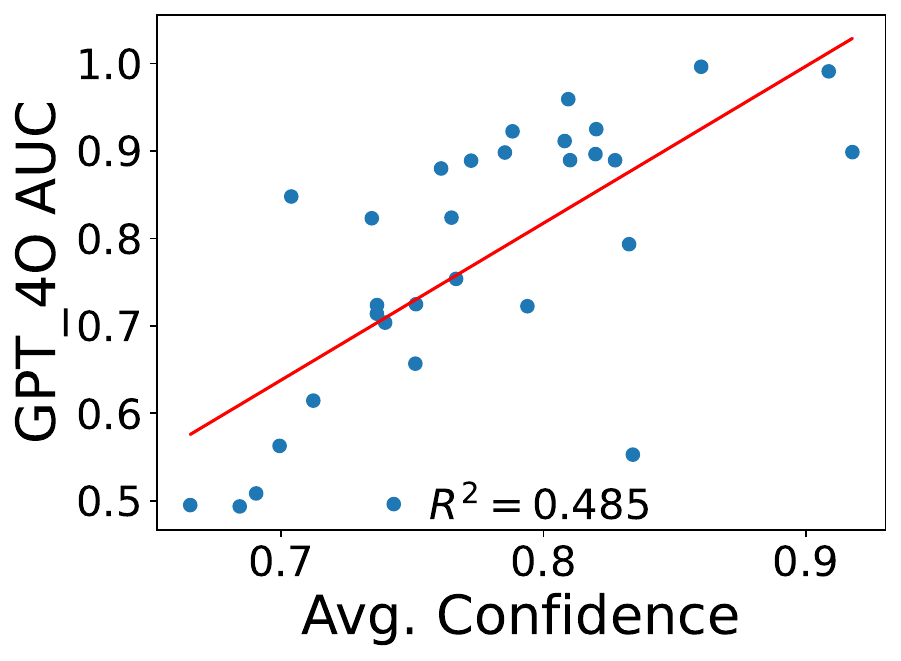}
  \caption{}
  \label{fig:sfig4o1-3}
\end{subfigure}%
\begin{subfigure}{.25\linewidth}
  \centering
\includegraphics[width=\linewidth]{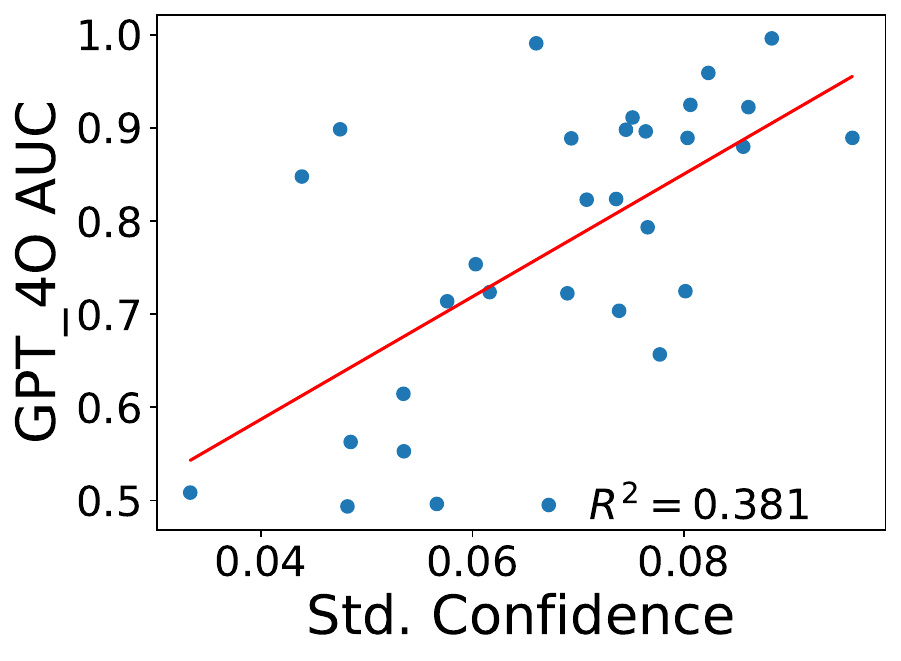}
  \caption{}
  \label{fig:sfig4o1-4}
\end{subfigure}

\begin{subfigure}{.25\linewidth}
  \centering
  \includegraphics[width=\linewidth]{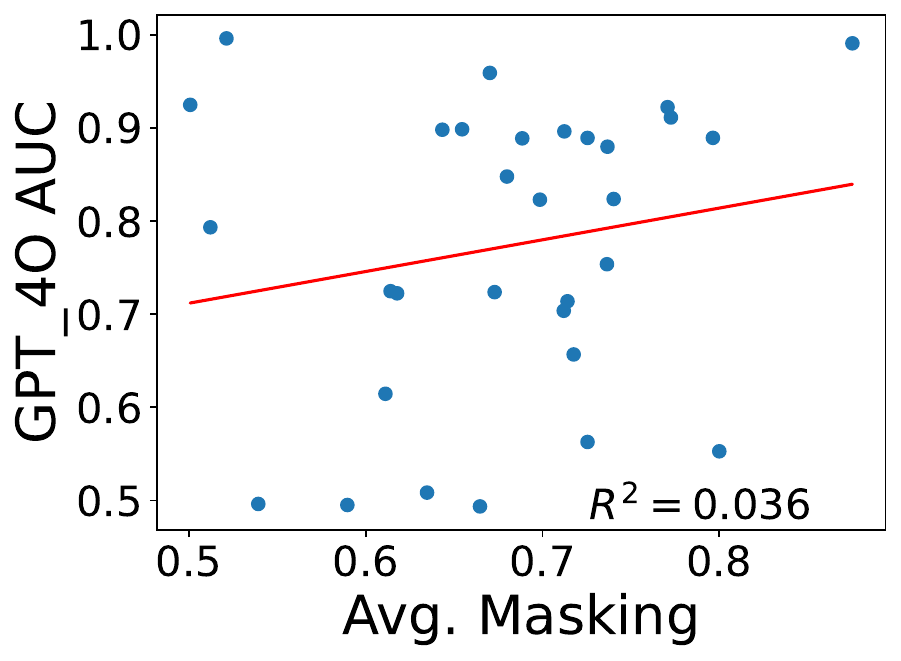}
  \caption{}
  \label{fig:sfig4o1-5}
\end{subfigure}%
\begin{subfigure}{.25\linewidth}
  \centering
  \includegraphics[width=\linewidth]{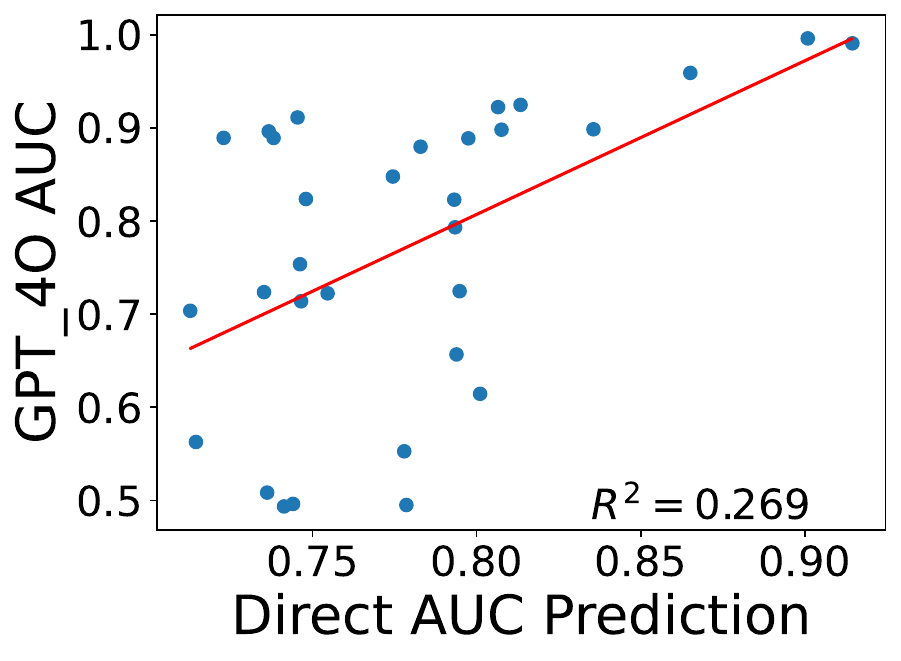}
  \caption{}
  \label{fig:sfig4o1-6}
\end{subfigure}%
\begin{subfigure}{.25\linewidth}
  \centering
  \includegraphics[width=\linewidth]{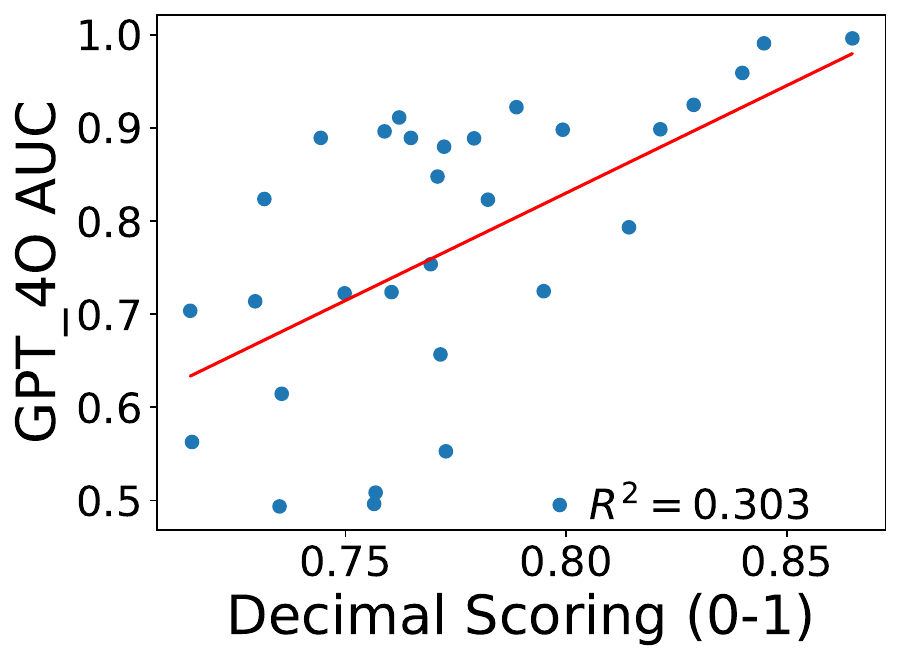}
  \caption{}
  \label{fig:sfig4o1-7}
\end{subfigure}%
\begin{subfigure}{.25\linewidth}
  \centering
  \includegraphics[width=\linewidth]{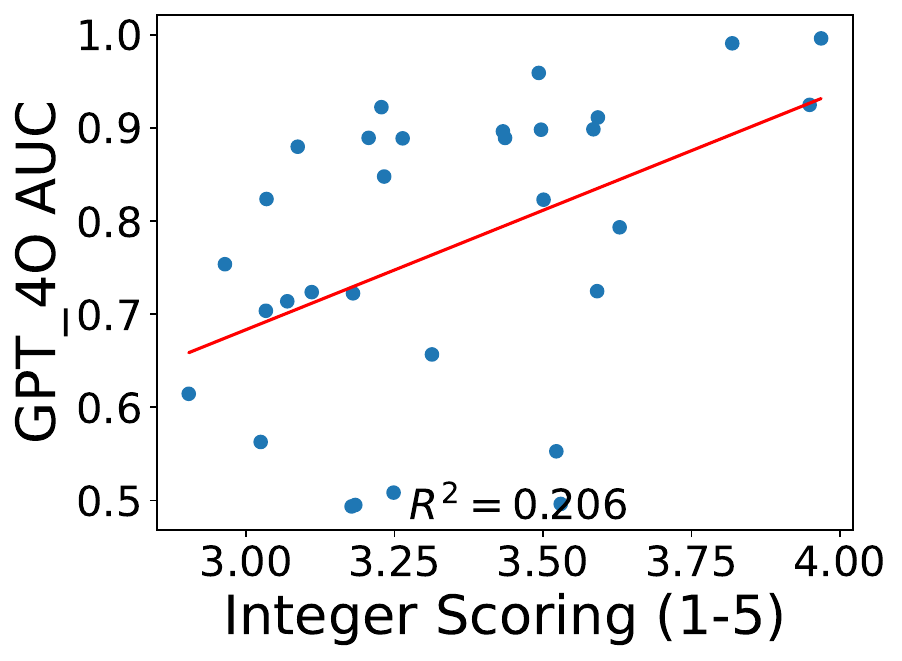}
  \caption{}
  \label{fig:sfig4o1-9}
\end{subfigure}
\caption{Correlation between aggregate metrics derived from our experiments on the unlabeled datasets and the AUC scores of GPT-4o on each of the datasets, where each point represents one dataset. We plot the best-fit line with its corresponding $R^2$ value for each metric.}
\label{fig:fig4o1}

\end{figure*}

\textbf{\textit{LLMs vary widely in their ability to anticipate their performance on new tasks, correlating with their baseline strength.}} We observe that newer and more capable LLMs demonstrate the strongest general performance on tabular data. GPT-4o, the newest and largest LLM we study, achieves the highest median AUC across tasks (0.82) and has a nontrivial correlation between metrics that do not leverage any unlabeled data (the "direct AUC prediction," "decimal scoring," and "integer scoring" metrics) and its actual performance (Figure~\ref{fig:fig4o1}). There is also a strong correlation between GPT-4o's task-level performance and its average confidence on unlabeled data points for that task. However, these metrics display much weaker correlations with task-level performance for the other LLMs studied. These models have both weaker predictive performance and little correlation between their self-predicted performance and actual performance. We note that aggregate LLM performance is exactly in line with the strength of the LLM (i.e., parameter count): in order from lowest to highest average AUC across the 31 original tasks, we have Mistral-7b-Instruct-v0.1 (AUC: 0.66), Llama-3.1-8b-Instruct (AUC: 0.69), GPT-4o-mini (AUC: 0.72), and GPT-4o (AUC: 0.77). These findings suggest that an LLM’s baseline capability correlates with its ability to anticipate its own performance, perhaps due to its inherent grasp of the prediction task.

The limitations of smaller models make this contrast even clearer. For smaller models (see Figure~\ref{fig:fig6}, Appendix~\ref{appendix:llamacor}), most evaluated metrics show little correlation with AUC. In particular, methods that do not use any unlabeled data are entirely uncorrelated with performance. Even for GPT-4o, these  metrics are weaker predictors of AUC than metrics that exploit unlabeled data, such as the standard deviation of risk scores (Figure~\ref{fig:fig4o1}). This suggests that confidence signals derived from unlabeled data are consistently more informative than self-estimates that rely solely on task descriptions, even while the strength of all metrics roughly scales with the complexity of the LLM.

 Surprisingly, the masking strategy (proxying the LLMs' performance at predicting a label by its performance at predicting features)  poorly predicts downstream AUC for all tested LLMs. Although one might expect that an LLM’s performance on masked columns would reflect its overall predictive capacity on a dataset, this assumption does not hold empirically. As shown in Figure~\ref{fig:fig9}, AUC scores vary widely across tasks within the same dataset, limiting the utility of dataset-averaged metrics. In other words, the variance in predictive quality across outcome variables makes it difficult for a dataset-level average to be a strong indicator of performance on any specific task within the dataset. As this appears to be a trend across many datasets, the results point to LLMs’ capabilities as the mechanism: we find no evidence to support the expectation that LLMs are reliably good within some domains and consistently bad at others. 
 
\textbf{\textit{Information describing the spread of risk scores provides particularly strong signals for downstream performance.}} As shown in Figures~\ref{fig:sfig6-2},~\ref{fig:sfig4o1-2},~\ref{fig:sfigmistral1-2},~\ref{fig:llamacor-2}, the standard deviation of risk scores correlates positively with downstream AUC. For all tested models, the $R^2$ of this relationship is the highest among all metrics evaluated (e.g., $R^2 = 0.605$ and $0.270$ for GPT-4o and Llama-3.1-8b-Instruct, respectively). A higher variance in risk scores may reflect greater separation between  classes in a model's predictions, suggesting that some failure modes are distinguished by the model giving similar predictions for most rows. Importantly, there are significant outliers from this relationship, indicating that large variance in risk scores is not a guarantee of good performance on a task. Nevertheless, since this metric exhibits by far the largest correlation with predictive performance, we conduct a deeper dive by querying the distribution of model predictions for the 285 additional masked-column prediction tasks in addition to the 31 original tasks of predicting the designated label for each dataset. This gives us a significantly larger task-level sample size for more detailed analysis.

\begin{figure}[!htbp]

\begin{subfigure}{.5\linewidth}
    \centering
    \includegraphics[width=\linewidth]{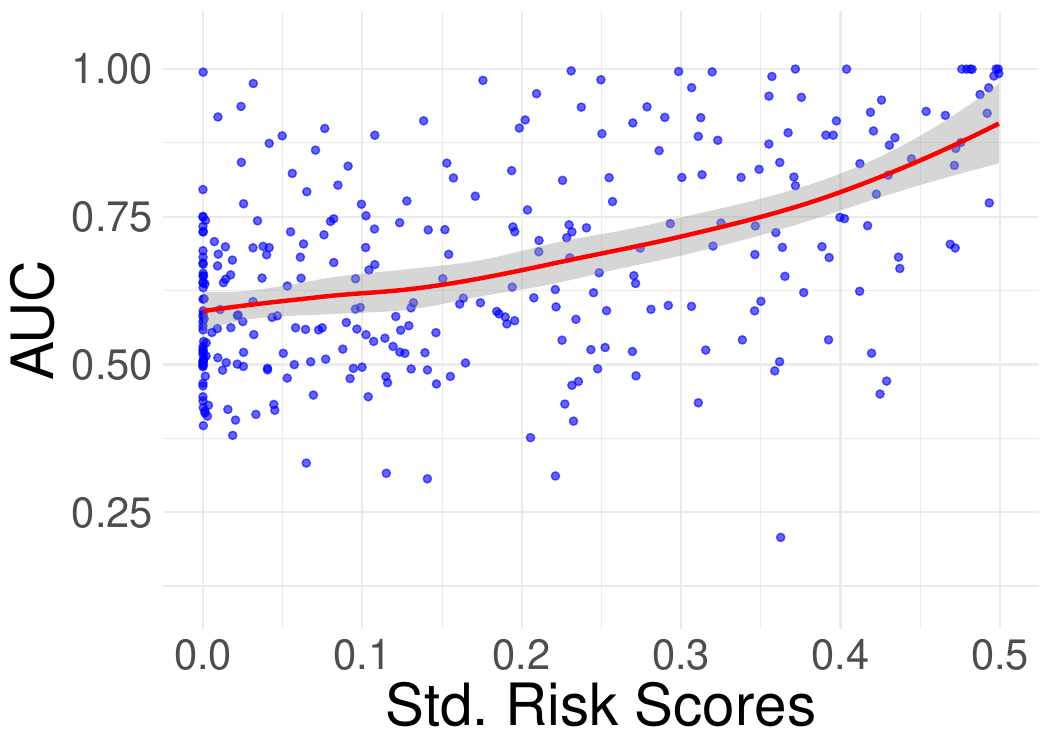}
    \caption{}
    \label{fig:sfig7-1}
\end{subfigure}%
\begin{subfigure}{.5\linewidth}
    \centering
    \includegraphics[width=\linewidth]{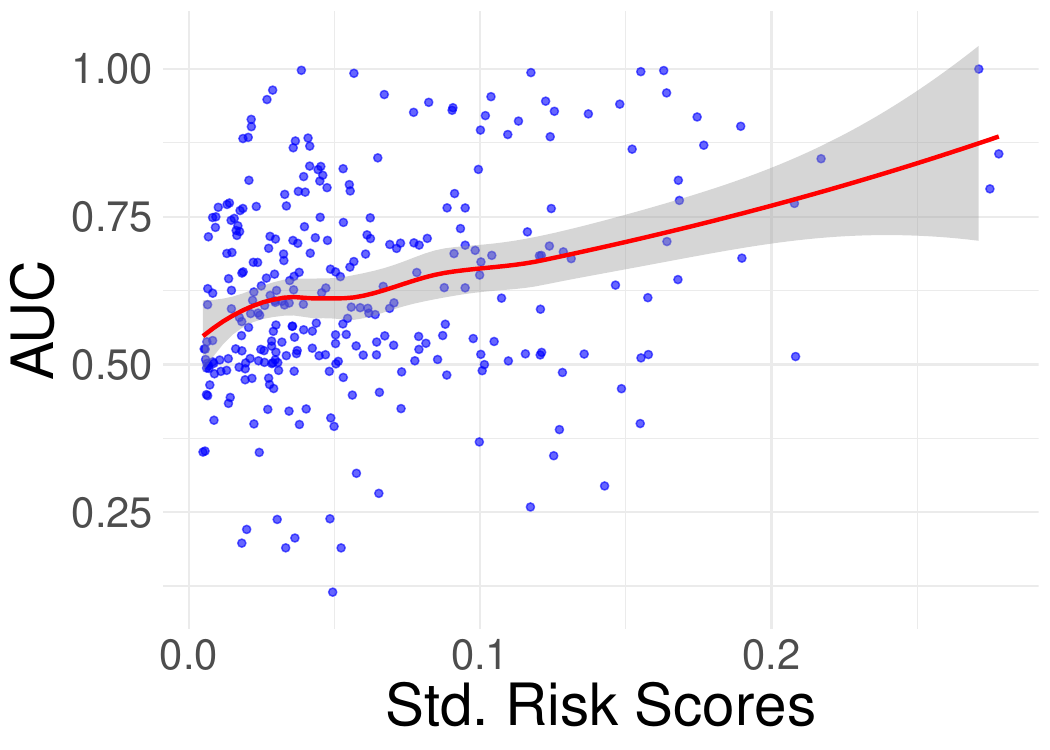}
    \caption{}
    \label{fig:sfig7-2}
\end{subfigure}

\begin{subfigure}{.5\linewidth}
    \centering
    \includegraphics[width=\linewidth]{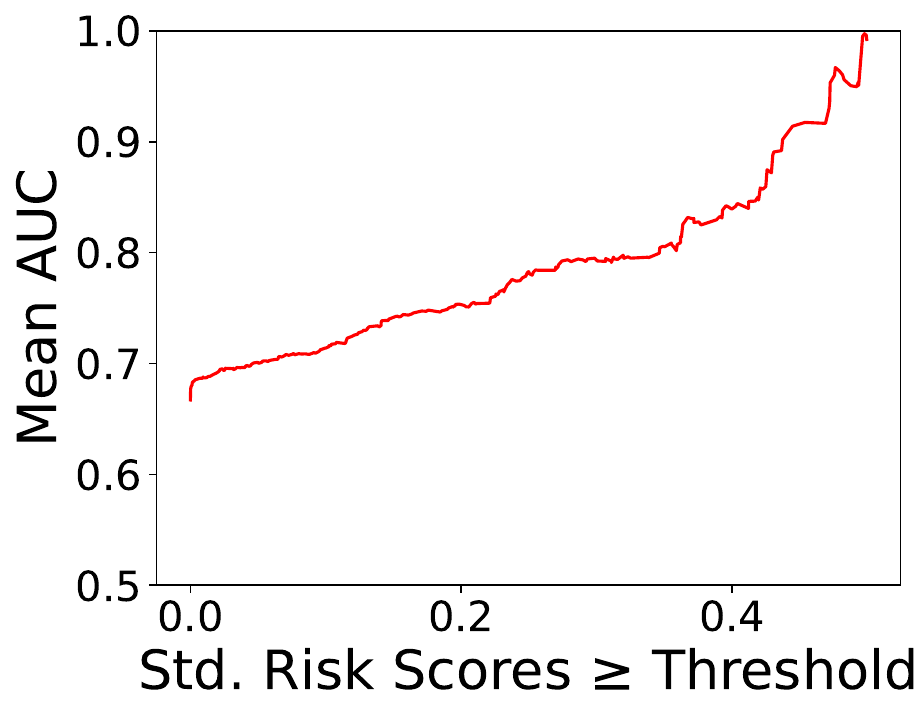}
    \caption{}
    \label{fig:sfig7-3}
\end{subfigure}%
\begin{subfigure}{.5\linewidth}
    \centering
    \includegraphics[width=\linewidth]{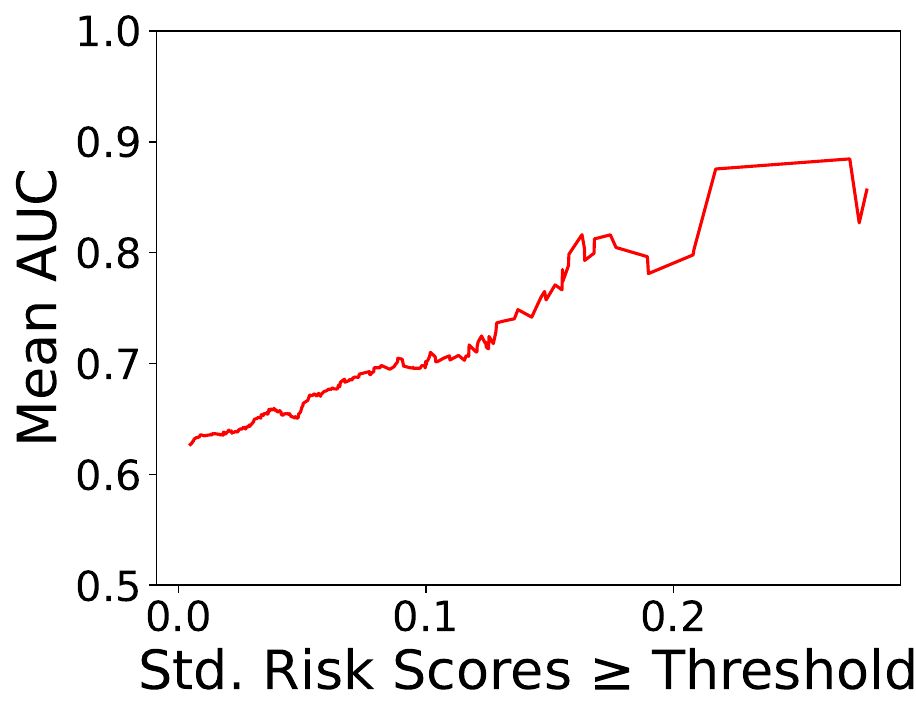}
    \caption{}
    \label{fig:sfig7-4}
\end{subfigure}
\caption{Proxy tasks in the Masking experiment using GPT-4o-mini (a,c) and Llama (b,d), including the original 31 tasks. LOESS curves with 95\% CI shown in (a,b); each point represents predictions on one dataset column. (c,d) show average AUC as the minimum threshold on standard deviation of risk scores increases.
}

\label{fig:fig7}

\end{figure}

\textbf{\textit{Checking the variance of risk scores can aid task-level abstention decisions.}} Aggregating results across all 316 tasks (Figures~\ref{fig:sfig7-1} and~\ref{fig:sfig7-2} for GPT-4o-mini and Llama; Appendix~\ref{appendix:appauc} for GPT-4o and Mistral), we still observe a monotonically increasing relationship between the standard deviation of risk scores and AUC. To measure whether a practitioner would get an informative signal by screening potential tasks according to this metric, Figures~\ref{fig:sfig7-3},~\ref{fig:sfig7-4},~\ref{fig:bryan1-3}, and~\ref{fig:bryan1-4} show the mean AUC on all tasks above a given minimum threshold for the standard deviation, for all LLMs. By raising this threshold, we are able to distinguish tasks with significantly higher than average AUC. For instance, for GPT-4o-mini, the set of all datasets with a standard deviation in risk scores of at least 0.4 has an average AUC of 0.8417, much higher than the average AUC over all datasets (0.7186). While it is important not to rely on this metric absolutely, we suggest that practitioners check whether LLMs make similarly-valued predictions for all individuals, since doing so can help flag datasets where LLMs may not be suitable for zero-shot prediction.

\begin{figure}[!htbp]

\begin{subfigure}{.5\linewidth}
    \centering
    \includegraphics[width=\linewidth]{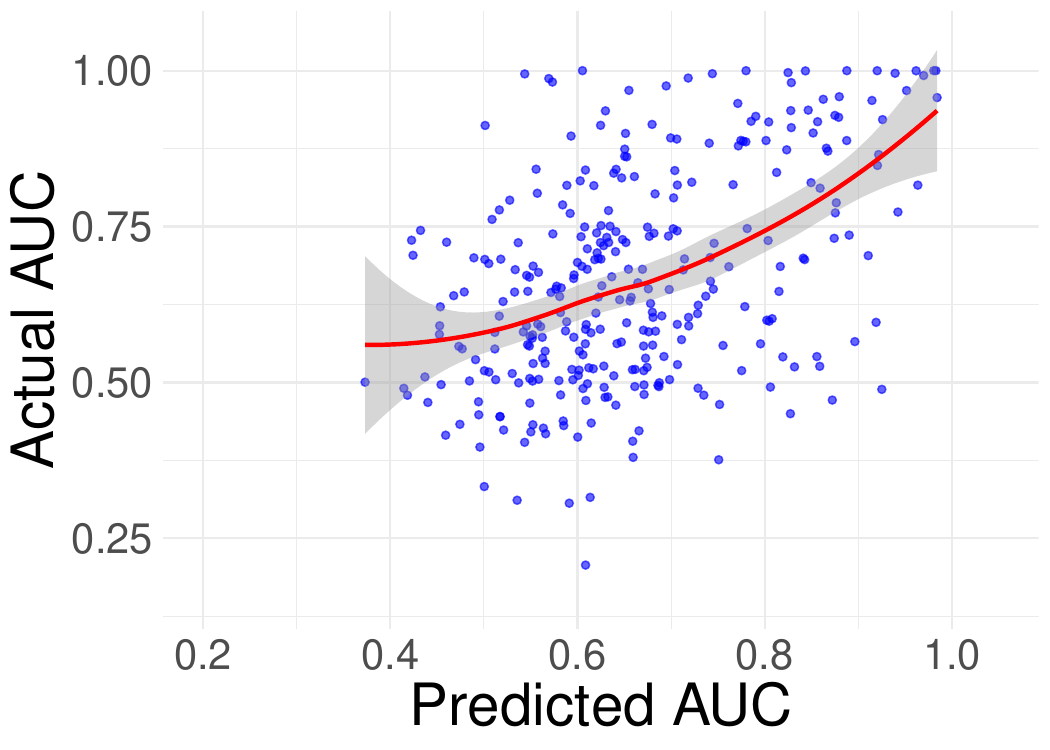}
    \caption{}
    \label{fig:sfig8-1}
\end{subfigure}%
\begin{subfigure}{.5\linewidth}
    \centering
    \includegraphics[width=\linewidth]{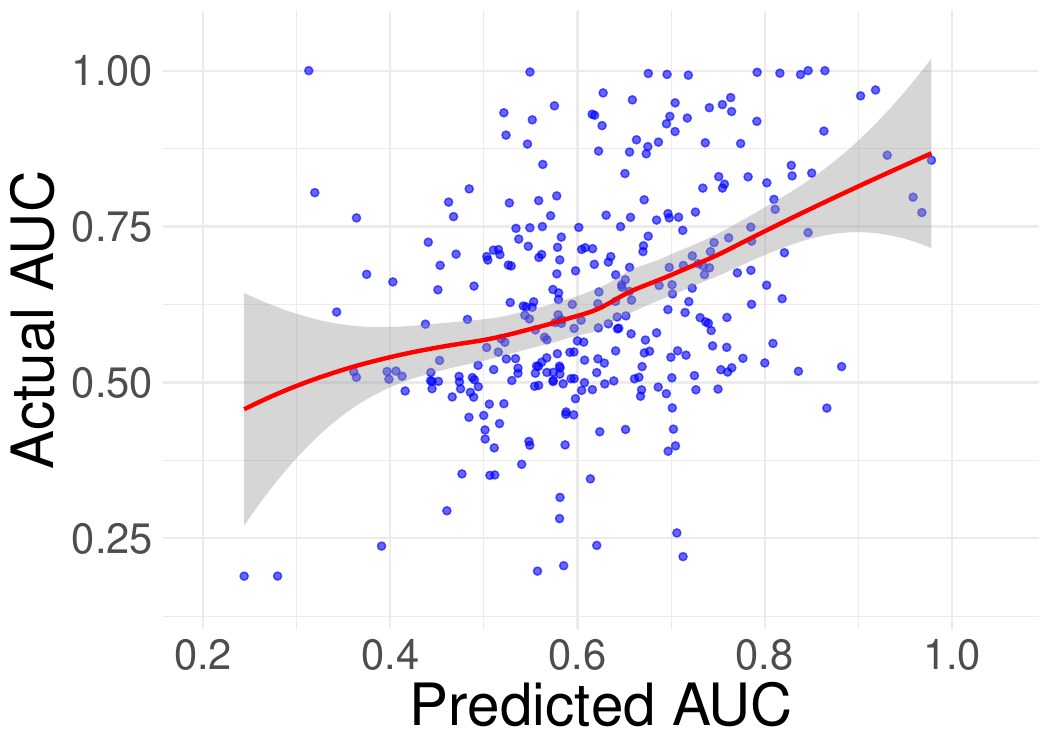}
    \caption{}
    \label{fig:sfig8-2}
\end{subfigure}

\begin{subfigure}{.5\linewidth}
    \centering
    \includegraphics[width=\linewidth]{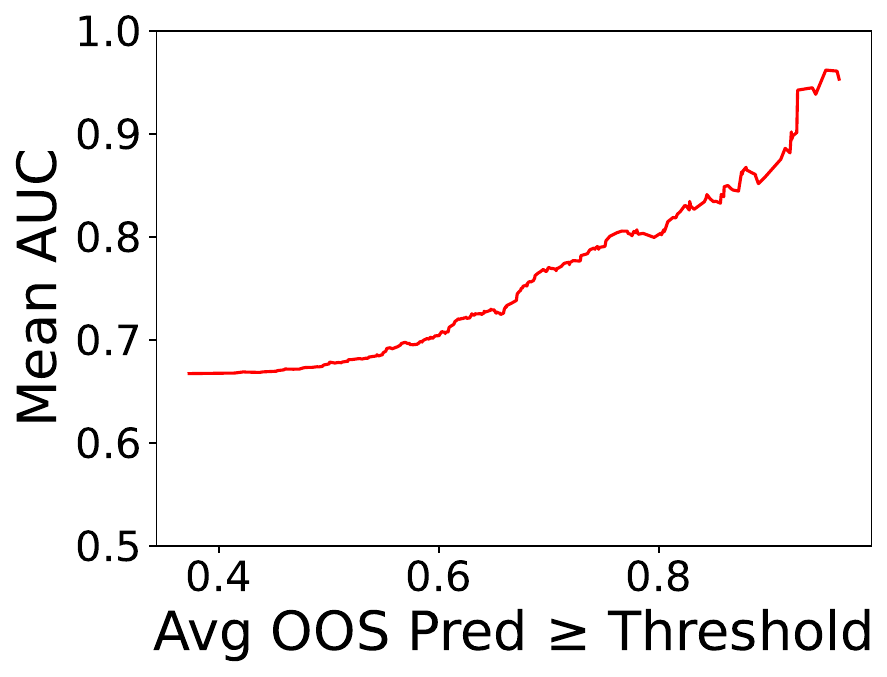}
    \caption{}
    \label{fig:sfig8-3}
\end{subfigure}%
\begin{subfigure}{.5\linewidth}
    \centering
    \includegraphics[width=\linewidth]{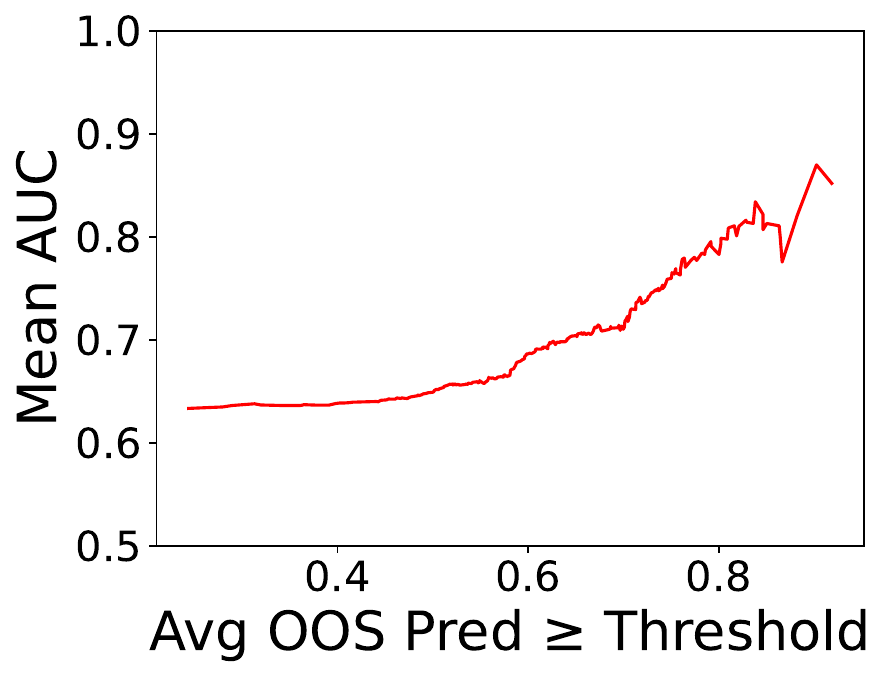}
    \caption{}
    \label{fig:sfig8-4}
\end{subfigure}
\caption{Proxy tasks in the Masking experiment using GPT-4o-mini (a,c) and Llama (b,d), including the original 31 tasks. (a,b) show LOESS fits (with 95\% CI) of actual vs. XGBoost-predicted AUCs, trained via grouped 5-fold cross-validation. Each point represents one prediction task. (c,d) show AUC averages after thresholding on predicted AUCs, analogous to Figures \ref{fig:sfig7-3} and \ref{fig:sfig7-4}.}

\label{fig:fig8}

\end{figure}

\textbf{\textit{The full distribution of predictions captures additional information about performance}}. As the standard deviation of the risk score distribution alone contains significant signal,  we test whether additional information about performance can be gleaned from the full distribution of risk scores. For each task, we discretize the distribution of risk scores from the LLM into 201 values giving each $\alpha$-percentile of the distribution, varying $\alpha$ by 0.5-percentile increments, and train XGBoost models to predict task-level AUC. We use 5-fold cross-validation, grouping by dataset to avoid leakage, so each task's out-of-sample prediction is based on the other 4 folds. Figure~\ref{fig:fig8} plots the average out-of-sample predicted AUC against the actual AUC, along with a LOESS-smoothed curve and 95\% confidence interval, for GPT-4o-mini and Llama (see Appendix~\ref{fig:bryan2} for an equivalent plot for GPT-4o and Mistral). The resulting trend is clearly positive, suggesting that the distribution of LLM-generated risk scores, computed solely on unlabeled data, contains meaningful information about task-level zero-shot performance. The relationship between predicted and actual AUCs becomes somewhat tighter than in Figure \ref{fig:fig7}, particularly for Llama, suggesting that while the standard deviation of the distribution carries much of the signal about performance, other features of the distribution can contribute additional information. Despite the strong correlation between coarser metrics (i.e., standard deviation of risk scores) and downstream AUC for GPT-4o, we still find that the full distribution of risk scores adds meaningful information when regressing on AUC (see Appendix~\ref{fig:bryan2-1}, \ref{fig:bryan2-3}).

\begin{figure}[!t]

\begin{subfigure}{.5\linewidth}
    \centering
    \includegraphics[width=\linewidth]{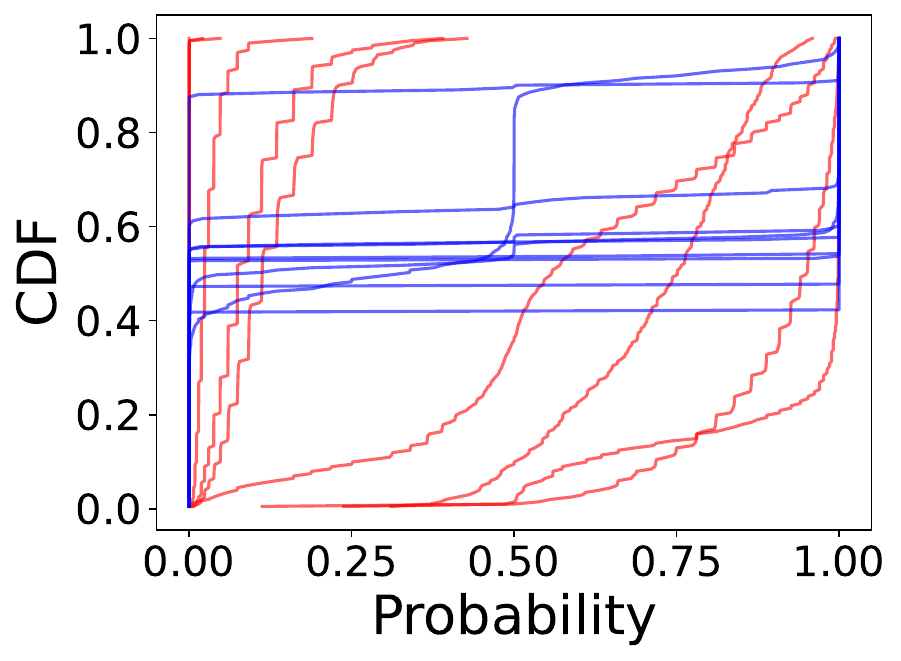}
    \caption{}
    \label{fig:cdfs-1}
\end{subfigure}%
\begin{subfigure}{.5\linewidth}
    \centering
    \includegraphics[width=\linewidth]{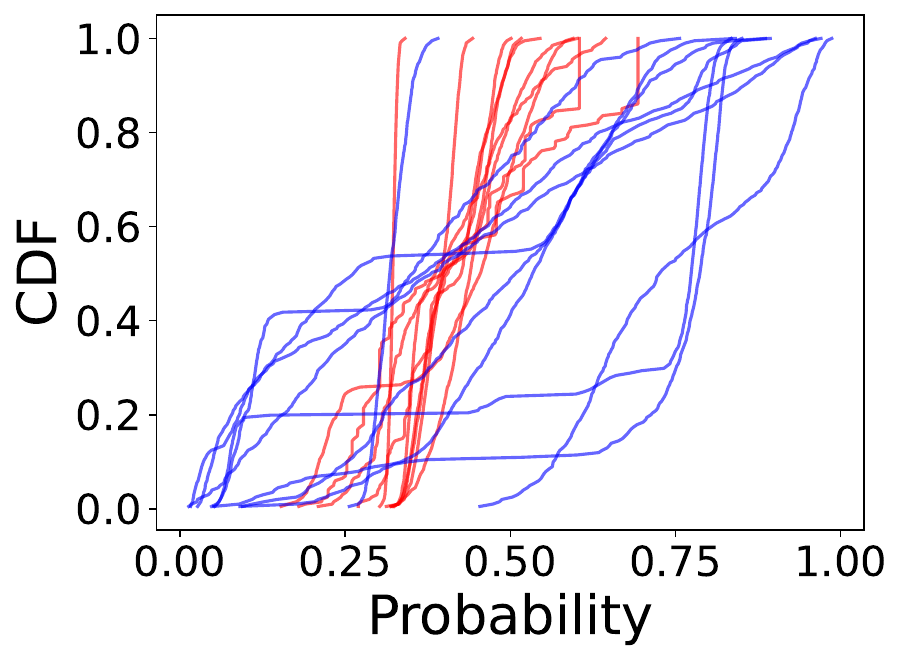}
    \caption{}
    \label{fig:cdfs-2}
\end{subfigure}

\caption{CDFs of the 10 highest (blue) and lowest (red) predicted AUCs over prediction tasks by XGBoost, using 201 percentile values along with standard deviation of risk scores to predict AUC. We observe clear trends within LLMs—for GPT-4o-mini (a), bimodal distributions of risk scores correlate with high XGBoost predictions, whereas for Llama (b), distributions encompassing a wide range of probabilities correlate with high predictions.
}

\label{fig:cdfs}

\end{figure}

To visualize what information the XGBoost models associate with high AUCs, Figure \ref{fig:cdfs} shows the cumulative distribution functions (CDFs) of the LLMs' risk scores for the 10 tasks with the highest and lowest predicted AUCs (see Appendix~\ref{appendix:cdf} for an equivalent plot for GPT-4o and Mistral). Notably, results differ between LLMs. For GPT-4o-mini (Figure \ref{fig:cdfs-1}), high AUC is associated with strongly bimodal risk scores, clustered near 0 or 1. In contrast, for Llama (Figure \ref{fig:cdfs-2}), high AUC aligns with broader, high-variance distributions, while tighter, low-variance distributions correspond to lower AUCs. These differences suggest that the qualitative signals of good performance vary across LLMs. Although both LLMs encode useful information, the way this information manifests differs, indicating a need to analyze distributional traits on a per-model basis.

\section{Conclusion}

While the zero-shot prediction capabilities of LLMs offer exciting opportunities, it remains unclear how to \textit{reliably} employ LLM predictions without validating their outputs on labeled data. We conduct a large-scale empirical study across 316 prediction tasks to explore whether LLMs can serve as reliable zero-shot predictors across a diverse collection of tabular classification tasks. We introduce eight novel task-level metrics for better estimating the LLMs' confidence in the prediction task.

Our findings indicate that performance is highly variable even within individual datasets, so success at one task is no guarantee of success at other tasks on similar data. Instead, measuring the distribution of risk scores for a new task yields both heuristics as well as more sophisticated models that capture a strong signal about the LLMs' performance on that task. However, enough variance in performance remains that such predictions of performance should be seen more as a way to prioritize more promising tasks or screen out ones with a low likelihood of success, not a substitute for eventual validation on labeled data for consequential applications.

\section*{Limitations}

This paper investigates the predictive performance of large language models (LLMs) in zero-shot settings on tabular data, using unlabeled data to estimate task-level performance while drawing new conclusions about individual-level calibrations. While our findings offer novel insights, several limitations merit discussion:

\textit{Memorization or data leakage.}  The datasets that we use are publicly accessible, raising the prospect that they may have appeared in LLM training sets. Our results do imply that LLMs have not memorized the data in the sense of perfectly replicating individual rows, as AUCs vary widely at predicting individual columns within the same dataset given the other columns. Our serialization strategy also alters the presentation of information from the original csv file, which has been found to disrupt some explicit memorization \cite{bordt2024much}. Beyond literal row-by-row memorization though, previous work shows that LLMs perform better at tasks seen more during training, especially for tasks related to retrieval of world knowledge \cite{kandpal2023large,wang2025generalization}. The impact of this phenomena depends on the application at hand—practitioners in many settings may  hope to actually benefit from LLMs having seen relevant data to their application during the training process. Accordingly, proxies for task-level performance that partly pick up on prior exposure to similar tasks may still serve their needs. However, using public data does represent a potential limitation in external validity for our results; we can't rule out that predictors of task-level performance might be different in domains that are completely unseen during LLM training.

\textit{Model access and scale.} We rely on LLMs that expose token-level probabilities (GPT-4o-mini, GPT-4o, Mistral-7b-Instruct-v0.1, and Llama-3.1-8b-Instruct), which may not generalize to other models without such access or with substantially different architectures. Larger models, or models with distinct fine-tuning or pretraining regimes, may behave differently.

\textit{Prompting.} Our serialization of tabular data uses fixed, template-based formats (i.e., “Feature: Value” pairs, followed by binary questions). Our prompting approaches do not explore alternative prompts, few-shot settings, or chain-of-thought reasoning.

\bibliography{custom}
\clearpage

\appendix
\onecolumn

\section{Appendix}
\label{sec:appendix}

\subsection{Dataset Descriptions, Sources, and Artifacts}
\label{appendix:datadesc}

\begin{table}[H]
\small
\centering
\resizebox{\textwidth}{!}{
\begin{tabular}{@{}p{2.5cm}p{8.5cm}p{4.5cm}@{}}
\toprule
\textbf{Dataset Name} & \textbf{Short Description} & \textbf{Source} \\
\midrule
acsincome & ACSIncome task from the folktables package. & \url{https://github.com/socialfoundations/folktables} \\
acsmobility & ACSMobility task from the folktables package. & \url{https://github.com/socialfoundations/folktables} \\
acspubcov & ACSPublicCoverage task from the folktables package. & \url{https://github.com/socialfoundations/folktables} \\
acstraveltime & ACSTravelTime task from the folktables package. & \url{https://github.com/socialfoundations/folktables} \\
acsunemployment & ACSEmployment task from the folktables package. & \url{https://github.com/socialfoundations/folktables} \\
airline & Predict flight delays based on scheduled departure info. & \url{https://www.openml.org/d/42493} \\
bank & Predict term deposit subscription in a marketing campaign. & \url{https://www.openml.org/d/1558} \\
brfssdiabetes & Predict whether a patient has diabetes (BRFSS survey). & \url{https://github.com/mlfoundations/tableshift} \\
brfsshbp & Predict hypertension diagnosis for 50+ age group. & \url{https://github.com/mlfoundations/tableshift} \\
brfsshighcholesterol & Predict high cholesterol in BRFSS survey data. & \url{https://github.com/mlfoundations/tableshift} \\
car & Predict acceptability of cars from evaluation records. & \url{https://archive.ics.uci.edu/dataset/19} \\
diabetes & Predict readmission of diabetic patients within 30 days. & \url{https://archive.ics.uci.edu/dataset/296} \\
glioma & Classify glioma (brain tumor) grade. & \url{https://archive.ics.uci.edu/dataset/759} \\
houses & Predict if California housing value exceeds \$200k. & \url{https://www.openml.org/d/537} \\
indiandiabetes & Predict diabetes using diagnostic features. & \url{https://www.kaggle.com/datasets/uciml/pima-indians-diabetes-database} \\
ipums & Predict facility birth in Latin/Caribbean countries. & \url{https://globalhealth.ipums.org/} \\
mushroom & Classify mushrooms as edible or poisonous. & \url{https://archive.ics.uci.edu/dataset/73} \\
nursery & Prioritize nursery school applications. & \url{https://archive.ics.uci.edu/dataset/76} \\
rice & Classify Turkish rice grains as Osmancik or Cammeo. & \url{https://archive.ics.uci.edu/dataset/545} \\
sepsis & Predict ICU patient risk of sepsis within 6 hours. & \url{https://github.com/mlfoundations/tableshift} \\
support2 & Predict hospital death of critically ill patients. & \url{https://archive.ics.uci.edu/dataset/880} \\
taxibog & Predict long taxi rides in Bogota. & \url{https://www.kaggle.com/datasets/mnavas} \\
taximex & Predict long taxi rides in Mexico City. & \url{https://www.kaggle.com/datasets/mnavas} \\
taxiuio & Predict long taxi rides in Quito. & \url{https://www.kaggle.com/datasets/mnavas} \\
telescope & Classify cosmic ray vs gamma signal events. & \url{https://archive.ics.uci.edu/dataset/159} \\
ucibreastcancer & Predict breast mass as malignant or benign. & \url{https://archive.ics.uci.edu/dataset/15} \\
ucidiabetes & Predict diabetes using lifestyle statistics. & \url{https://archive.ics.uci.edu/dataset/15} \\
uciheart & Predict heart disease diagnosis. & \url{https://archive.ics.uci.edu/dataset/45} \\
ucispambase & Classify email as spam or not spam. & \url{https://archive.ics.uci.edu/dataset/94} \\
ucistatloggerman & Classify credit risk from attributes. & \url{https://archive.ics.uci.edu/dataset/144} \\
usaccidents & Predict severity of US traffic accidents. & \url{https://www.kaggle.com/datasets/sobhanmoosavi/us-accidents} \\
\bottomrule
\end{tabular}
}
\end{table}

For more details regarding our dataset sources and other artifacts:

\begin{itemize}
    \item The \texttt{car}, \texttt{diabetes}, \texttt{glioma}, \texttt{mushroom}, \texttt{nursery}, \texttt{rice}, \texttt{support2}, \texttt{telescope}, \texttt{ucibreastcancer}, \texttt{ucidiabetes}, \texttt{uciheart}, \texttt{ucispambase}, and \texttt{ucistatloggerman} datasets all come from the UCI repository \citep{Dua:2019}.
    \item The \texttt{acsincome}, \texttt{acsmobility}, \texttt{acspubcov}, \texttt{acstraveltime}, and \texttt{acsunemployment} datasets all come from the Folktables repository \citep{ding2021retiring}.
    \item The \texttt{brfssdiabetes}, \texttt{brfsshbp}, \texttt{brfsshighcholesterol}, and \texttt{sepsis} datasets all come from the Tableshift repository \citep{gardner2023tableshift}.
    \item The \texttt{airline}, \texttt{bank}, and \texttt{house} datasets all come from OpenML \citep{openml}.
    \item The \texttt{indiandiabetes}, \texttt{taxibog}, \texttt{taximex}, \texttt{taxiuio}, and \texttt{usaccidents} datasets all come from Kaggle \citep{ucidiabeteskaggle, mnavas2022taxidata, moosavi2020usaccidents}.
    \item the \texttt{ipums} dataset is curated from the IPUMS Global Health repository of international health survey data \citep{ipums}.
    \item All datasets are publicly available and we release our data for replication, with the exception of the \texttt{ipums} data, which required individual-level dataset requests for the de-identified data on maternal outcomes, and is thus not released. All data is compliant with anonymization policies (i.e., de-identified) and does not contain offensive or sensitive content.
    \item We use GPT-4o-mini, GPT-4o \citep{hurst2024gpt}, Mistral-7b-Instruct-v0.1 \citep{chaplot2023albert}, and Llama-3.1-8b-Instruct \citep{grattafiori2024llama} as LLMs for predictive modeling, for all experiments. All models contain publicly available APIs for personal and research use. Furthermore, Mistral-7b-Instruct-v0.1 and Llama-3.1-8b-Instruct make their model weights publicly available.

\end{itemize}

\clearpage

\subsection{LLM Metrics Table}
\label{appendix:metrics}

\begin{table}[H]
\centering
\footnotesize
\begin{tabularx}{\textwidth}{lXXXX}
\toprule
\textbf{Dataset} & \textbf{GPT-4o-mini AUC} & \textbf{GPT-4o-mini ECE} & \textbf{Llama AUC} & \textbf{Llama ECE} \\
\midrule
taxiuio & 0.8794 & 0.0971 & 0.7929 & 0.3087 \\
mushroom & 0.8881 & 0.2900 & 0.6931 & 0.1676 \\
acsincome & 0.8655 & 0.1939 & 0.8481 & 0.2812 \\
support2 & 0.8904 & 0.1369 & 0.8644 & 0.2953 \\
telescope & 0.4322 & 0.6490 & 0.4900 & 0.3038 \\
nursery & 0.8368 & 0.2425 & 0.7776 & 0.1163 \\
diabetes & 0.4979 & 0.0960 & 0.5235 & 0.1940 \\
brfssdiabetes & 0.6497 & 0.1540 & 0.7144 & 0.0706 \\
airline & 0.4768 & 0.0697 & 0.4779 & 0.1867 \\
bank & 0.6805 & 0.1115 & 0.5507 & 0.0854 \\
acspubcov & 0.7232 & 0.2211 & 0.6963 & 0.0723 \\
ucistatloggerman & 0.4499 & 0.4677 & 0.4589 & 0.4457 \\
brfsshbp & 0.7249 & 0.4550 & 0.7052 & 0.1633 \\
usaccidents & 0.5974 & 0.1980 & 0.7300 & 0.1535 \\
uciheart & 0.8756 & 0.2504 & 0.8117 & 0.1348 \\
IndianDiabetes & 0.7882 & 0.4330 & 0.7971 & 0.0877 \\
taxibog & 0.8730 & 0.0770 & 0.8202 & 0.1923 \\
ucispambase & 0.8921 & 0.2954 & 0.7491 & 0.1632 \\
ucidiabetes & 0.6624 & 0.3149 & 0.7133 & 0.4992 \\
glioma & 0.8837 & 0.2426 & 0.3511 & 0.2808 \\
rice & 0.4907 & 0.3090 & 0.6011 & 0.2925 \\
acstraveltime & 0.6599 & 0.3724 & 0.6556 & 0.0357 \\
acsmobility & 0.5803 & 0.1427 & 0.5779 & 0.1153 \\
car & 0.9121 & 0.1308 & 0.8564 & 0.1067 \\
acsunemployment & 0.8880 & 0.4190 & 0.8711 & 0.2171 \\
houses & 0.4935 & 0.4083 & 0.4659 & 0.1077 \\
sepsis & 0.5936 & 0.0273 & 0.5950 & 0.2442 \\
brfsshighcholesterol & 0.6977 & 0.4171 & 0.6846 & 0.1722 \\
ucibreastcancer & 0.8115 & 0.5732 & 0.9436 & 0.4302 \\
taximex & 0.8859 & 0.0494 & 0.8180 & 0.2640 \\
ipums & 0.6970 & 0.3519 & 0.7080 & 0.1563 \\
\bottomrule
 \end{tabularx}
\end{table}

\clearpage

\begin{table}[H]
\centering
\footnotesize
\begin{tabularx}{\textwidth}{lXXXX}
\toprule
\textbf{Dataset} & \textbf{GPT-4o AUC} & \textbf{GPT-4o ECE} & \textbf{Mistral AUC} & \textbf{Mistral ECE} \\
\midrule
taxiuio & 0.9115 & 0.0845 & 0.7166 & 0.4622 \\
mushroom & 0.9964 & 0.2077 & 0.8277 & 0.2423 \\
acsincome & 0.8801 & 0.1696 & 0.8102 & 0.1016 \\
support2 & 0.9226 & 0.0921 & 0.7242 & 0.5444 \\
telescope & 0.6145 & 0.2760 & 0.6420 & 0.0480 \\
nursery & 0.7935 & 0.2672 & 0.8622 & 0.1841 \\
diabetes & 0.4935 & 0.1304 & 0.4973 & 0.2322 \\
brfssdiabetes & 0.8239 & 0.0894 & 0.7785 & 0.1827 \\
airline & 0.5084 & 0.1529 & 0.5114 & 0.2466 \\
bank & 0.8480 & 0.4554 & 0.5559 & 0.1709 \\
acspubcov & 0.7238 & 0.2866 & 0.6122 & 0.0780 \\
ucistatloggerman & 0.4962 & 0.3348 & 0.4436 & 0.2079 \\
brfsshbp & 0.7538 & 0.2986 & 0.6665 & 0.0886 \\
usaccidents & 0.5527 & 0.2241 & 0.6028 & 0.3827 \\
uciheart & 0.8983 & 0.2046 & 0.7593 & 0.2384 \\
IndianDiabetes & 0.8231 & 0.2400 & 0.7723 & 0.1712 \\
taxibog & 0.8895 & 0.1253 & 0.7495 & 0.3091 \\
ucispambase & 0.9594 & 0.1051 & 0.7447 & 0.0519 \\
ucidiabetes & 0.7248 & 0.5399 & 0.6910 & 0.2811 \\
glioma & 0.8988 & 0.1425 & 0.4919 & 0.1840 \\
rice & 0.4950 & 0.0449 & 0.2742 & 0.0616 \\
acstraveltime & 0.7038 & 0.3325 & 0.5685 & 0.1266 \\
acsmobility & 0.5628 & 0.1987 & 0.5293 & 0.1105 \\
car & 0.9250 & 0.0986 & 0.8565 & 0.2582 \\
acsunemployment & 0.8896 & 0.1346 & 0.5654 & 0.2648 \\
houses & 0.8890 & 0.1692 & 0.4260 & 0.0345 \\
sepsis & 0.6568 & 0.1264 & 0.5929 & 0.3422 \\
brfsshighcholesterol & 0.7139 & 0.3755 & 0.6982 & 0.0806 \\
ucibreastcancer & 0.9911 & 0.0358 & 0.9597 & 0.3950 \\
taximex & 0.8965 & 0.0868 & 0.7839 & 0.4085 \\
ipums & 0.7225 & 0.1912 & 0.6658 & 0.2763 \\
\bottomrule
\end{tabularx}
\end{table}

\clearpage

\subsection{Additional AUC and ECE Histograms}
\label{appendix:hist}

\begin{figure}[ht]
\begin{subfigure}{.5\linewidth}
  \centering
  \includegraphics[width=\linewidth]{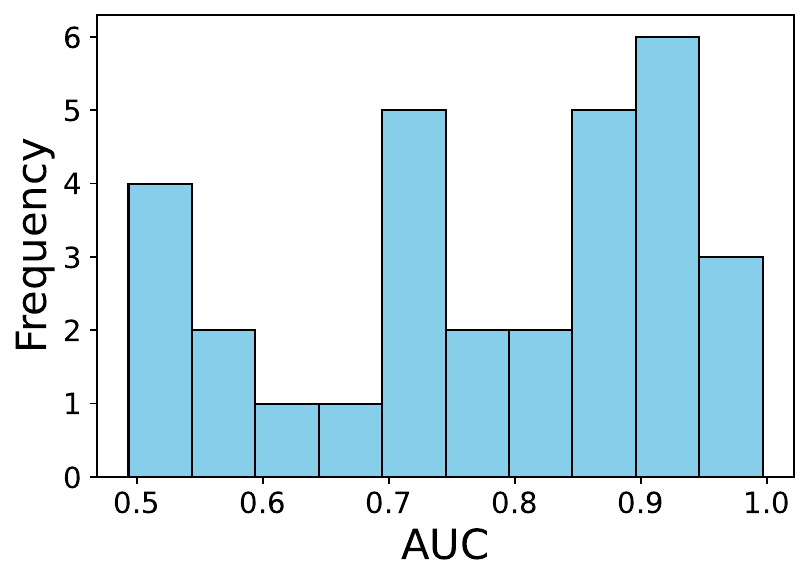}
  \caption{}
  \label{fig:sfigtable-1}
\end{subfigure}%
\begin{subfigure}{.5\linewidth}
  \centering
  \includegraphics[width=\linewidth]{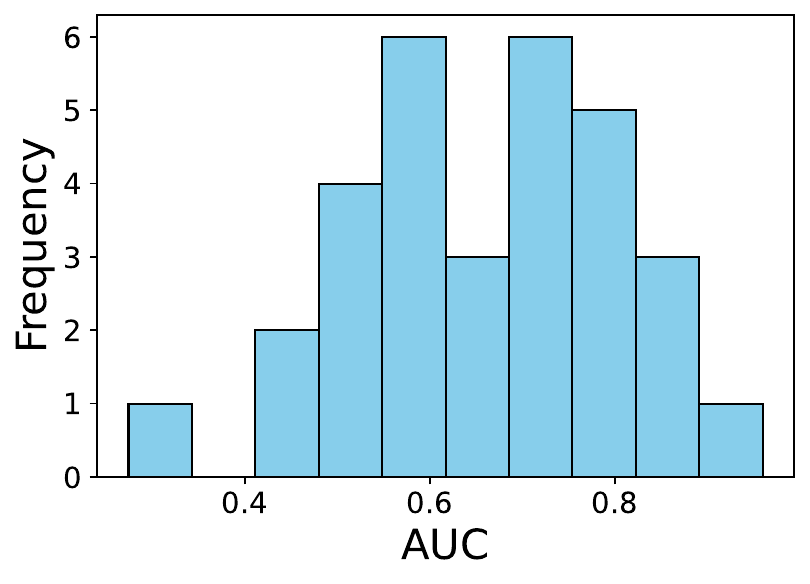}
  \caption{}
  \label{fig:sfigtable-2}
\end{subfigure}

\begin{subfigure}{.5\linewidth}
  \centering
  \includegraphics[width=\linewidth]{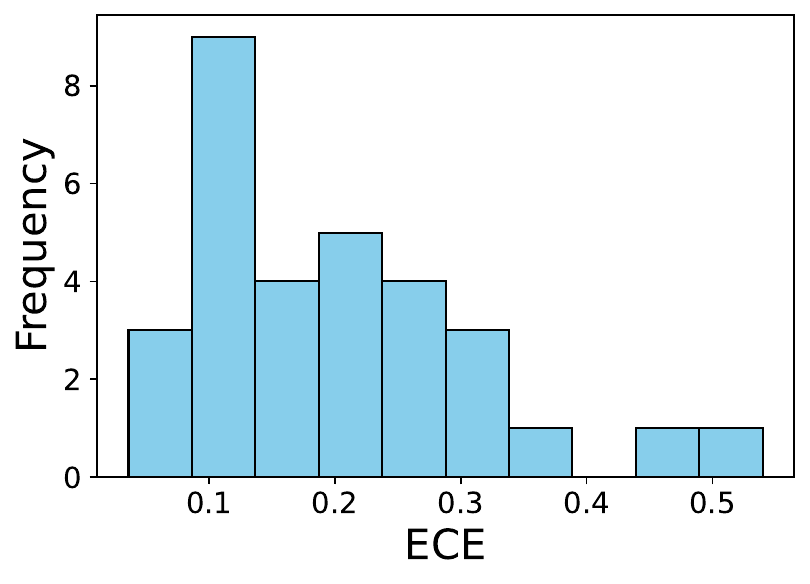}
  \caption{}
  \label{fig:sfigtable-3}
\end{subfigure}%
\begin{subfigure}{.5\linewidth}
  \centering
  \includegraphics[width=\linewidth]{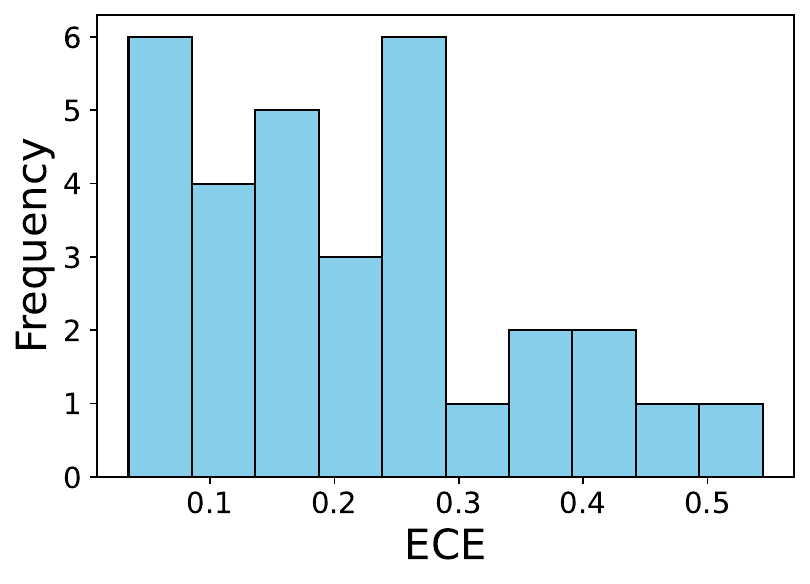}
  \caption{}
  \label{fig:sfigtable-4}
\end{subfigure}

\caption{Histograms of AUC and ECE over all datasets, for GPT-4o (a,c) and Mistral-7b-Instruct-v0.1 (b,d).}
\label{fig:apptable}
\end{figure}

\clearpage

\subsection{Correlation between Metrics and AUC}
\label{appendix:llamacor}

\begin{figure}[H]
\begin{subfigure}{.25\linewidth}
  \centering
  \includegraphics[width=\linewidth]{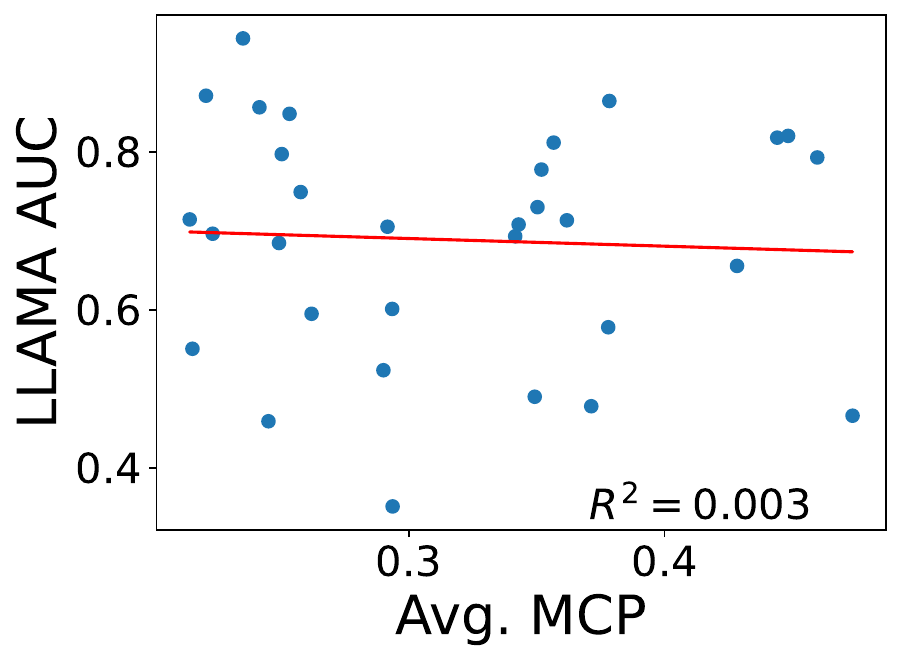}
  \caption{}
  \label{fig:llamacor-1}
\end{subfigure}%
\begin{subfigure}{.25\linewidth}
  \centering
  \includegraphics[width=\linewidth]{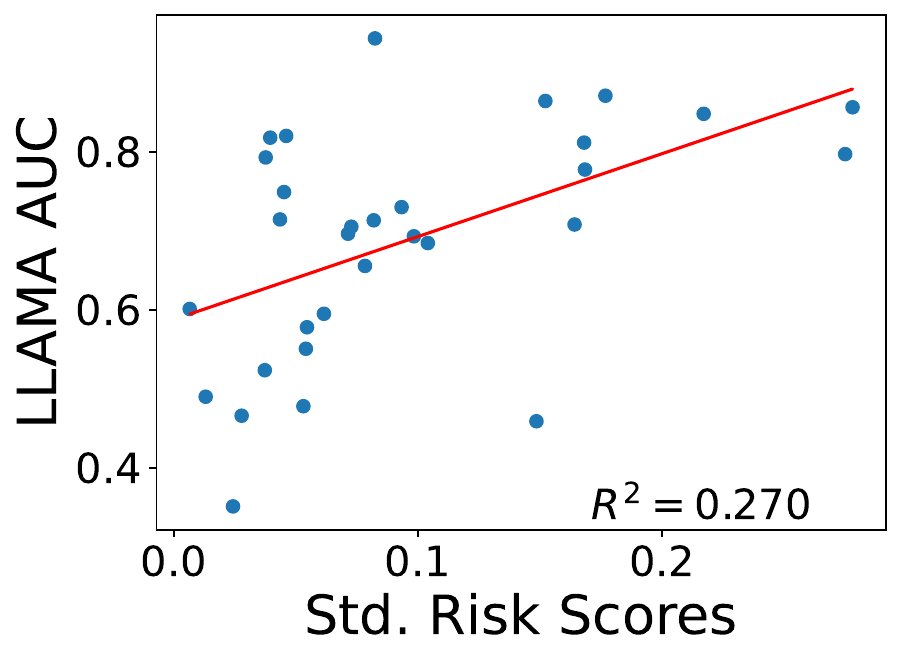}
  \caption{}
  \label{fig:llamacor-2}
\end{subfigure}%
\begin{subfigure}{.25\linewidth}
  \centering
  \includegraphics[width=\linewidth]{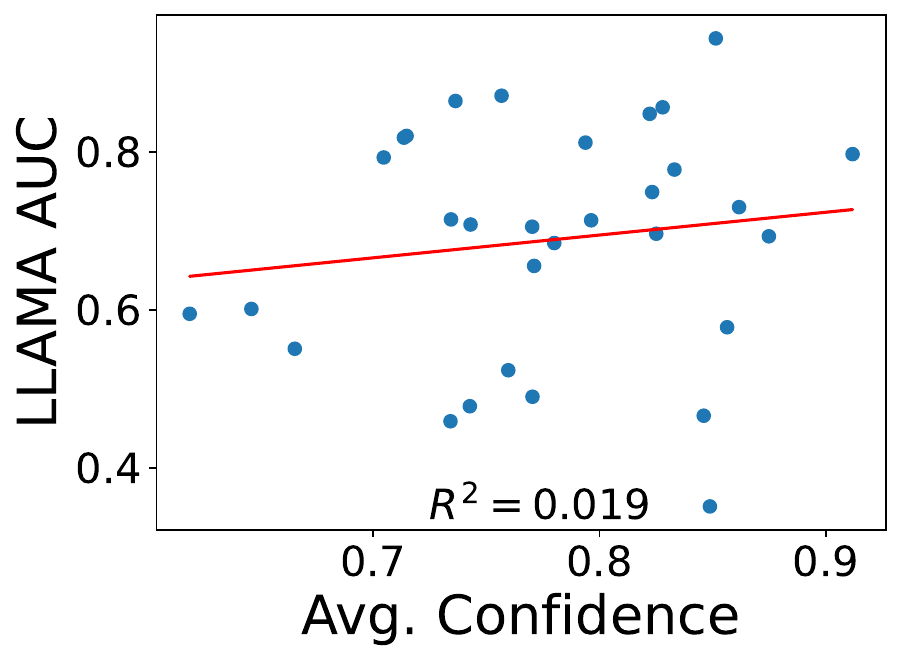}
  \caption{}
  \label{fig:llamacor-3}
\end{subfigure}%
\begin{subfigure}{.25\linewidth}
  \centering
  \includegraphics[width=\linewidth]{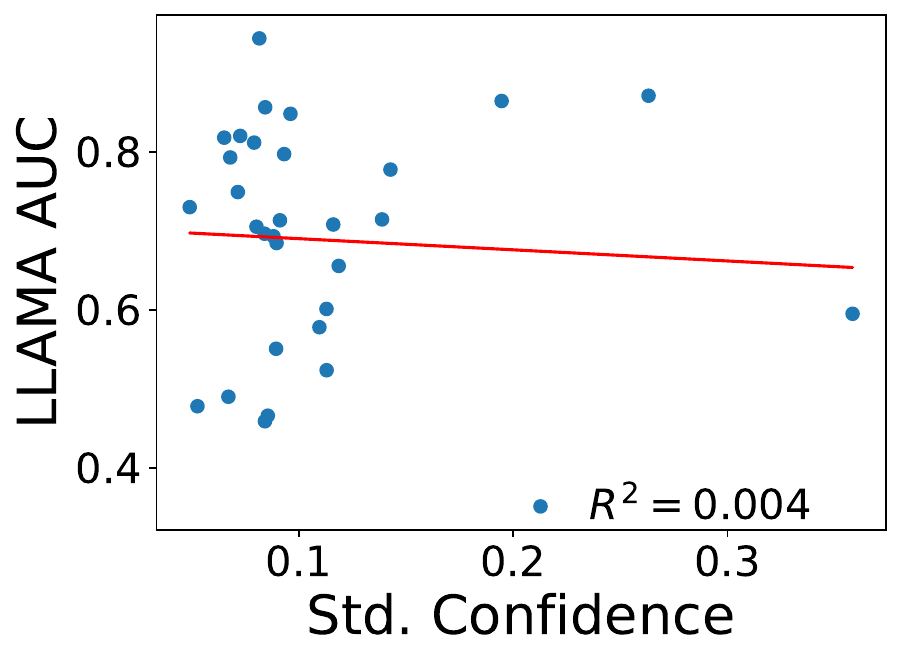}
  \caption{}
  \label{fig:llamacor-4}
\end{subfigure}

\begin{subfigure}{.25\linewidth}
  \centering
  \includegraphics[width=\linewidth]{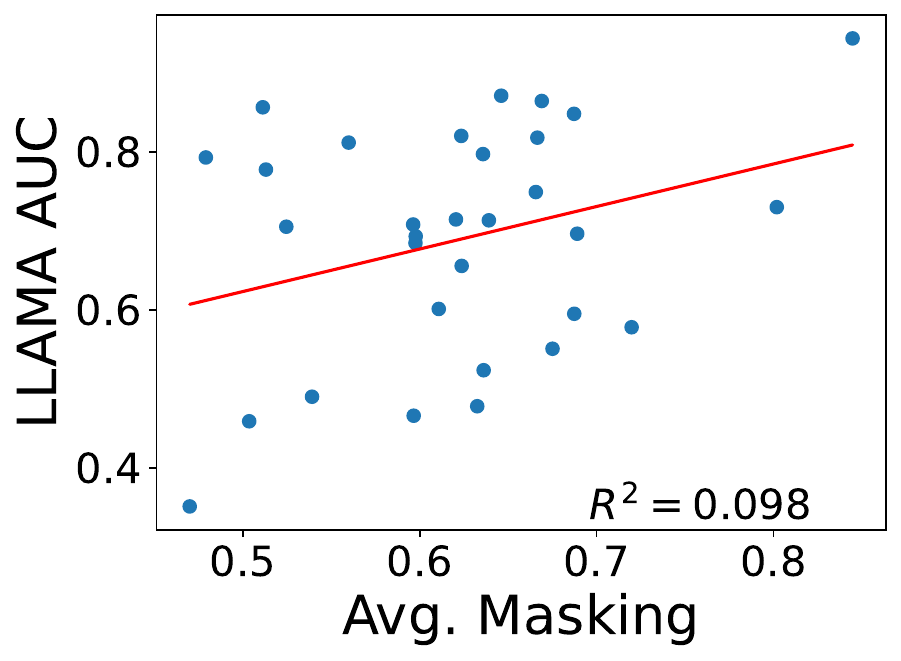}
  \caption{}
  \label{fig:llamacor-5}
\end{subfigure}%
\begin{subfigure}{.25\linewidth}
  \centering
  \includegraphics[width=\linewidth]{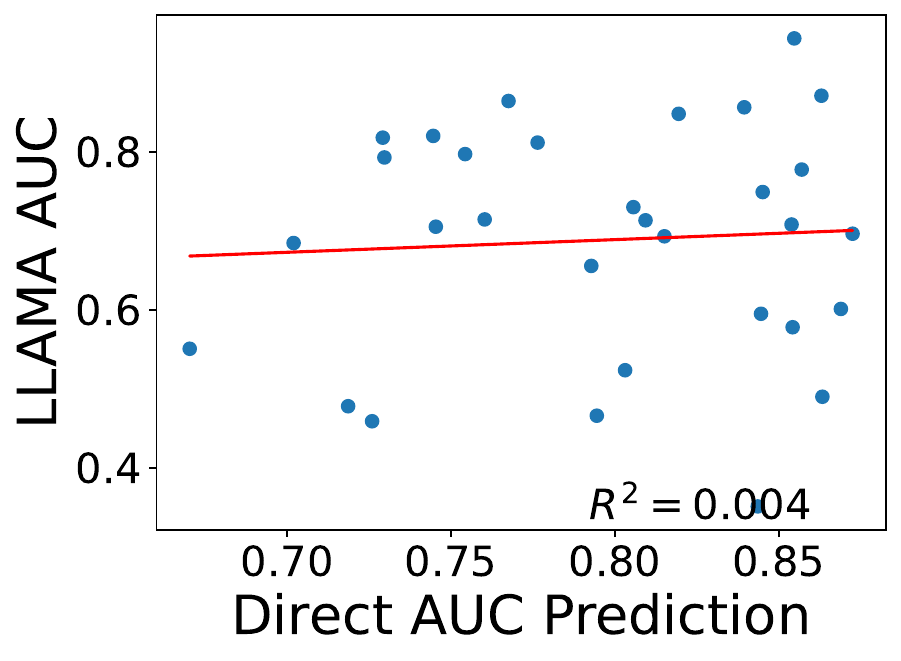}
  \caption{}
  \label{fig:llamacor-6}
\end{subfigure}%
\begin{subfigure}{.25\linewidth}
  \centering
  \includegraphics[width=\linewidth]{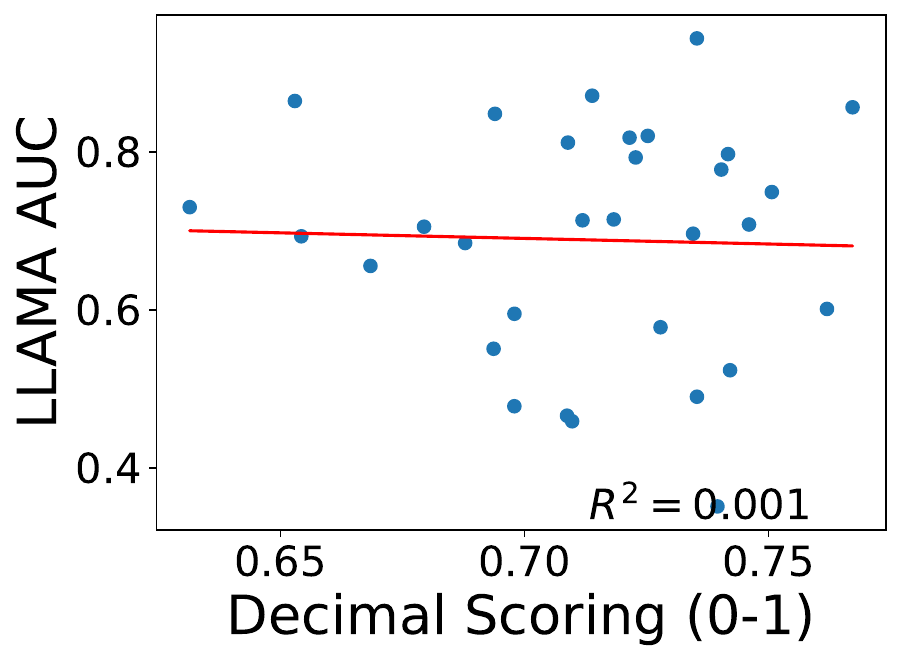}
  \caption{}
  \label{fig:sfigaucraw-7}
\end{subfigure}%
\begin{subfigure}{.25\linewidth}
  \centering
  \includegraphics[width=\linewidth]{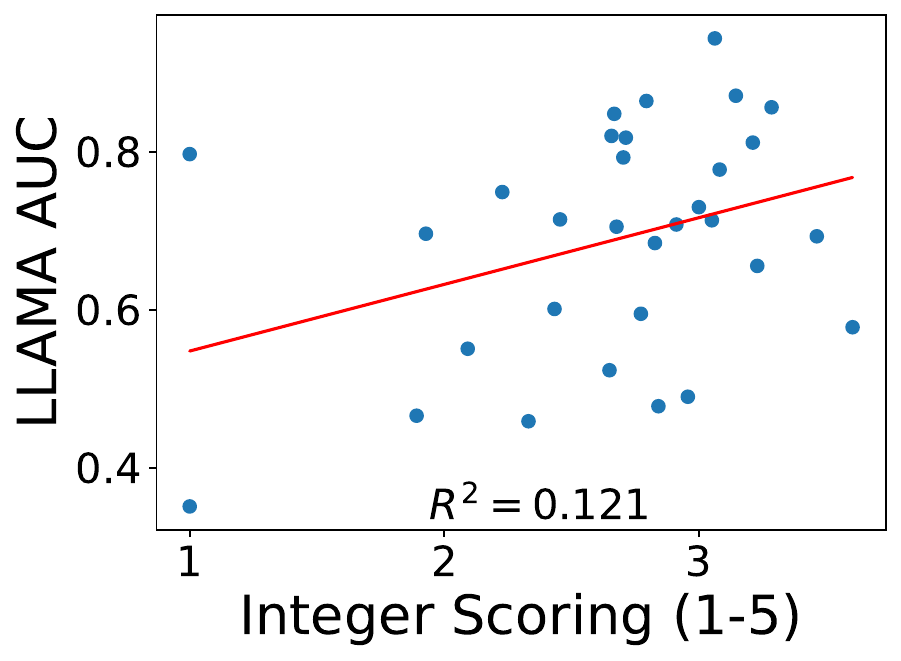}
  \caption{}
  \label{fig:llamacor-9}
\end{subfigure}
\caption{Correlation between aggregate metrics derived from our experiments on the unlabeled datasets and the AUC scores of Llama-3.1-8b-Instruct on each of the datasets, where each point represents one dataset. We plot the best-fit line with its corresponding $R^2$ value for each metric.}
\label{fig:llamacor}

\end{figure}

\begin{figure}[H]
\begin{subfigure}{.25\linewidth}
  \centering
\includegraphics[width=\linewidth]{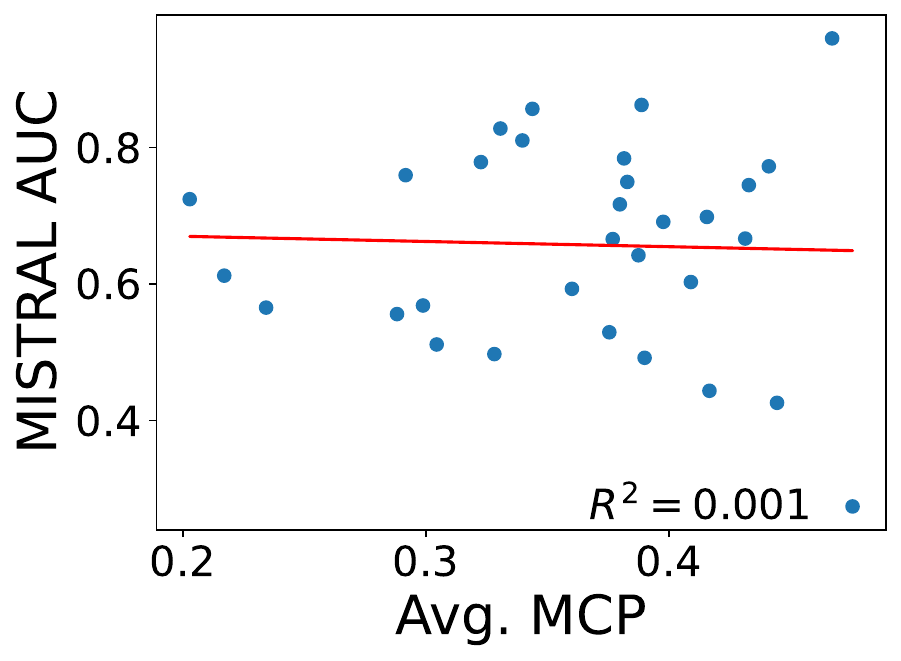}
  \caption{}
  \label{fig:sfigmistral1-1}
\end{subfigure}%
\begin{subfigure}{.25\linewidth}
  \centering
\includegraphics[width=\linewidth]{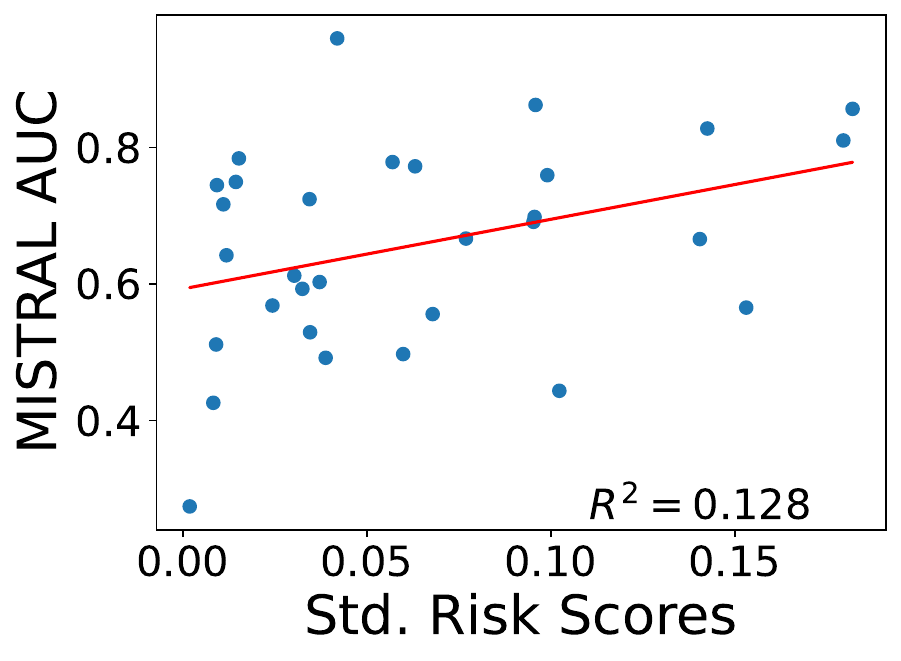}
  \caption{}
  \label{fig:sfigmistral1-2}
\end{subfigure}%
\begin{subfigure}{.25\linewidth}%
  \centering
\includegraphics[width=\linewidth]{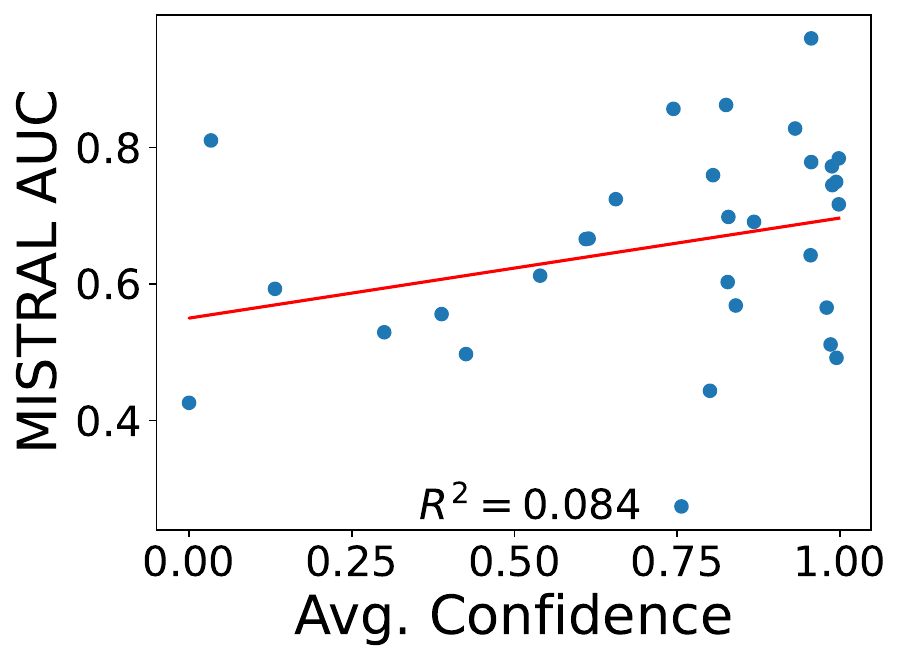}
  \caption{}
  \label{fig:sfigmistral1-3}
\end{subfigure}%
\begin{subfigure}{.25\linewidth}
  \centering
\includegraphics[width=\linewidth]{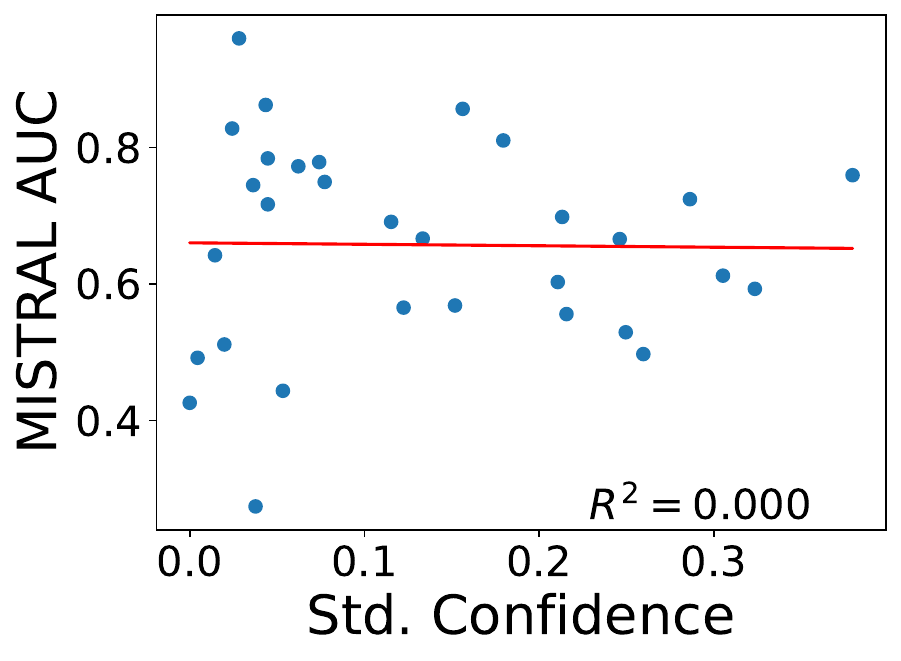}
  \caption{}
  \label{fig:sfigmistral1-4}
\end{subfigure}

\begin{subfigure}{.25\linewidth}
  \centering
  \includegraphics[width=\linewidth]{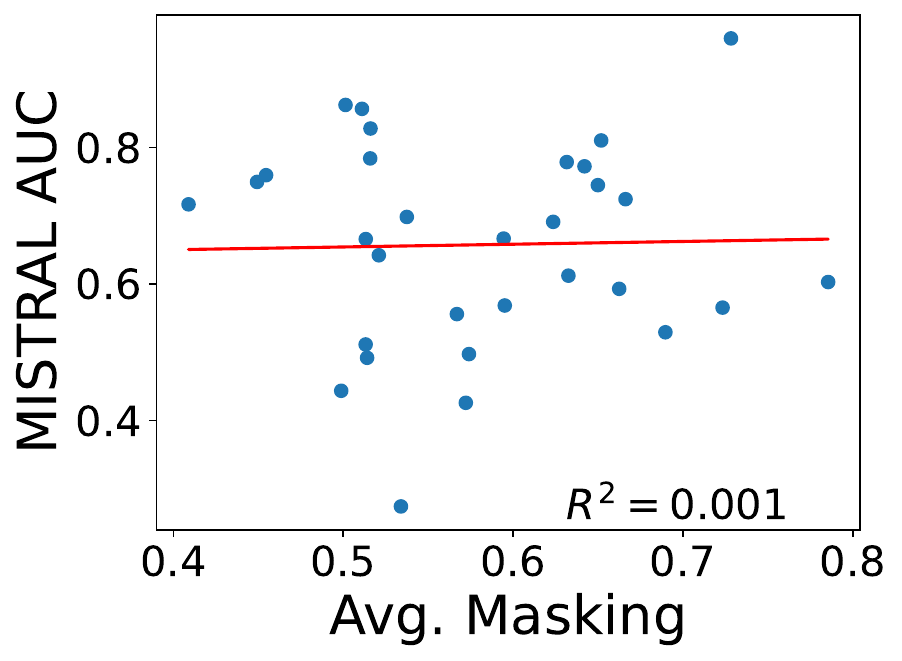}
  \caption{}
  \label{fig:sfigmistral1-5}
\end{subfigure}%
\begin{subfigure}{.25\linewidth}
  \centering
  \includegraphics[width=\linewidth]{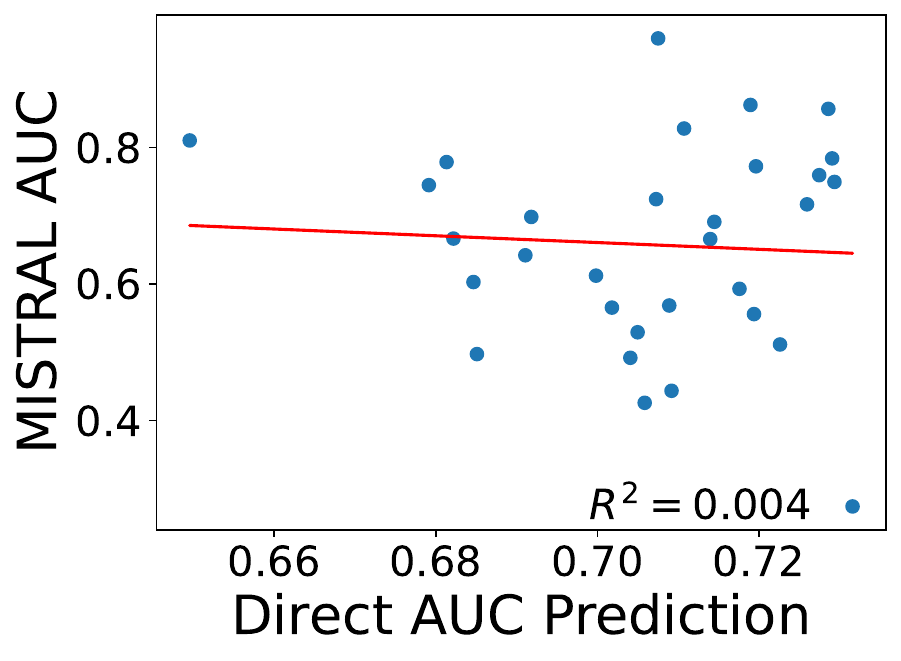}
  \caption{}
  \label{fig:sfigmistral1-6}
\end{subfigure}%
\begin{subfigure}{.25\linewidth}
  \centering
  \includegraphics[width=\linewidth]{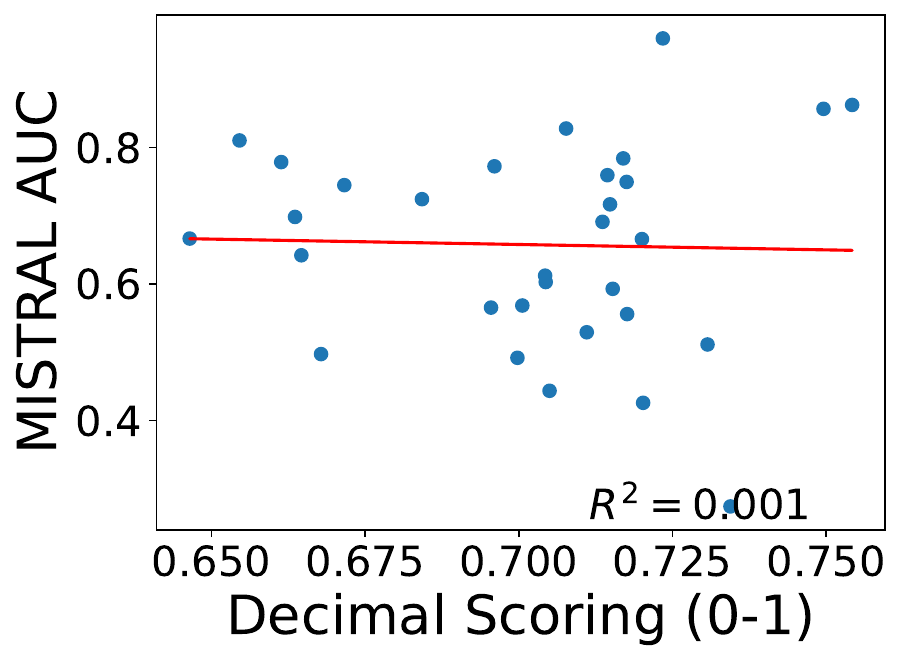}
  \caption{}
  \label{fig:sfigmistral1-7}
\end{subfigure}%
\begin{subfigure}{.25\linewidth}
  \centering
  \includegraphics[width=\linewidth]{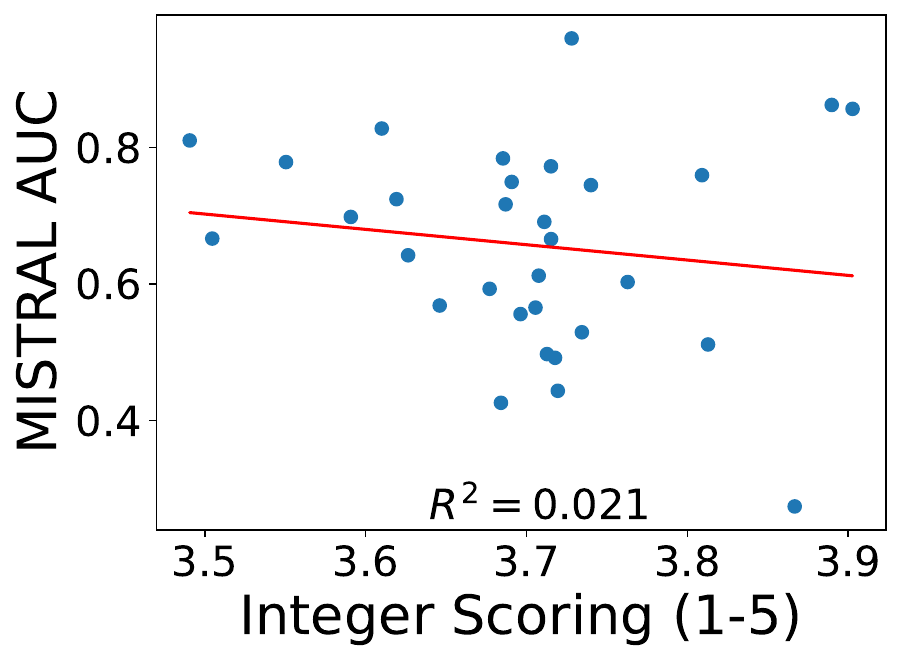}
  \caption{}
  \label{fig:sfigmistral1-9}
\end{subfigure}
\caption{Correlation between aggregate metrics derived from our experiments on the unlabeled datasets and the AUC scores of Mistral-7B-Instruct-v0.1 on each of the datasets, where each point represents one dataset. We plot the best-fit line with its corresponding $R^2$ value for each metric.}
\label{fig:figmistral1}

\end{figure}
\clearpage

\subsection{Normalized AUC Scores, Masking Experiment}
\label{appendix:normauc}

\begin{figure}[H]

\begin{subfigure}{0.42\linewidth}
    \centering
    \includegraphics[width=\linewidth]{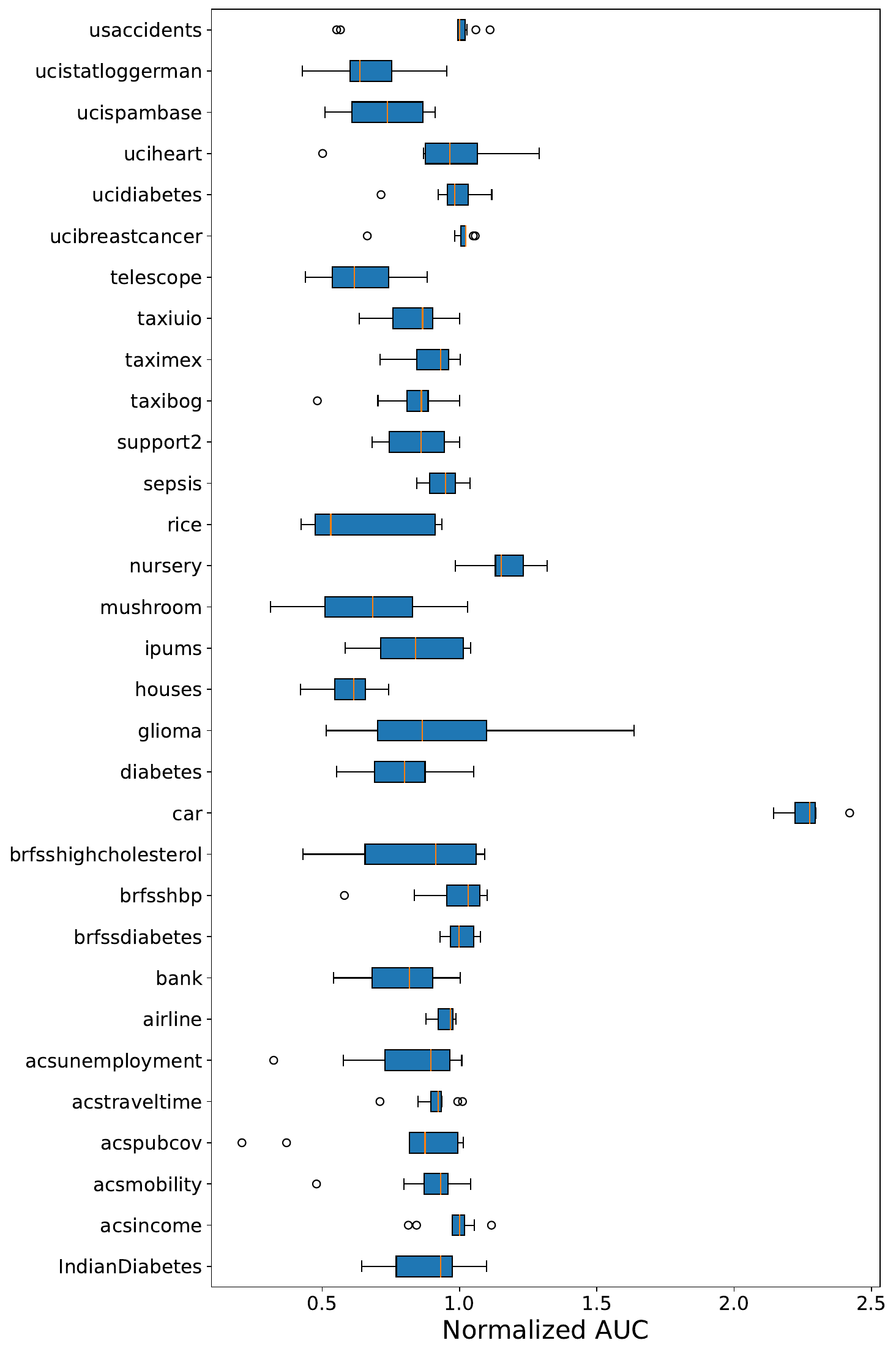}
    \caption{}
    \label{fig:norm-1}
\end{subfigure}%
\begin{subfigure}{0.42 \linewidth}
    \centering
    \includegraphics[width=\linewidth]{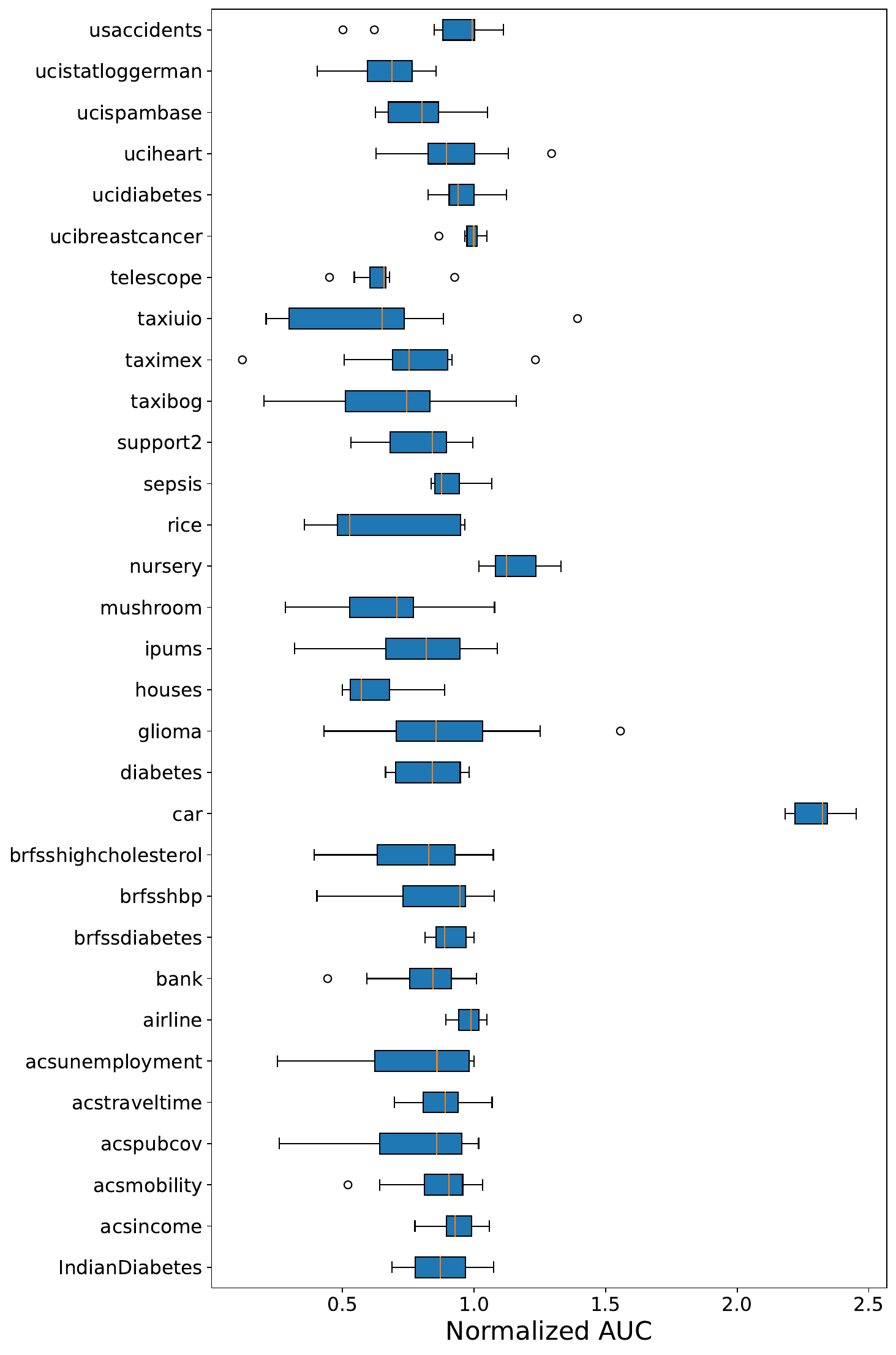}
    \caption{}
    \label{fig:norm-2} 
\end{subfigure}

\begin{subfigure}{0.42\linewidth}
    \centering
    \includegraphics[width=\linewidth]{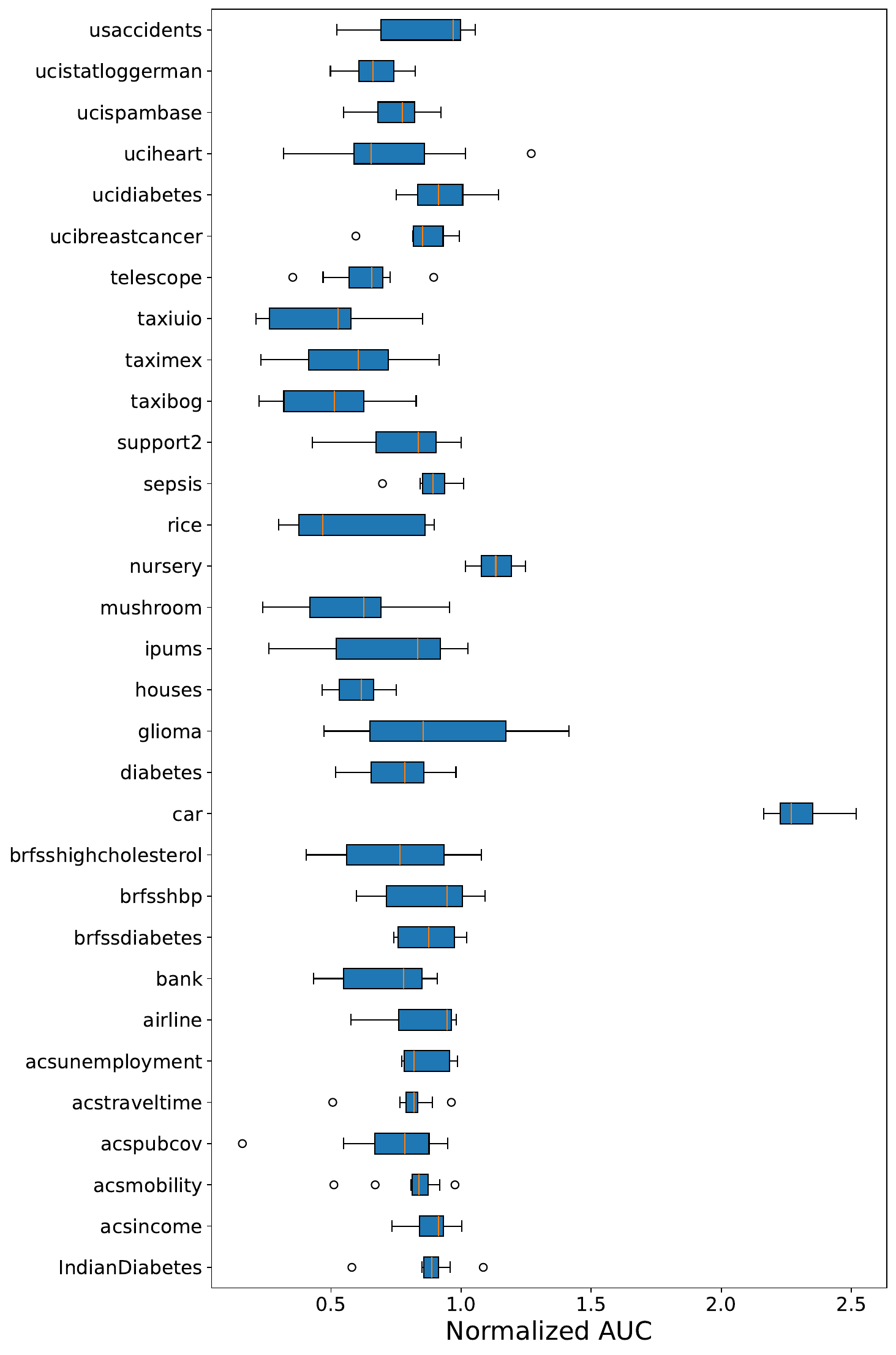}
    \caption{}
    \label{fig:norm-3}
\end{subfigure}%
\begin{subfigure}{0.42 \linewidth}
    \centering
    \includegraphics[width=\linewidth]{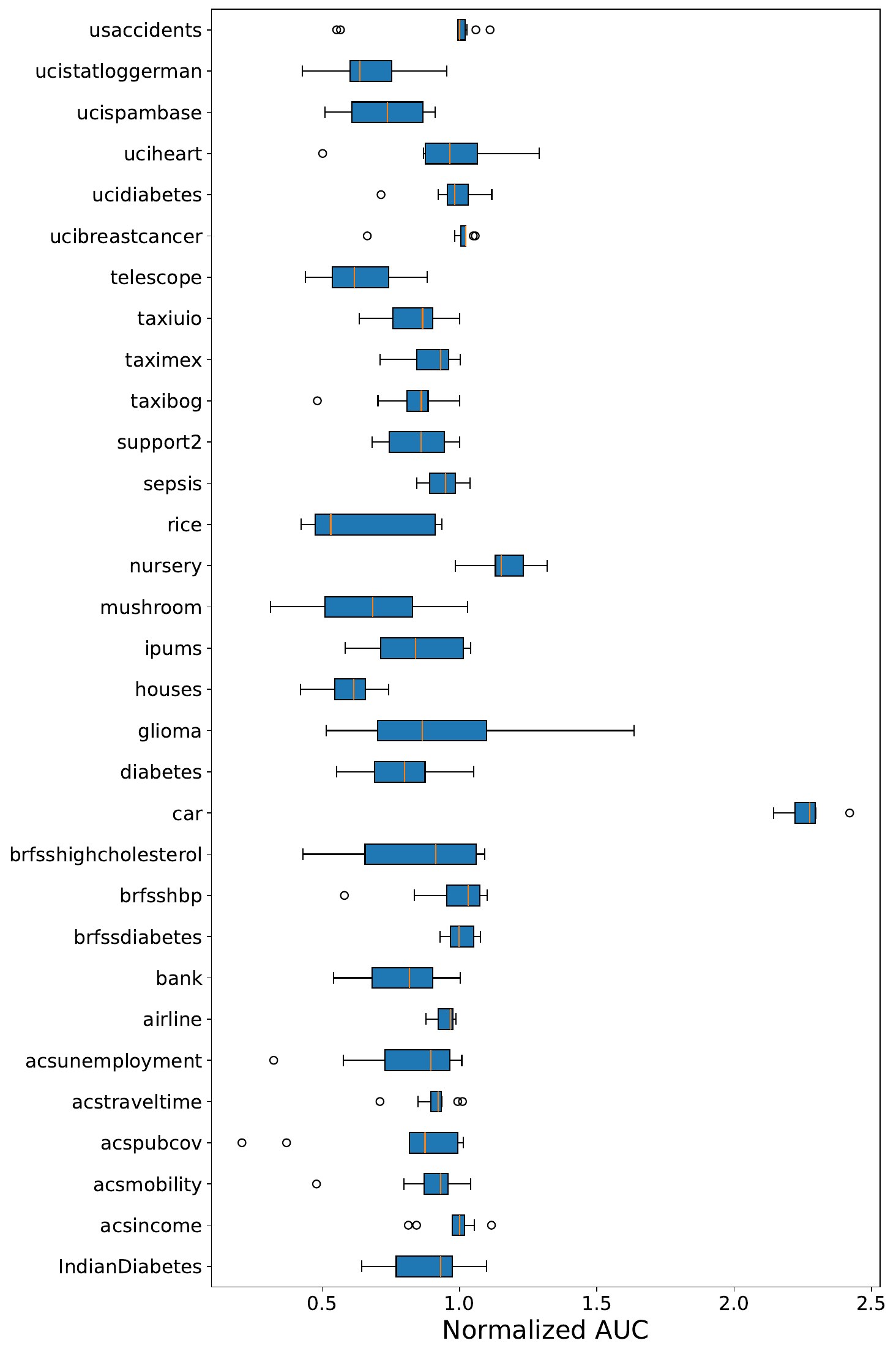}
    \caption{}
    \label{fig:norm-4} 
\end{subfigure}

\caption{Box plots of AUC scores over masked-out columns in the Masking experiment, for all datasets, where each AUC is divided by the AUC achieved by an XGBoost classifier on the same prediction task. Results shown for GPT-4o-mini (a), Llama (b), Mistral (c), and GPT-4o (d).}

\label{fig:norm}

\end{figure}

\clearpage

\subsection{ AUC Scores, Masking Experiment}
\label{appendix:rawauc}

\begin{figure}[H]
    \begin{subfigure}{0.42\linewidth}
    \includegraphics[width=\linewidth]{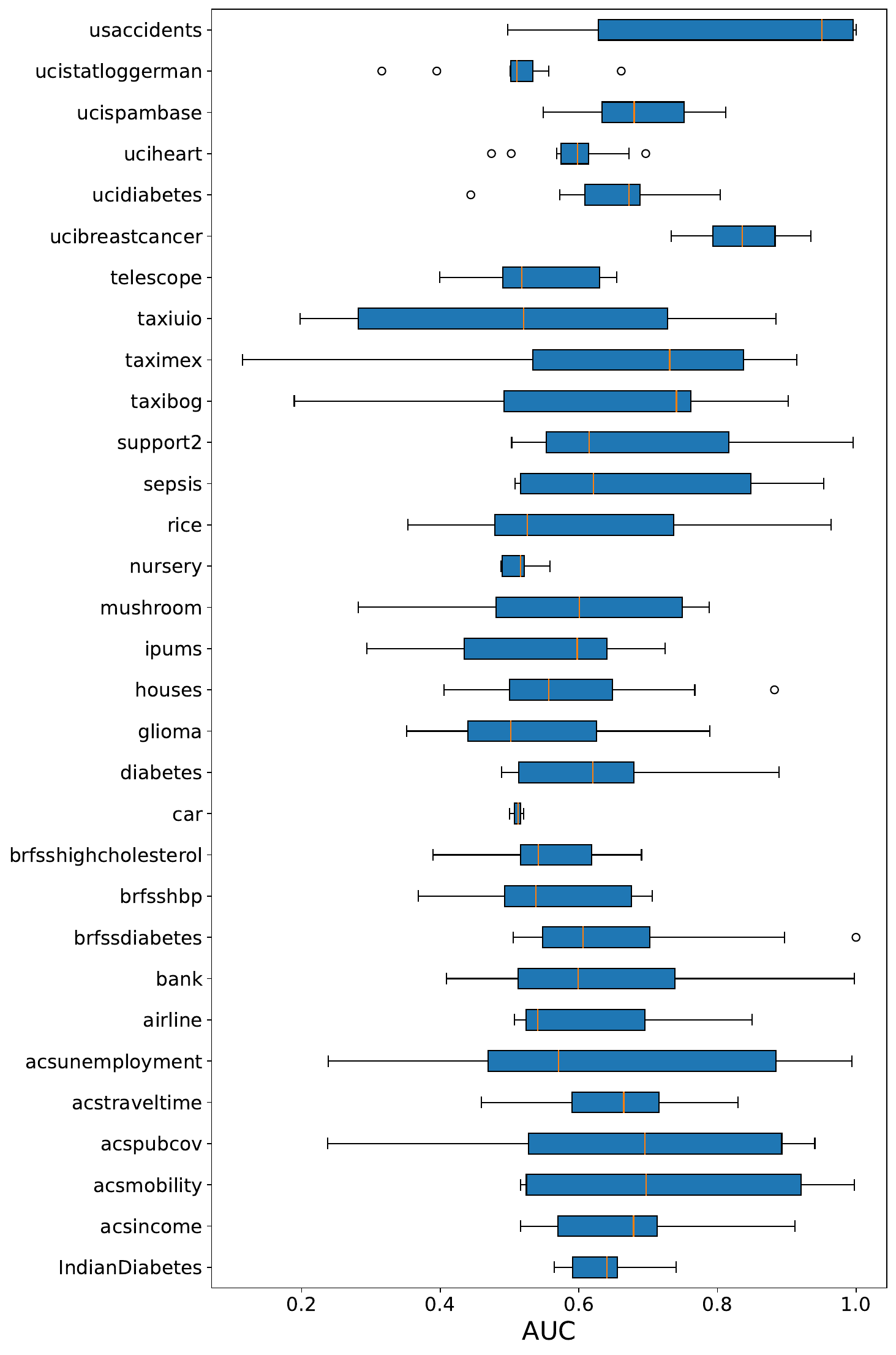}
    \label{fig:sfigraw-1}
    \end{subfigure}

    \begin{subfigure}{0.42\linewidth}
    \centering
    \includegraphics[width=\linewidth]{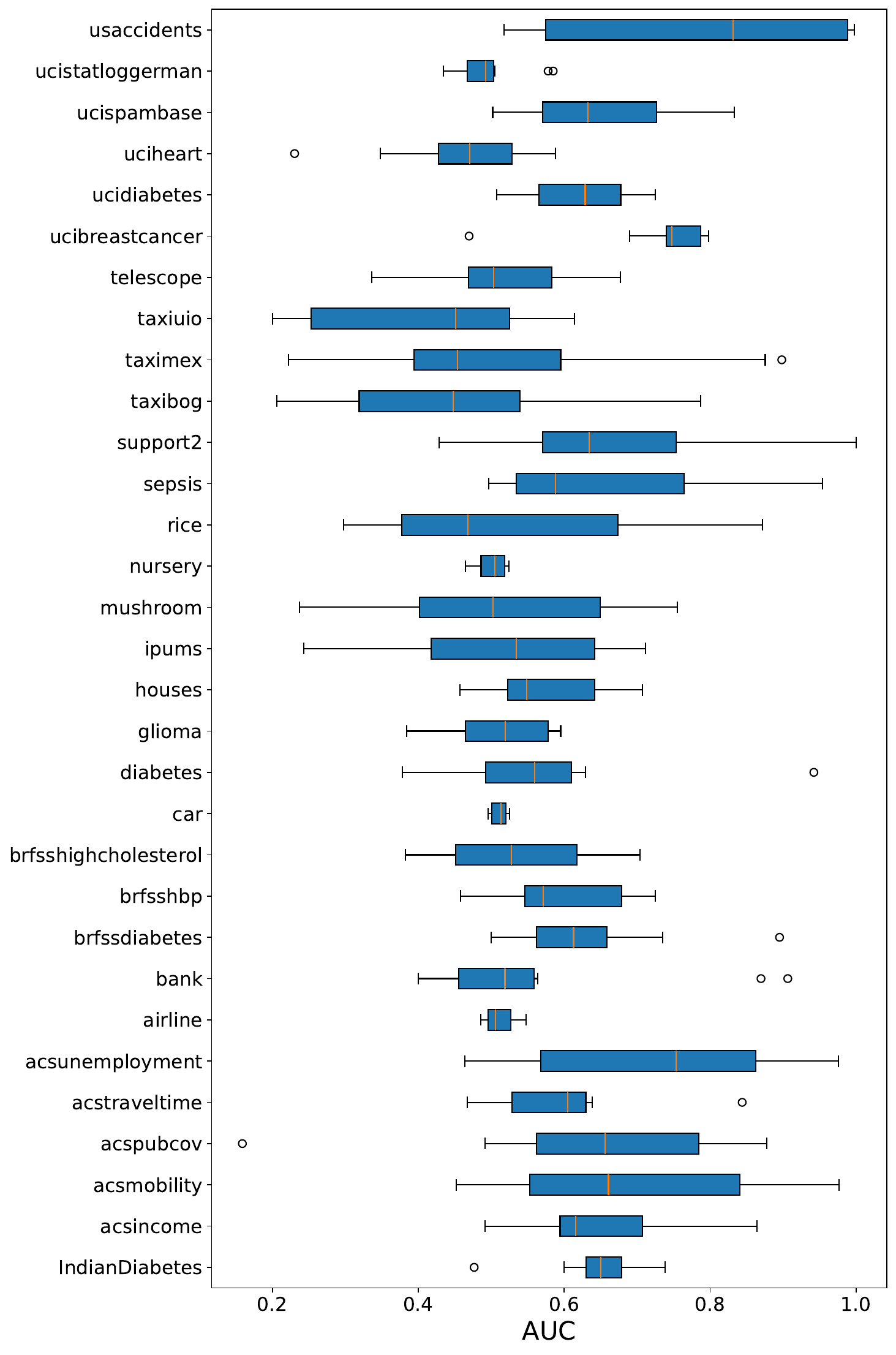}
    \caption{}
    \label{fig:sfigraw-2}
    \end{subfigure}%
    \begin{subfigure}{0.42 \linewidth}
    \centering
    \includegraphics[width=\linewidth]{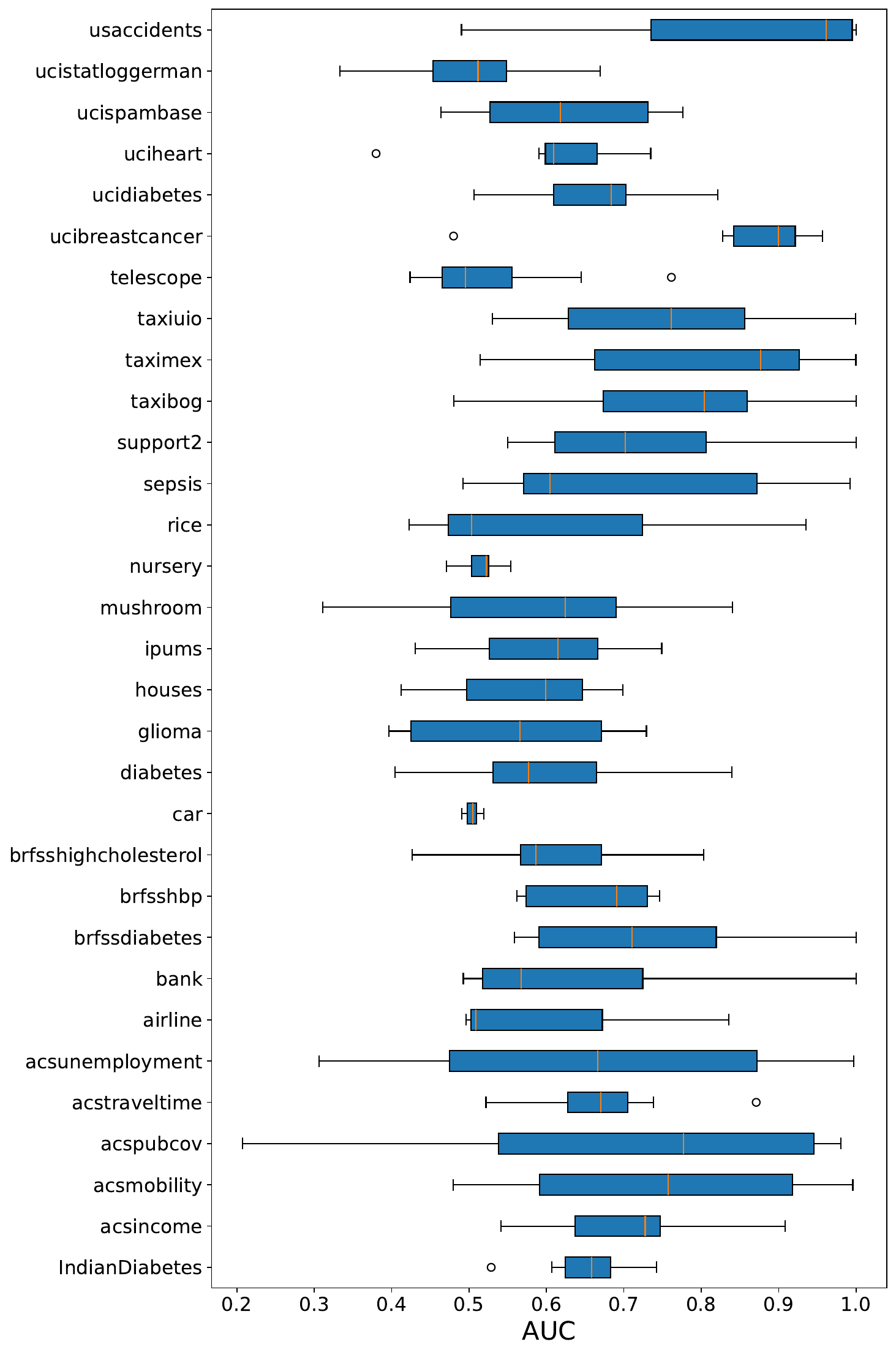}
    \caption{}
    \label{fig:sfigraw-3} 
    \end{subfigure}
\caption{Box plots of AUC scores over masked-out columns in the Masking experiment, for all datasets. Results shown for Llama (a), Mistral (b), and GPT-4o (c).}
\end{figure}

\clearpage

\subsection{Additional Agreement Plots}
\label{appendix:agreement}

\begin{figure}[H]
    \begin{subfigure}{0.42\linewidth}
    \includegraphics[width=\linewidth]{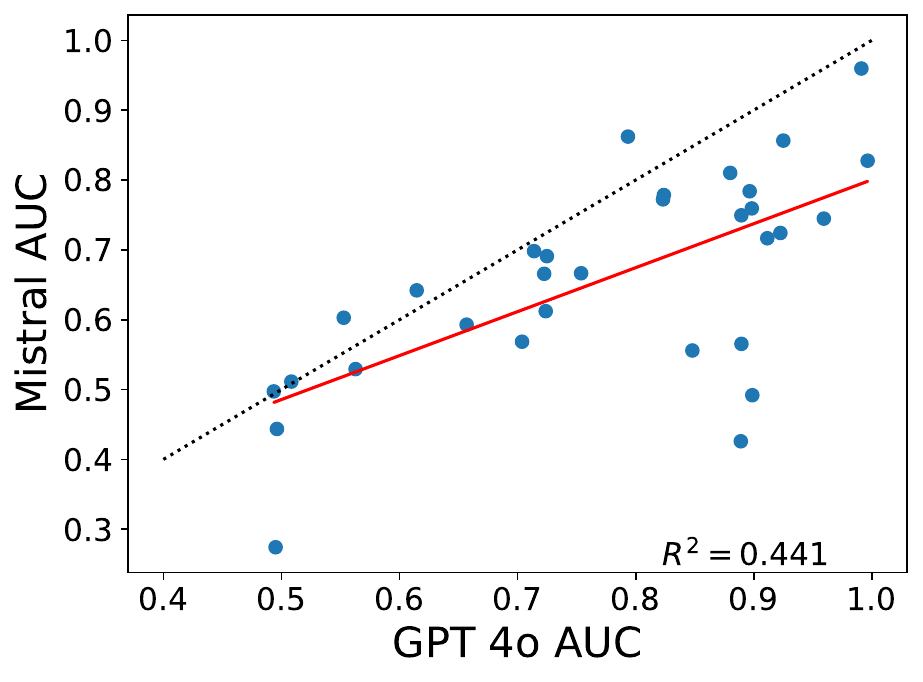}
    \label{fig:agree-1}
    \end{subfigure}

    \begin{subfigure}{0.42\linewidth}
    \centering
    \includegraphics[width=\linewidth]{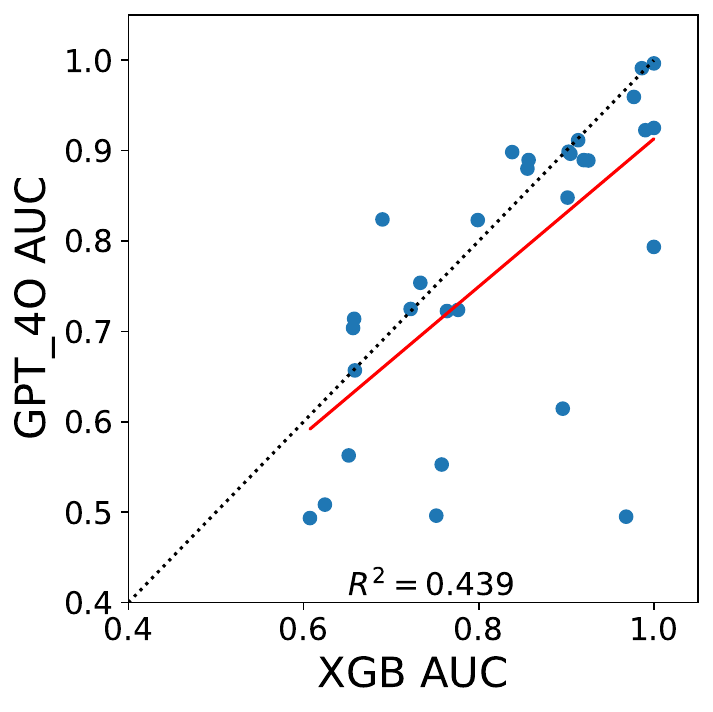}
    \caption{}
    \label{fig:agree-2}
    \end{subfigure}%
    \begin{subfigure}{0.42 \linewidth}
    \centering
    \includegraphics[width=\linewidth]{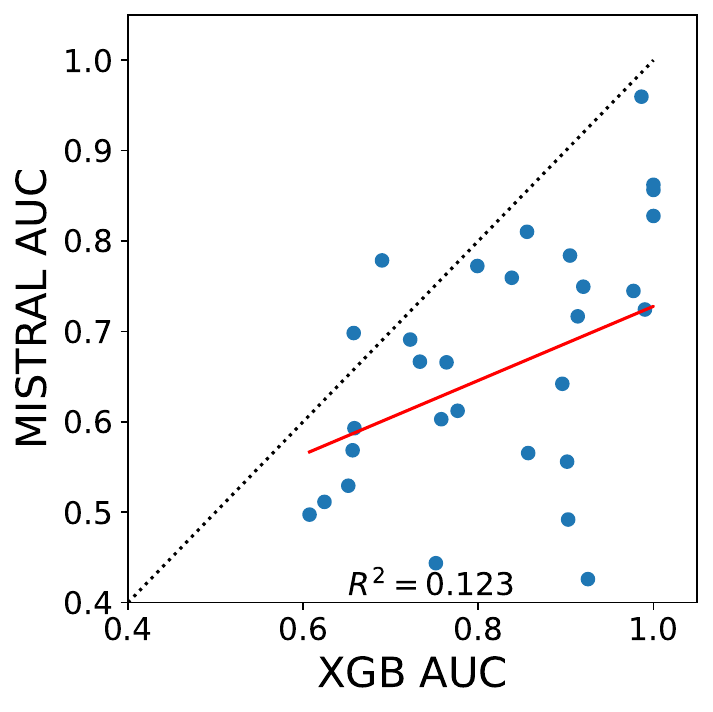}
    \caption{}
    \label{fig:agree-3} 
    \end{subfigure}
\caption{(a) Plot of AUC scores for each of the datasets, for
both GPT-4o and Mistral-7b-Instruct-v0.1. Best-fit
line with R2 value plotted in red. (b,c) Correlation between AUC scores of GPT-4o (b) and Mistral-7b-Instruct-v0.1 (c) over prediction tasks on each dataset, along with the AUCs achieved by training an
XGBoost classification model on a subset of the training
set, and evaluating on a disjoint validation set.}
\end{figure}

\clearpage

\subsection{Additional Calibration Curve Plots}
\label{appendix:fail}

\begin{figure}[!ht]
\begin{subfigure}{.5\linewidth}
    \centering
    \includegraphics[width=\linewidth]{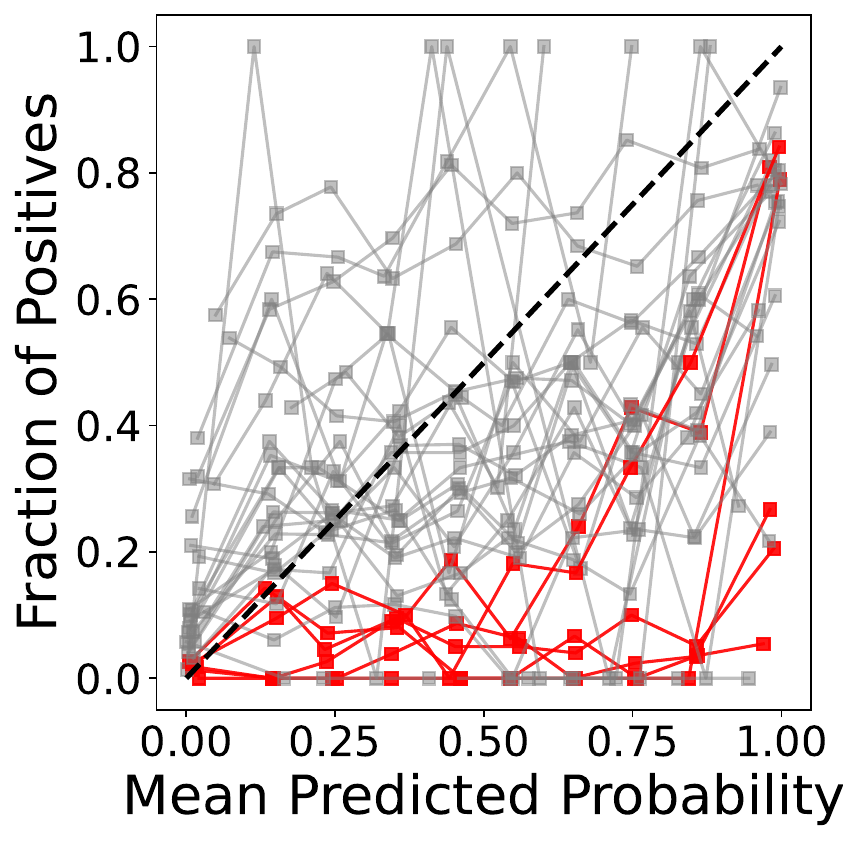}
    \caption{}
    \label{fig:cal-1}
\end{subfigure}%
\begin{subfigure}{.5\linewidth}
    \centering
    \includegraphics[width=\linewidth]{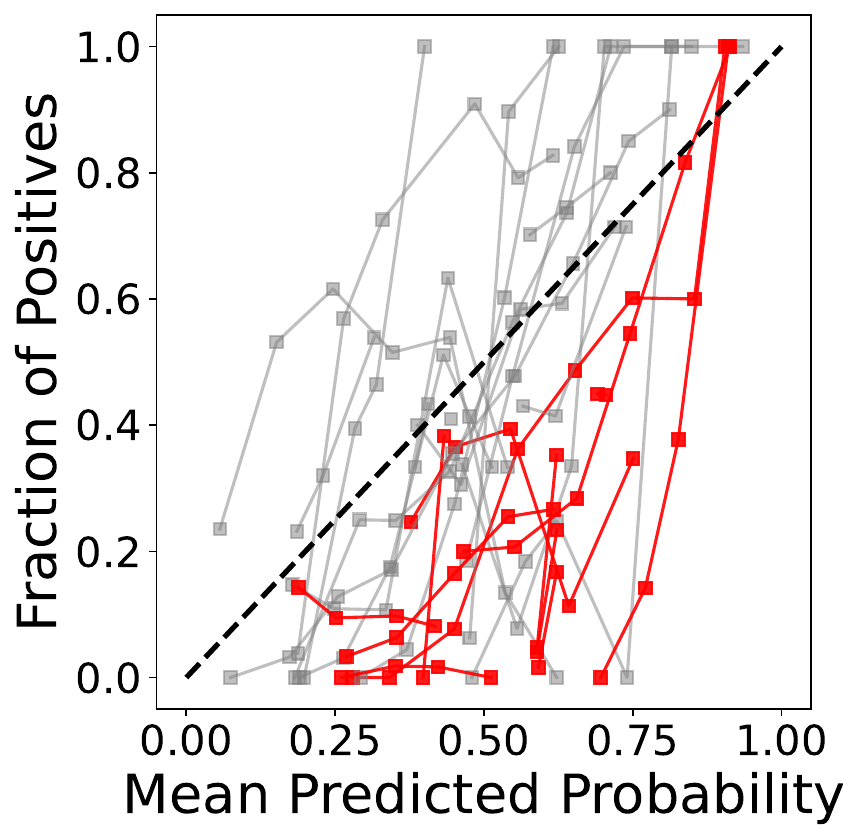}
    \caption{}
    \label{fig:cal-2}
\end{subfigure}

\caption{Calibration curves for GPT-4o (a) and Mistral-7b-Instruct-v0.1 (b) across 31 datasets. Each curve corresponds to a prediction task. Curves crossing the identity line are shown in grey; those consistently above or below are blue and red, respectively. Concretely, all curves that a) are on average 0.2 above the identity line and b) have no points more than .1 below the identity line are colored in blue; curves on average .2 below the identity line and with no points more than .1 above are colored in red.}
\label{fig:cal}
\end{figure}

\clearpage

\subsection{Additional Failure Analysis Plots}
\label{appendix:failpred}

\begin{figure}[!ht]
\begin{subfigure}{.5\linewidth}
    \centering
    \includegraphics[width=\linewidth]{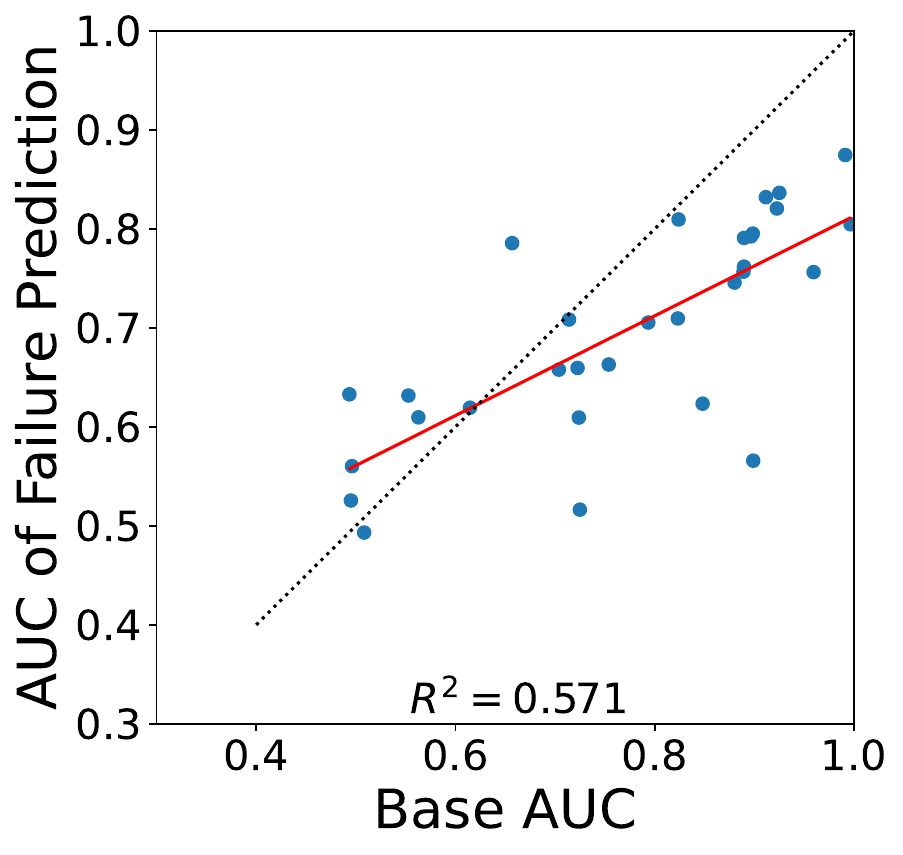}
    \caption{}
    \label{fig:sfigadditionalfail-1}
\end{subfigure}%
\begin{subfigure}{.5\linewidth}
    \centering
    \includegraphics[width=\linewidth]{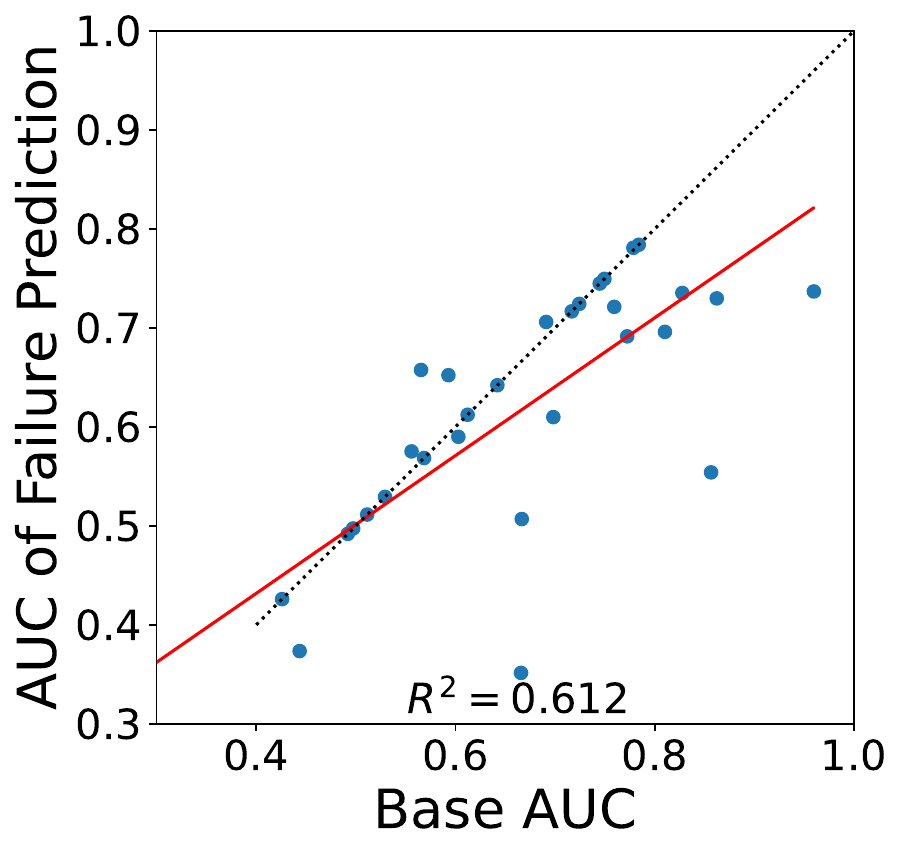}
    \caption{}
    \label{fig:sfigadditionalfail-2}
\end{subfigure}

\caption{
Correlation between AUC scores of failure prediction and predicting the outcome variable for all datasets, for GPT-4o (a) and Mistral-7B-Instruct-v0.1 (b).
}
\label{fig:figadditionalfail}
\end{figure}

\clearpage

\subsection{Additional Regression on AUC}
\label{appendix:appauc}

\begin{figure}[!htbp]

\begin{subfigure}{.5\linewidth}
    \centering
    \includegraphics[width=\linewidth]{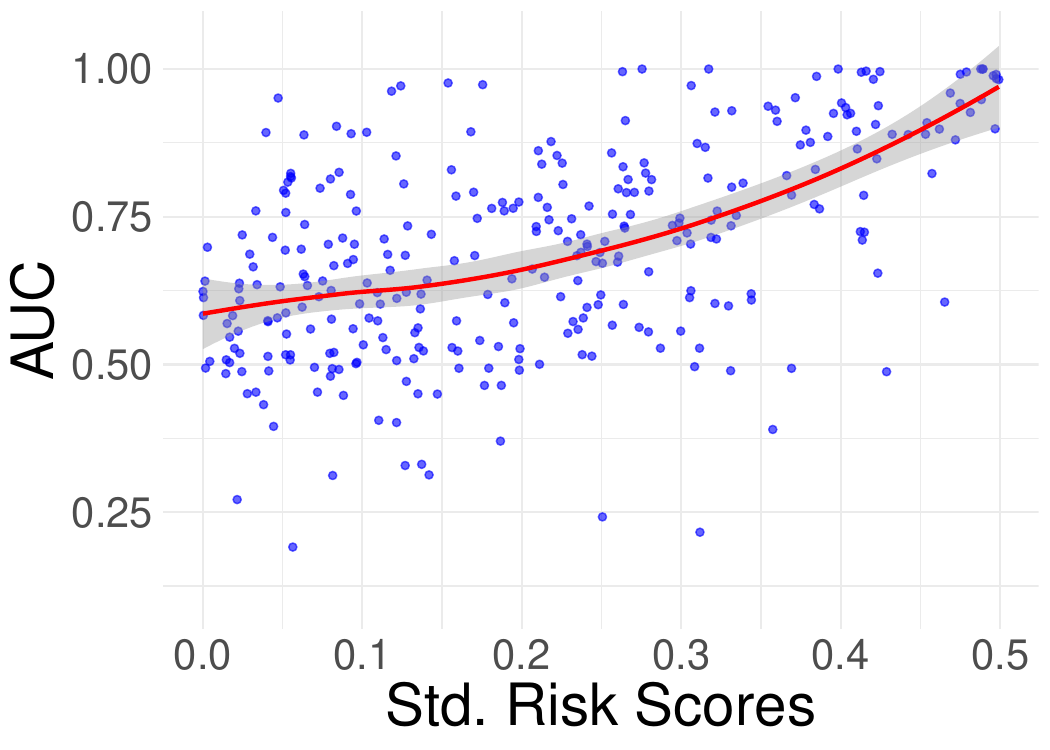}
    \caption{}
    \label{fig:bryan1-1}
\end{subfigure}%
\begin{subfigure}{.5\linewidth}
    \centering
    \includegraphics[width=\linewidth]{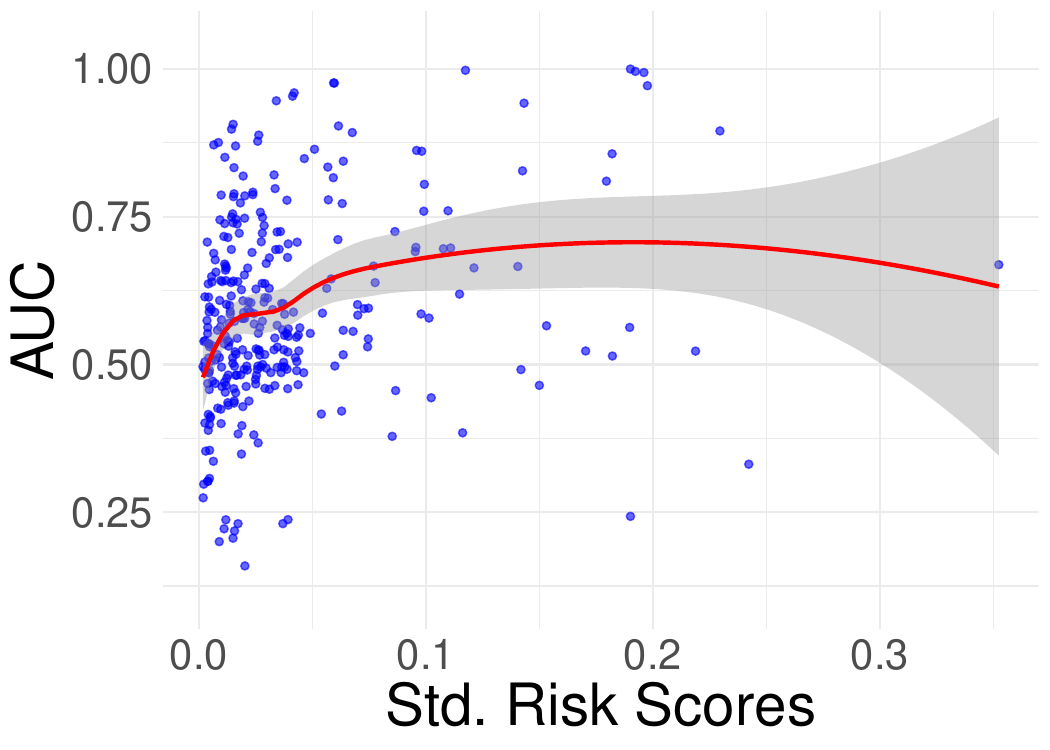}
    \caption{}
    \label{fig:bryan1-2}
\end{subfigure}

\begin{subfigure}{.5\linewidth}
    \centering
    \includegraphics[width=\linewidth]{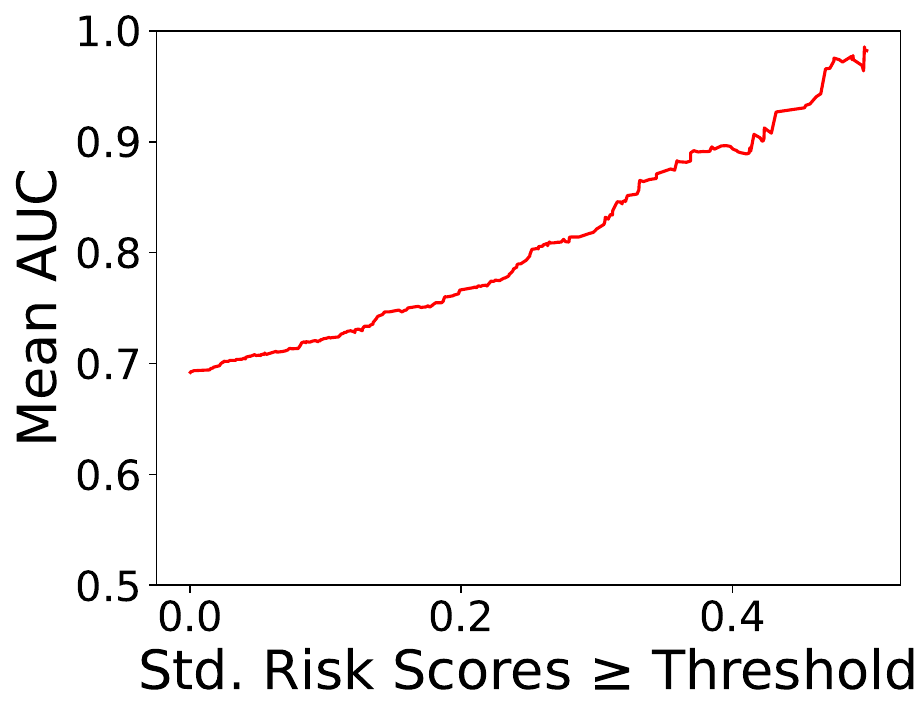}
    \caption{}
    \label{fig:bryan1-3}
\end{subfigure}%
\begin{subfigure}{.5\linewidth}
    \centering
    \includegraphics[width=\linewidth]{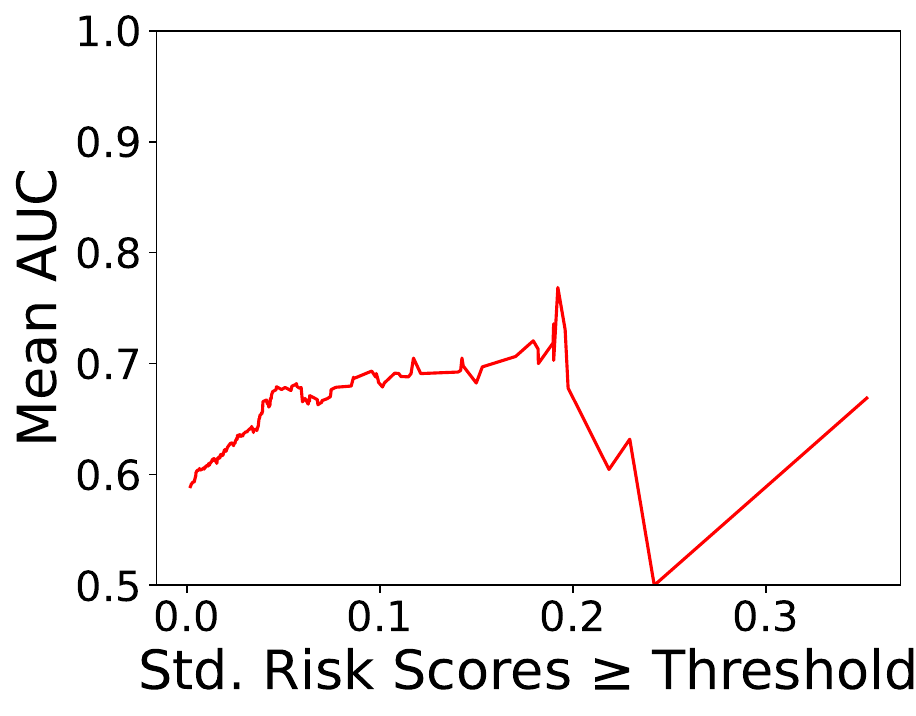}
    \caption{}
    \label{fig:bryan1-4}
\end{subfigure}
\caption{Proxy tasks in the Masking experiment using GPT-4o (a,c) and Mistral-7b-Instruct-v0.1 (b,d), including the original 31 tasks. LOESS curves with 95\% CI shown in (a,b); each point represents predictions on one dataset column. (c,d) show average AUC as the minimum threshold on standard deviation of risk scores increases.
}

\label{fig:bryan1}

\end{figure}

\begin{figure}[!htbp]

\begin{subfigure}{.5\linewidth}
    \centering
    \includegraphics[width=\linewidth]{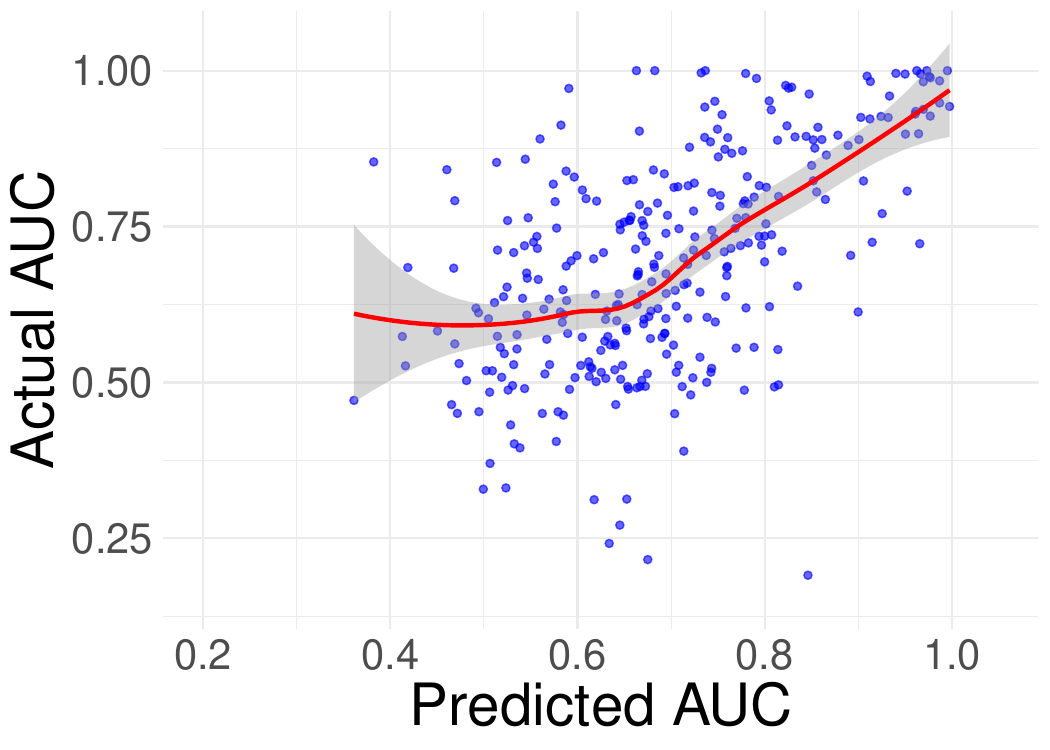}
    \caption{}
    \label{fig:bryan2-1}
\end{subfigure}%
\begin{subfigure}{.5\linewidth}
    \centering
    \includegraphics[width=\linewidth]{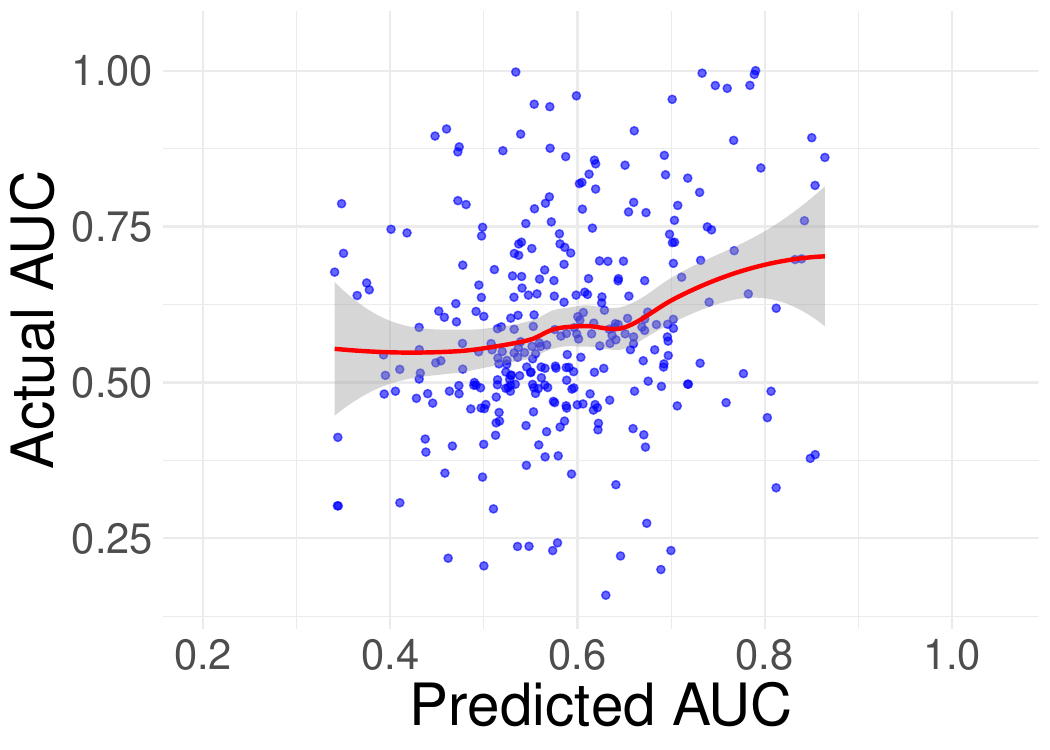}
    \caption{}
    \label{fig:bryan2-2}
\end{subfigure}

\begin{subfigure}{.5\linewidth}
    \centering
    \includegraphics[width=\linewidth]{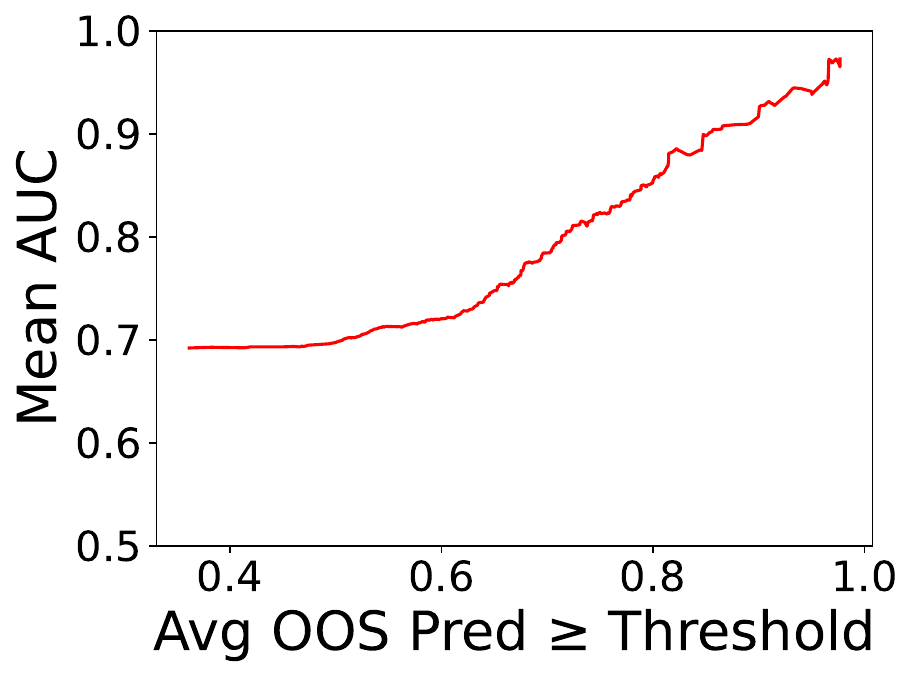}
    \caption{}
    \label{fig:bryan2-3}
\end{subfigure}%
\begin{subfigure}{.5\linewidth}
    \centering
    \includegraphics[width=\linewidth]{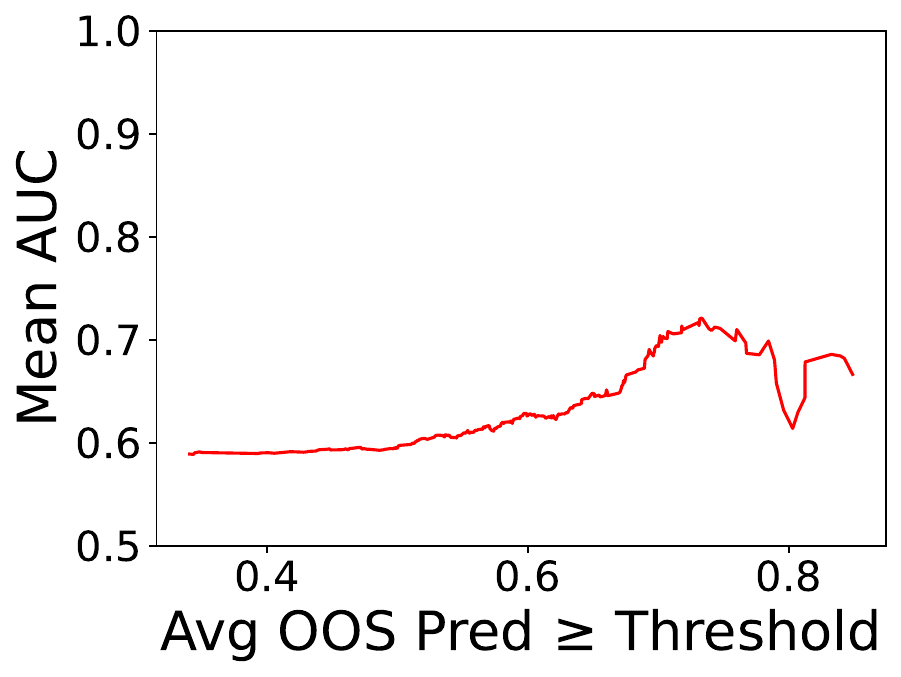}
    \caption{}
    \label{fig:bryan2-4}
\end{subfigure}
\caption{Proxy tasks in the Masking experiment using GPT-4o (a,c) and Mistral-7b-Instruct-v0.1 (b,d), including the original 31 tasks. (a,b) show LOESS fits (with 95\% CI) of actual vs. XGBoost-predicted AUCs, trained via grouped 5-fold cross-validation. Each point represents one prediction task. (c,d) show AUC averages after thresholding on predicted AUCs, analogous to Figures \ref{fig:sfig7-3} and \ref{fig:sfig7-4}.}

\label{fig:bryan2}

\end{figure}

\clearpage

\subsection{Additional CDF Plots}
\label{appendix:cdf}

\begin{figure}[H]

\begin{subfigure}{.5\linewidth}
    \centering
    \includegraphics[width=\linewidth]{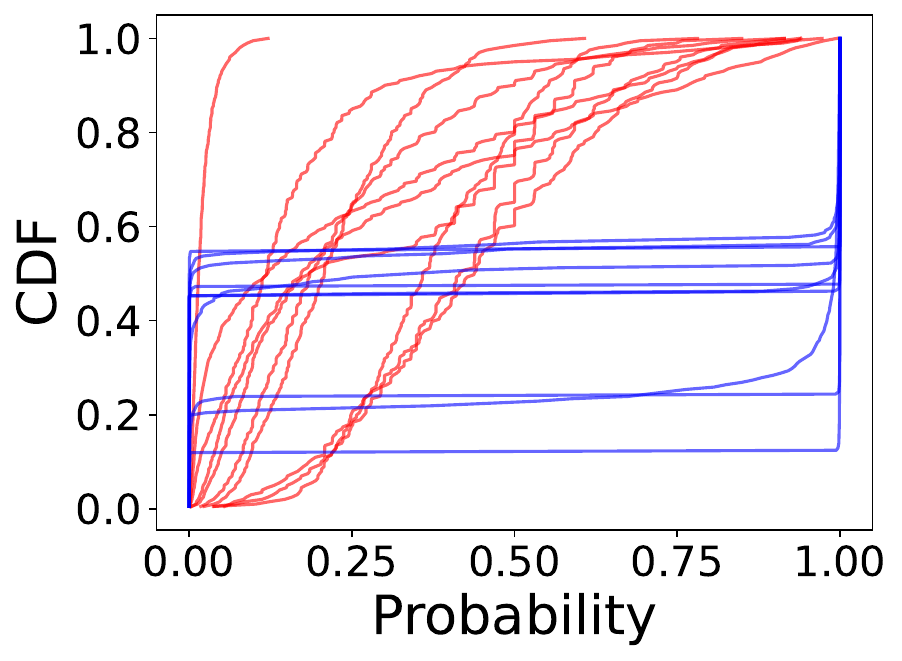}
    \caption{}
    \label{fig:appcdf-1}
\end{subfigure}%
\begin{subfigure}{.5\linewidth}
    \centering
    \includegraphics[width=\linewidth]{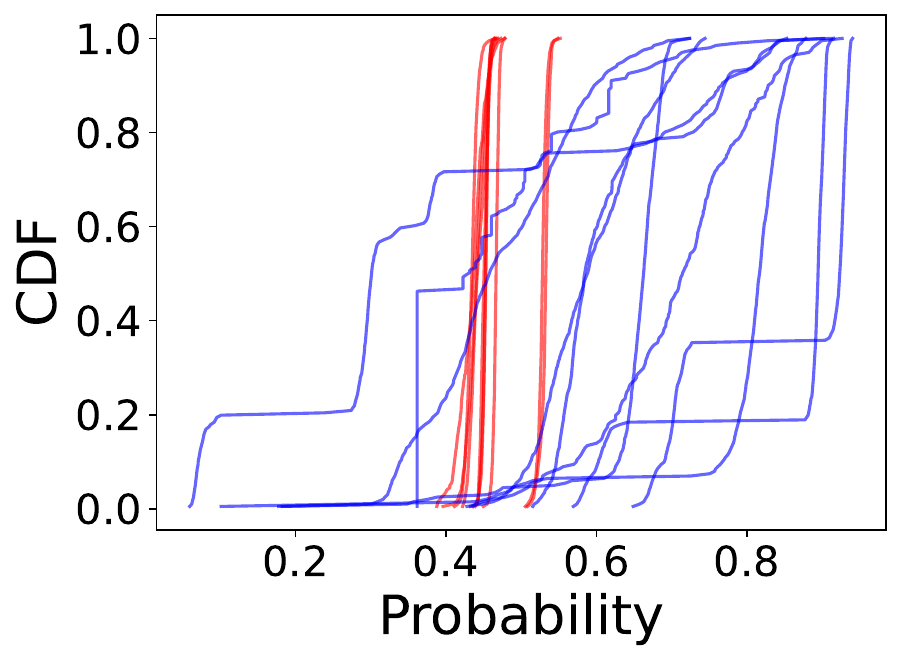}
    \caption{}
    \label{fig:appcdf-2}
\end{subfigure}

\caption{CDFs of the 10 highest (blue) and lowest (red) predicted AUCs over prediction tasks by XGBoost, using 201 percentile values along with standard deviation of risk scores to predict AUC. Shown for GPT-4o (a) and Mistral-7b-Instruct-v0.1 (b).
}

\label{fig:appcdf}

\end{figure}

\clearpage

\subsection{Prompting Templates}
\label{appendix:prompts}

We provide the templates used to generate each of our dataset-level metrics below.

\begin{table}[H]
\centering
\begin{tabular}{l p{6cm} p{6cm}}
\begin{tabularx}{\textwidth}{lXX}

\toprule
\textbf{} & \textbf{Context} & \textbf{Content} \\
\midrule
\textbf{Risk Scores} & \texttt{"Please respond with a single letter."} & \texttt{\$DESCRIPTION OF DATASET\$\textbackslash n\textbackslash n Information: \$SERIALIZED ROW\$\textbackslash n\textbackslash n Question: \$QUESTION\$\textbackslash n A. \$POSITIVE LABEL TEXT\$\textbackslash n B. \$NEGATIVE LABEL TEXT\$} \\
\textbf{Verb. Confidence} & \texttt{\$DESCRIPTION OF DATASET\$} & \texttt{\$SERIALIZED ROW\$ Provide your best guess and the probability that it is correct (0.0 to 1.0) for\textbackslash n the following question. Give ONLY the guess and probability, no other words or\textbackslash n explanation. For example:\textbackslash n\textbackslash n Guess: <most likely guess, as short as possible; not\textbackslash n a complete sentence, just the 
guess!>\textbackslash n Probability: <the probability between 0.0\textbackslash n and 1.0 that your guess is correct, without any extra commentary whatsoever; just\textbackslash n the probability!>\textbackslash n \textbackslash n The question is: \$QUESTION\$} 
\\
\bottomrule
\end{tabularx}
\end{tabular}
\end{table}

\subsection{AI Assistants In Research Or Writing}
\label{appendix:ai}

As our paper centers around the zero-shot capabilities of LLMs for tabular data, all of our experiments necessarily deal with AI assistants (GPT-4o-Mini, Llama-3.1-8b-Instruct) to generate core research results. 
We also utilize AI assistants (Copilot, GPT) for assistance with rewording and clarity during the paper writing process, along with providing starter code for generating plots.

\subsection{Risks}
\label{appendix:risks}
One risk with our findings is the potential misuse of our proposed metrics. While we identify metrics, such as the standard deviation of risk scores, that correlate with LLM performance, these signals should not be interpreted as guarantees of success. Practitioners may be tempted to rely upon our metrics as substitutes for evaluation on labeled data, leading to over-confidence in model outputs. This is particularly of concern in high-stakes domains (e.g., healthcare or finance), where systematically inaccurate predictions carry serious consequences. We emphasize that our metrics are diagnostic tools or guides to which tasks are more promising as opposed to actionable decision rules. They should be used in conjunction with domain knowledge and do not substitute for eventual labeled-data evaluation in high-stakes settings.

\subsection{Hardware Details.}
\label{appendix:hardware}

For GPT-4o-mini and GPT-4o, we conduct all inference via the OpenAI API, and so we do not require any GPU assistance. However, we run Llama-3.1-8b-Instruct and Mistral-7b-Instruct-v0.1 locally with Huggingface. To do this, we utilize a single NVIDIA Tesla V100 GPU, and require 80 GPU hours to run all experiments.

\end{document}